\documentclass[11pt]{article}
\usepackage[utf8]{inputenc}
\usepackage[T1]{fontenc}
\usepackage{amsmath, amsthm, amssymb, amsthm, outlines,parskip, mathtools}
\usepackage[margin=1in]{geometry}
\usepackage{hyperref}
\usepackage[capitalise]{cleveref}
\Crefname{assumption}{Assumption}{Assumptions}
\usepackage{color}
\usepackage{booktabs}
\usepackage{nicefrac}
\usepackage{url, graphicx}
\usepackage{algcompatible}
\usepackage{algorithm}
\usepackage{authblk}
\usepackage{enumitem}
\usepackage{array,multirow}
\usepackage[dvipsnames]{xcolor}

\usepackage{subcaption}
\usepackage[sort,comma,authoryear]{natbib}

\newcommand{\mybigskip}{\bigskip}
\newcommand{\MarginL}{\gamma}
\def\edit{}
\def\blockedit{}
\def\bedit{}
\def\eedit{}

\newcommand{\pr}{\mathbb{P}}
\newcommand{\expect}{\mathbb{E}}

\newcommand{\ind}{\mathbb{I}}
\newcommand{\sign}{\operatorname{sign}}

\newcommand{\rpm}{\raisebox{.2ex}{$\scriptstyle\pm$}}
\newcommand{\leb}{\operatorname{Leb}}

\newcommand{\cube}{\operatorname{Cube}}
\newcommand{\good}{\mathcal{G}}
\newcommand{\many}{\mathcal{M}}
\newcommand{\ogood}{\overline{\mathcal{G}}}
\newcommand{\omany}{\overline{\mathcal{M}}}
\newcommand{\ball}{\mathcal{B}}
\newcommand{\rand}{\mathcal{R}}
\newcommand{\exploit}{\mathcal{E}}
\newcommand{\de}{\mathcal{D}}

\newcommand{\prns}[1]{\left(#1\right)}
\newcommand{\braces}[1]{\left\{#1\right\}}
\newcommand{\bracks}[1]{\left[#1\right]}
\newcommand{\E}{\expect}
\newcommand{\Eb}[1]{\E\bracks{{#1}}}
\newcommand{\ie}{\textit{i.e.}}
\newcommand{\eg}{\textit{e.g.}}

\newcommand{\argmax}{\operatorname{argmax}}

\DeclarePairedDelimiter\ceil{\lceil}{\rceil}
\usepackage{suffix}
\WithSuffix\newcommand\floor*[1]{\mathfrak b(#1)}
\DeclarePairedDelimiter{\abs}{\lvert}{\rvert}
\DeclarePairedDelimiter{\magd}{\lVert}{\rVert}

\newtheorem{theorem}{Theorem}
\newtheorem{corollary}{Corollary}
\newtheorem{lemma}{Lemma}
\newtheorem{proposition}{Proposition}
\newtheorem{assumption}{Assumption}
\newtheorem{definition}{Definition}

\newcommand*\samethanks[1][\value{footnote}]{\footnotemark[#1]}
\title{Smooth Contextual Bandits:\\Bridging the Parametric and Non-differentiable Regret Regimes}
\author{Yichun Hu\thanks{Alphabetical order.},~~~~Nathan Kallus\samethanks,~~~~Xiaojie Mao\samethanks\\Cornell University}
\date{}

\begin{document}
\maketitle
\begin{abstract}
We study a nonparametric contextual bandit problem where the expected reward functions belong to a H\"older class with smoothness parameter $\beta$. We show how this interpolates between two extremes that were previously studied in isolation: non-differentiable bandits ($\beta\leq1$), where rate-optimal regret is achieved by running separate non-contextual bandits in different context regions, and parametric-response bandits (satisfying $\beta=\infty$), where rate-optimal regret can be achieved with minimal or no exploration due to infinite extrapolatability. We develop a novel algorithm that carefully adjusts to all smoothness settings and we prove its regret is rate-optimal by establishing matching upper and lower bounds, recovering the existing results at the two extremes. In this sense, our work bridges the gap between the existing literature on parametric and non-differentiable contextual bandit problems and between bandit algorithms that exclusively use global or local information, shedding light on the crucial interplay of complexity and regret in contextual bandits.\end{abstract}

\section{Introduction}

In many domains, including healthcare and e-commerce, we frequently encounter the following decision-making problem: we sequentially and repeatedly receive context information $X$ (\eg, features of patients or users), need to choose an action \edit{$A\in\mathcal A$ from among $\abs{\mathcal A}<\infty$ actions (\eg, with which therapy if any to treat a patient or which ad if any to show to a user)}, and receive a reward $Y(A)$ (\eg, patient's health outcome or user's click minus ad spot costs) corresponding to the chosen action.
Our goal is to collect the most reward over time.
When contexts $X$ and potential rewards \edit{$\{Y(a):a\in\mathcal A\}$} are drawn from a stationary, but unknown, distribution, this setting is modeled by the stochastic bandit problem \citep{wang2005bandit,MAL-024}. A special case is the multi-armed bandit (MAB) problem where there is no contextual information \citep{Lai:1985:AEA:2609660.2609757,Auer:2002:FAM:599614.599677}.
In these problems, we quantify the quality of an algorithm for choosing actions based on available historical data in terms of its \emph{regret} for every horizon $T$: the expected additional cumulative reward up to time $T$ that we would obtain if we had full knowledge of the stationary context-reward distribution (but not the realizations).
The \emph{minimax regret} is the best (over algorithms) worst-case regret (over problem instances).

The relevant part of the context-reward distribution for maximum-expected-reward decision-making is the conditional mean reward functions, $\eta_a(x)=\Eb{Y(a)\mid X=x}$, for \edit{$a\in\mathcal A$}: if we knew these functions, we would know what arm to pull.
Since we only observe the reward of the chosen action, $Y(A)$, and never that of the unchosen action\edit{s, $Y(a)\;\forall a\neq A$}, we face the oft-noted trade-off between exploration and exploitation: we are motivated to greedily exploit the arm we currently think is best for the context so to collect the highest reward right now, but we also need to explore \edit{other arms} to learn about its expected reward function for fear of missing better options in the future due to lack of information.

The trade-off between exploration and exploitation crucially depends on how we model the relationship between the context and the reward, \ie, $\eta_a$.
\label{minimax regret explanation}\edit{When we restrict $\eta_a$ to a \emph{model}, such as linear functions, minimax regret gives rigorous meaning to our not knowing the particular instance being faced at the onset and needing to \emph{learn} the reward structure. Specifically, it answers the question, given only the information $\eta_a$ belongs to a certain model, how small can one ensure regret is \emph{no matter what} by learning and adapting to \emph{any one} instance.}
In the stochastic setting, previous literature has considered two extreme cases in isolation: a parametric reward model, usually linear \citep{goldenshluger2013,bastani2015online,Bastani17}; and a nonparametric, non-differentiable reward model \citep{Rigollet10,perchet2013,fontaine2019regularized}. We review these below before describing our contribution. We define the problem in complete formality in \cref{section: formulation}.

\begin{figure}[t!]\centering%
\includegraphics[width=0.225\textwidth]{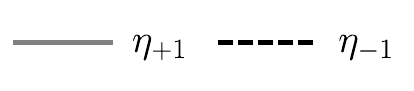}\\%
\begin{subfigure}[t]{0.3\textwidth}\includegraphics[width=\textwidth]{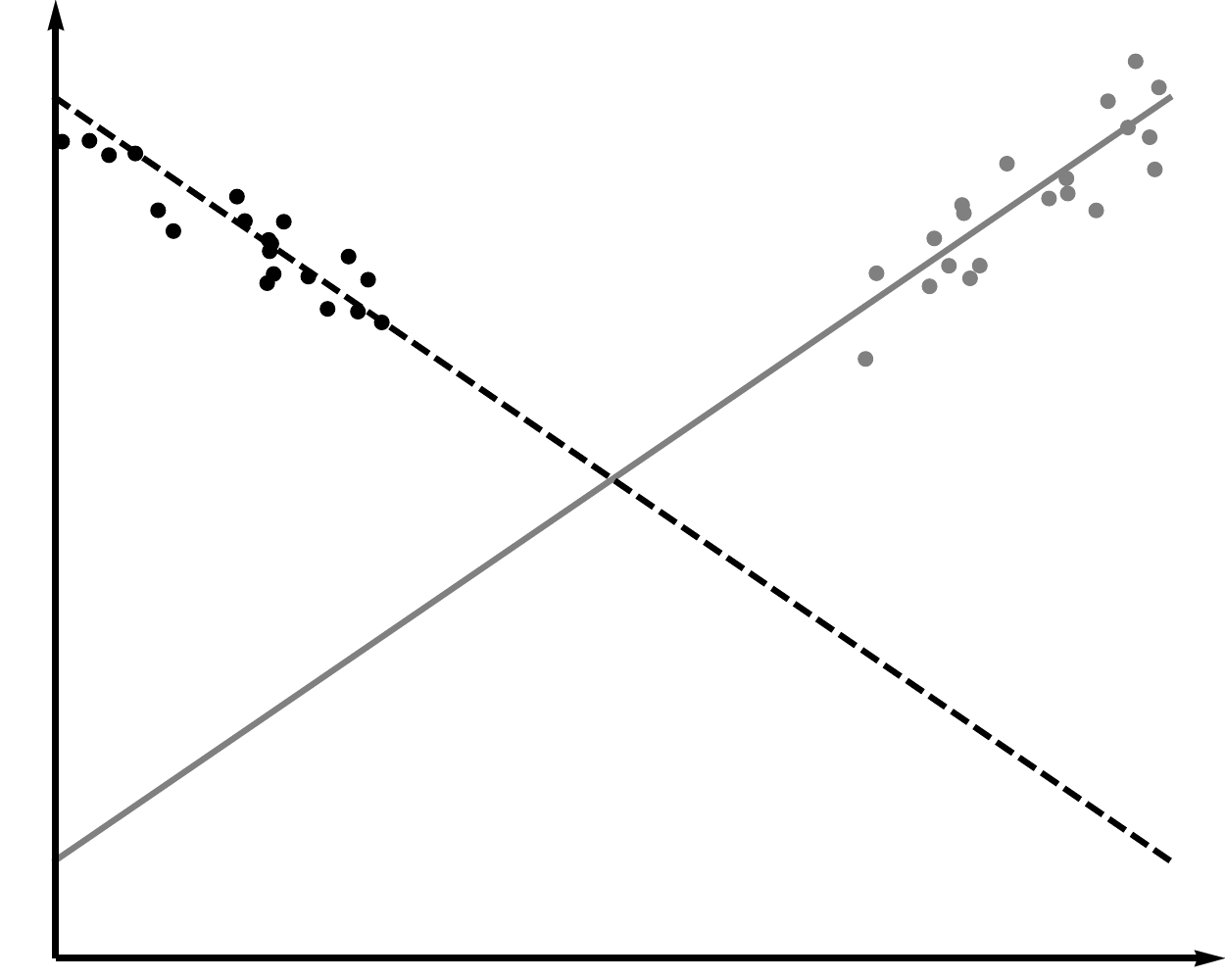}\caption{A linear response bandit: samples in one context region are fully informative about expected rewards in any other context region.}\label{fig: linear bandit}\end{subfigure}%
\hfill\begin{subfigure}[t]{0.3\textwidth}\includegraphics[width=\textwidth]{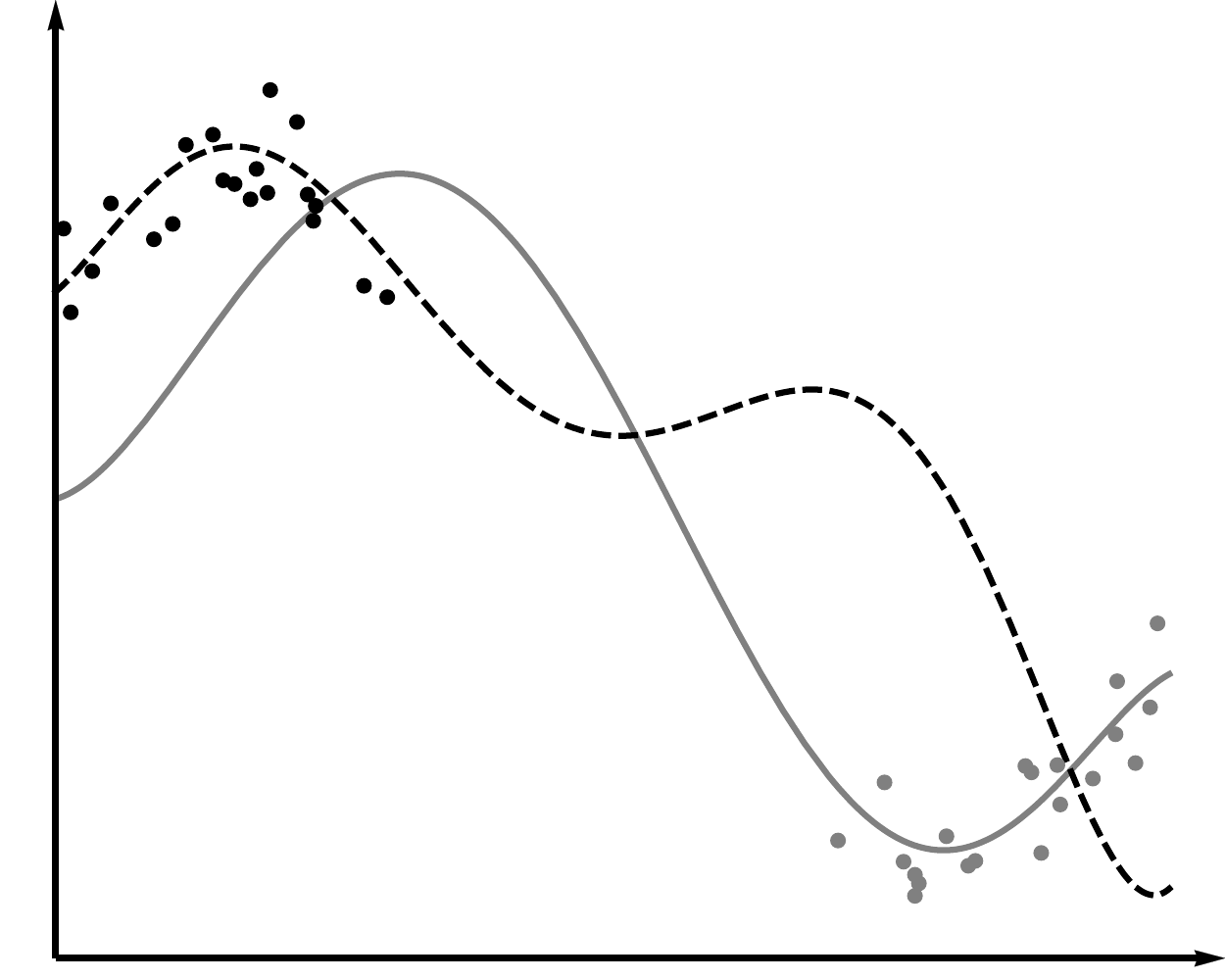}\caption{A nonparametric-response bandit: samples offer only limited extrapolation to learn expected rewards at nearby context values.}\label{fig: smooth bandit}\end{subfigure}%
\hfill\begin{subfigure}[t]{0.3\textwidth}\includegraphics[width=\textwidth]{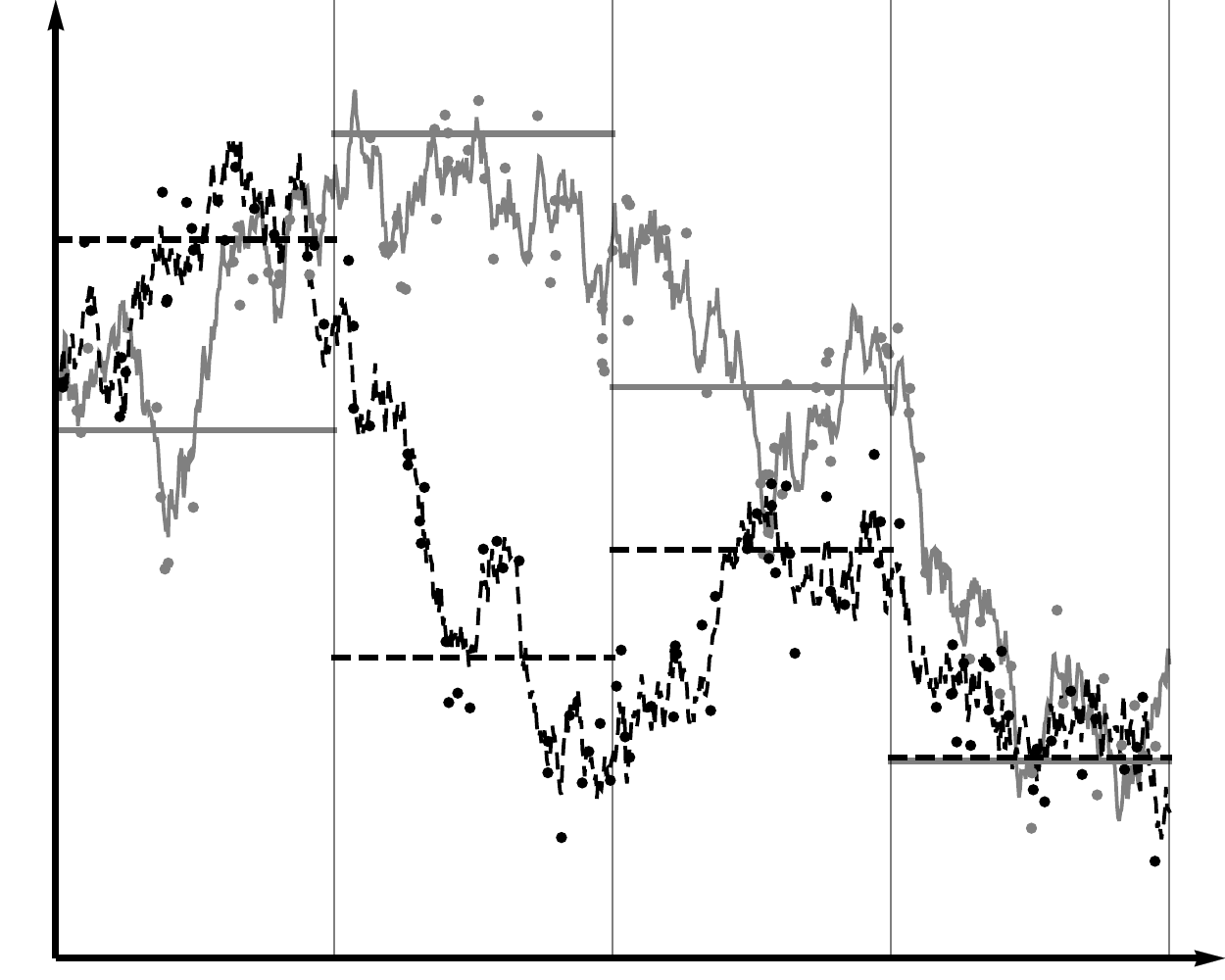}\caption{A non-differentiable-response bandit: rate-optimal regret obtainable by reducing the contextual bandit into multiple, separate MAB problems.}\label{fig: nondifferentiable bandit}\end{subfigure}%
\vspace{\baselineskip}\caption{The fundamental nature of contextual bandit problems depends crucially on the assumed structure of expected reward functions, $\eta_a$\edit{, for two arms $\mathcal A=\{-1,+1\}$}.}
\label{fig: connection}
\end{figure}

\paragraph*{Linear-response bandit.}
One extreme is the linear-response bandit where the expected reward function is assumed to be linear in context, $\eta_a(x)=\theta_a^\top x$ \citep{goldenshluger2013,bastani2015online}.
This parametric assumption imposes a global structure on the expected reward function and permits extrapolation, since \emph{all} samples from arm $a$ are informative about the finite-dimensional parameters $\theta_a$ regardless of the context (see \cref{fig: linear bandit}). Dramatically, this global structure almost entirely obviates the need for forced exploration. In particular, \cite{Bastani17} proved that, under very mild conditions, the greedy algorithm is rate optimal for linear reward models, achieving logarithmic regret. Consequently, the result shows that the classic trade-off that characterizes contextual bandit problems is often not present in linear-response bandits.
Similar behavior generally occurs when we impose other parametric models on expected rewards.
At the same time, while theoretically regret is consequently very low, linear- and parametric-response bandit algorithms may actually have linear regret in practice since the parametric assumption usually fails to hold exactly.

\paragraph*{Non-differentiable nonparametric-response bandit.}
Another line of literature considers nonparametric reward models that satisfy a H\"older continuity condition \citep{Rigollet10,perchet2013}, the strongest form of which is Lipschitz continuity.
In stark contrast to the linear case, such functions need not even be differentiable. \edit{(Note the difference to H\"older \emph{smoothness}, which we use in this paper and which imposes H\"older continuity on \emph{derivatives}.)}
In any nonparametric-response bandit, extrapolation is limited, since only nearby samples are informative about the reward functions at each context value (\cref{fig: smooth bandit}).
Thus, we need to take a more localized learning strategy: we have to actively explore in \emph{every} context region and learn the expected reward functions using nearby samples. In the non-differentiable extreme, \citet{Rigollet10} showed that one can achieve rate-optimal regret by partitioning the context space into small hypercubes and running completely separate MAB algorithms (\eg, UCB) within each hypercube in isolation (\cref{fig: nondifferentiable bandit}). In other words, we can almost ignore the contextual structure because we obtain so little information across contexts. However, the regret is also correspondingly very high.

\paragraph*{Our contribution: smooth contextual bandits.}
In this paper, we consider a nonparametric-response bandit problem with \emph{smooth} expected reward functions. This bridges the gap between the infinitely-smooth linear-response bandit and the unsmooth non-differentiable-response bandit.
We characterize the smoothness of the expected reward functions in terms of the highest order of continuous derivatives, or more generally in terms of a H\"older smoothness parameter $\beta$, which generalizes both non-differentiable H\"older continuous functions ($\beta\leq1)$ and infinitely-extrapolatable functions (such as linear, which we denote by $\beta=\infty$).
\cref{table: literature} summarizes the landscape of the current literature and where our paper lies in terms of this new smoothness perspective and in terms of the sharpness $\alpha$ of the margin (see \cref{assump: margin}).

\begin{table}[t!]\small%
\centering%
\def\arraystretch{2.2}
\begin{tabular}{ccccc}
\toprule
&&\parbox{3cm}{\center~\\[0.25em]$\beta\leq1$}&
\parbox{3cm}{\center\textit{Smoothness}\\[0.25em]$1\leq\beta<\infty$}
&\parbox{3cm}{\center~\\[0.25em]$\beta=\infty$}\\
\midrule
\multirow{3}{*}{\rotatebox[origin=c]{90}{\textit{Margin Sharpness}}}&$0\leq\alpha<1$&\multirow{2}{*}{\rotatebox[origin=c]{45}{\parbox{2cm}{\center\citet{Rigollet10}}}}&\multirow{3}{*}{\rotatebox[origin=c]{90}{\textit{--- This paper ---}}}&\parbox{4.9cm}{\center\citet{Bastani17}}\\
&$\alpha=1$&&&{\parbox{4.9cm}{\center\citet{goldenshluger2013}}}\\
&$\alpha>1$&{\parbox{4.3cm}{\center\citet{perchet2013}}}&&\parbox{4.9cm}{\center\citet{Bastani17}}\\[.2em]
\bottomrule
\end{tabular}%
\caption{%
The lay of the literature on stochastic contextual bandits in terms of our smoothness perspective. For the most part, there has been a significant and wide divide between non-differentiable-response and parametric-response bandits. Our work shows that (up to polylogs) the minimax regret rate $\tilde{\Theta}(T^{\frac{\beta + d - \alpha\beta}{2\beta + d}}\edit{+1})$ reigns across \emph{all} regimes; see also \cref{fig: regret rates}. (Note that additional linear restrictions are made in the $\beta=\infty$ column.)}%
\label{table: literature}%
\end{table}

We propose a novel algorithm for every level of smoothness $1\leq\beta<\infty$ and prove that it achieves the minimax optimal regret rate up to polylogs. In particular, when $\beta>1$, we must leverage information across farther-apart contexts and running separate MAB algorithms will be suboptimal. And, because $\beta<\infty$, we must ensure sufficient exploration everywhere. Thus, our algorithm interpolates between the fully-global learning of the linear-response bandit (which satisfies $\beta = \infty$) and the fully-local learning of the non-differentiable bandit ($0 < \beta \le 1$), according to the smoothness of the expected reward functions. The smoother the expected reward functions, the more global reward information we incorporate.
Moreover, our algorithm judiciously balances exploration and exploitation: it exploits only when we have certainty about which arm is optimal, and it explores economically in a shrinking margin region with fast diminishing error costs.
As a result, our algorithm achieves regret bounded by $\tilde{O}(T^{\frac{\beta + d - \alpha\beta}{2\beta + d}} \edit{+ 1})$ (where $\tilde{O}$ means up to polylogarithmic factors). We show that, for any algorithm, there exists an instance on which it must have regret lower bounded by the same rate, showing that our algorithm is rate optimal and establishing the minimax regret rate for the problem. Consequently, the minimax regret, $\mathcal R_T$, which we define in \cref{sec:minimaxreget}, satisfies $\mathcal R_T=\tilde{\Theta}(T^{\frac{\beta + d - \alpha\beta}{2\beta + d}}\edit{+1})$ and hence $\lim_{T\to\infty}\log(\mathcal R_T)/\log(T)=\frac{\prns{\beta + d - \alpha\beta}_+}{2\beta + d}$.

While this rate has the same \emph{form} as the regret in the non-differentiable case studied by \citet{Rigollet10}, our results extend to the smooth ($\beta>1$) regime where our algorithm can attain much lower regret, arbitrarily approaching polylogarithmic rates as smoothness increases. Our algorithm is fundamentally different, leveraging contextual information from farther away as smoothness increases without deteriorating estimation resolution, and our analysis is necessarily much finer. Our work connects seemingly disparate contextual bandit problems, and reveals the whole spectrum of minimax regret over varying levels of function complexity.

\begin{figure}[t!]\centering
\begin{subfigure}[t]{0.33\textwidth}\includegraphics[width=\textwidth]{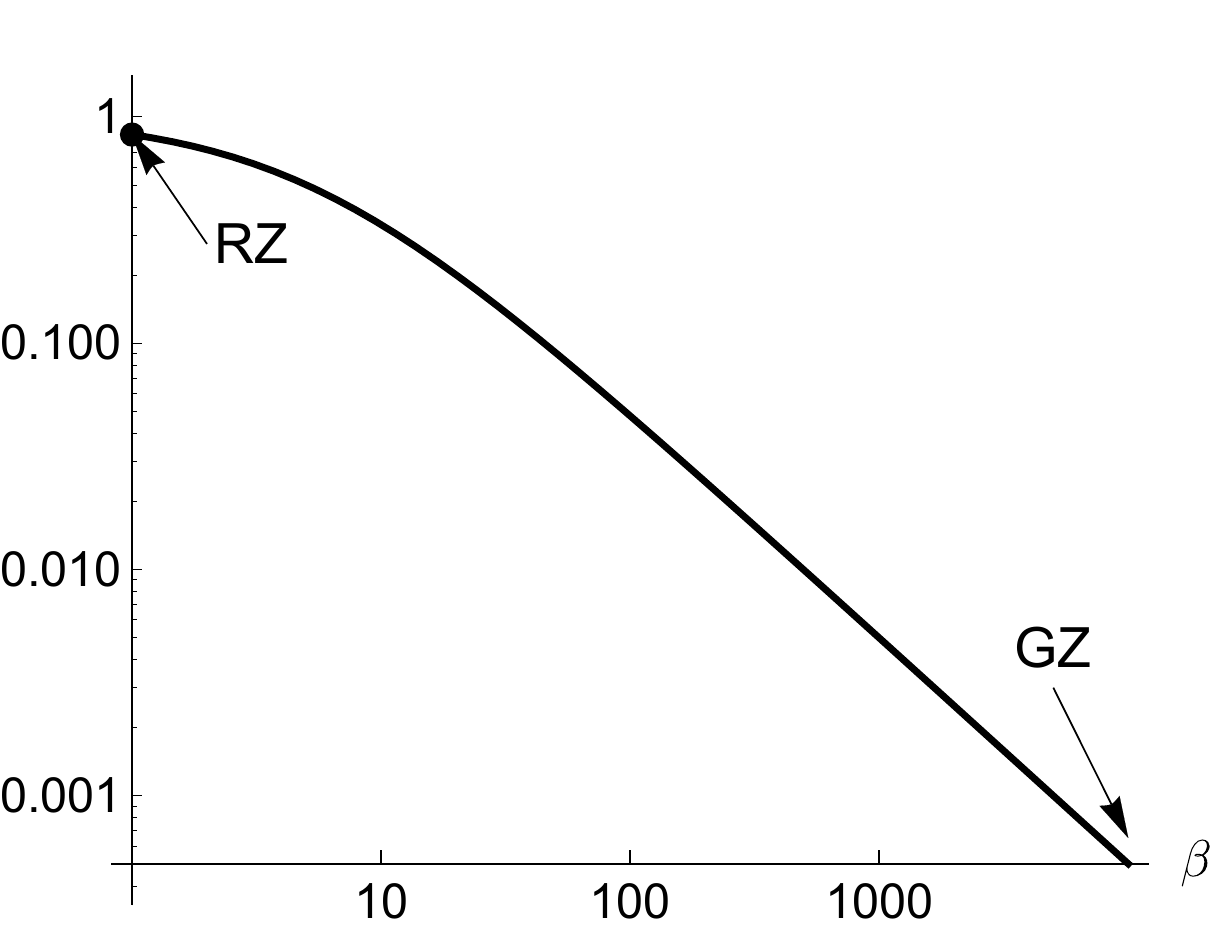}\caption{$\alpha=1,\,d=10$}\label{fig: regret rate alpha 1.0}\end{subfigure}%
\hfill\begin{subfigure}[t]{0.33\textwidth}\includegraphics[width=\textwidth]{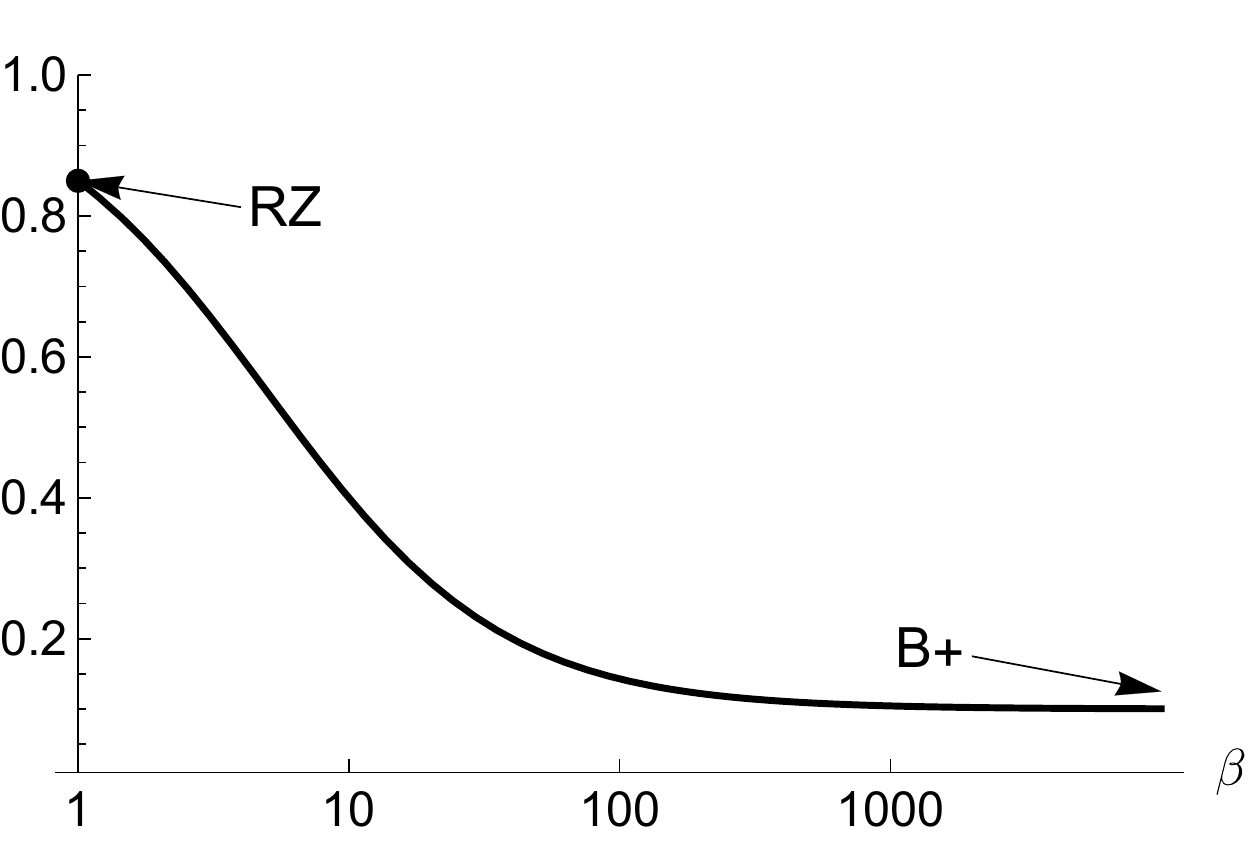}\caption{$\alpha=0.8,\,d=10$}\label{fig: regret rate alpha 0.8}\end{subfigure}%
\hfill\begin{subfigure}[t]{0.33\textwidth}\includegraphics[width=\textwidth]{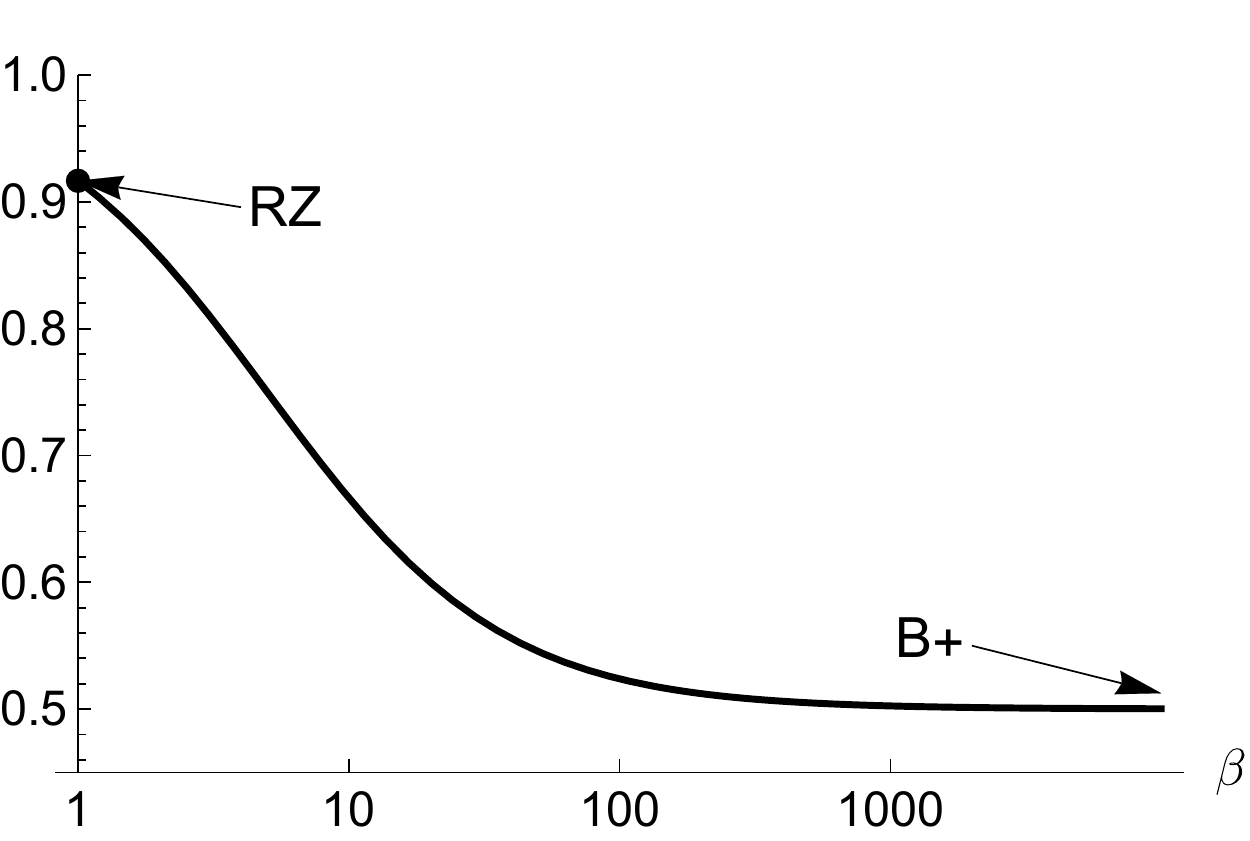}\caption{$\alpha=0,\,d=10$}\label{fig: regret rate alpha 0.0}\end{subfigure}%
\vspace{\baselineskip}\caption{The minimax regret rate exponent, $\lim_{T\to\infty}\log(\mathcal R_T)/\log(T)=\frac{\prns{\beta + d - \alpha\beta}_+}{2\beta + d}$, as shown by our \cref{thm: regret,lb}.
The minimax regret $\mathcal R_T$ is defined in \cref{sec:minimaxreget}.
Existing results shown with arrows only characterize the two extreme regimes (\textsf{RZ} refers to \citealp{Rigollet10}, \textsf{GZ} refers to \citealp{goldenshluger2013}, \textsf{B+} refers to \citealp{Bastani17}). In between, our results reveal the effect of complexity on regret.}\label{fig: regret rates}%
\end{figure}

\subsection{Related Literature} \label{subsection: literature}

\paragraph*{Nonparametric regression.}

Our algorithm leverages nonparametric regression to learn expected reward functions, namely local polynomial regression.
Nonparametric regression seeks to estimate regression (aka, conditional expectation) functions without assuming that they belong to an \emph{a priori} known parametric family.
One of the most popular nonparametric regression methods is the Nadaraya–Watson kernel regression estimator \citep{Nadaraya,watson}, which estimates the conditional expectation at a query point as the weighted average of observed outcomes, weighted by their closeness to the query using a similarity-measuring function known as a kernel.
Local polynomial estimators generalize this by fitting a polynomial by kernel-weighted least squares \citep{stone1977}, where fitting a constant recovers the former. \citet{stone1980} considered function classes with different levels of \edit{smoothness $\beta$} and showed that local polynomial regression achieves rate-optimal point convergence. \citet{stone1982} further showed that a modification of this estimator can achieve rate-optimal convergence in $p$-norm for $0<p\leq\infty$.
There are a variety of other nonparametric estimators that can achieve rate optimality in these classes, such as sieve estimators \citep[\eg, ][]{chen2007large, belloni2015some}, but we do not use these in our algorithm.
For more detail and an exhaustive bibliography on nonparametric regression, see \citet{Tsybakov:2008:INE:1522486}.

Nonparametric regression also has broad applications in decision making. In classification problems, \citet{audibert2007} established fast convergence rates for the 0-1 error of plug-in estimators based on local polynomial regression by leveraging a finite-sample concentration bound. The rate depends on a so-called margin condition number $\alpha$ originally proposed by \citet{mammen1999smooth,tsybakov2004optimal} that quantifies how well-separated the classes are, where larger $\alpha$ corresponds to more separation (see \cref{assump: margin}).
\citet{bertsimas2019predictive} use similar locally-weighted nonparametric regression methods to solve conditional stochastic optimization problems with auxiliary observations and show that this provides model-free asymptotic optimality.

\paragraph*{Contextual bandits.}

While the literature above usually considers an off-line problem with a given exogenous sample of data, the literature on contextual bandit problems considers adaptive data collection and sequential decision-making (see \citealp{MAL-024} for a complete bibliography). Some contextual bandit literature allows for adversarially chosen contexts (\eg, \citealp{Langford:2007:EAC:2981562.2981665, pmlr-v15-beygelzimer11a}), but this leads to high regret and may be too pessimistic in real-world applications. For example, in clinical trials for a non-infectious disease, the treatment decisions for one patient do not have direct impacts on the personal features of the next patient. One line of literature captured this stochastic structure by assuming that contexts and rewards are drawn i.i.d. (independently and identically distributed) from a stationary but unknown distribution \citep[\eg,][]{wang2005bandit,Dudik:2011:EOL:3020548.3020569,Agarwal:2014:TMF:3044805.3045075}. The aforementioned linear- and nonparametric-response bandits both fall in this setting. \citet{Rigollet10,goldenshluger2013,goldenshluger2009woodroofe,perchet2013} introduced the use of the margin condition in this setting to quantify how well-separated the arms are, a well-known determiner of regret in the simpler MAB problem \citep{Lai:1985:AEA:2609660.2609757}.

\citet{goldenshluger2013} assumed a linear model between rewards and covariates for each arm and proposed a novel rate-optimal algorithm that worked by maintaining two sets of parameter estimates for each arm. \citet{Bastani17} showed that the greedy algorithm is optimal under mild covariate diversity conditions. \citet{bastani2015online} considered a sparse linear model and used a LASSO estimator to accommodate high-dimensional contextual features.
While \citet{goldenshluger2013,bastani2015online} assume a sharp margin ($\alpha=1$), \citet{goldenshluger2009woodroofe} also considers more general margin conditions in the one-armed linear-response setting and \citet[Appendix E]{Bastani17} considers these in the multi-armed linear-response setting.
All of the above achieve regret bounds of order $\log T$ under a sharp margin condition ($\alpha=1$).
However, as discussed before, this relies heavily on the fact that every observation is informative about expected rewards everywhere.

\edit{\label{valko discussion}\citet{valko2013finite} assume that arm rewards belong to a reproducing kernel Hilbert space (RKHS) with a bounded kernel function (\eg, Gaussian). While this model considerably generalizes the linear model, it is similar in two crucial ways: the learning rate is similar and extrapolation is still possible. For offline regression in an RKHS, the rate is at worst $O(n^{-1/2})$ and at best $\tilde O(n^{-1})$ \citep[Corollary 6.7]{bartlett2005local}, which stands in stark contrast to the rate possible when only assuming limited differentiability, which only approaches $O(n^{-1/2})$ as the number of derivatives increases infinitely \citep{stone1980,stone1982}. Furthermore, assuming a bounded RKHS norm essentially enables extrapolation: \eg, for the Gaussian kernel, if two functions agree on a nonempty open set they agree everywhere, meaning we can extrapolate from such a subset \citep[Corollary 3.9]{steinwart2006explicit}. In contrast, the lower bound we prove on regret in our problem (\cref{lb}) relies on constructing an example with arbitrary constant values in different regions, forcing one to explore each region as extrapolation is \emph{not} possible. \citet{valko2013finite} indeed obtain a regret bound of $\tilde O(\sqrt{T})$, which matches the bounds for linear response (or, our bound as $\beta\to\infty$) without a  margin condition ($\alpha=0$), as \citet{valko2013finite} indeed do not impose the margin condition.}

\citet{Rigollet10,perchet2013} study the case where we only assume that the expected reward functions are H\"older continuous, \ie, that $\abs{\eta_a(x)-\eta_a(x')}\leq\magd{x-x'}^\beta$. Note that $\beta=1$ corresponds to Lipschitz continuity and is the strongest variant of this assumption, since $\beta>1$ requires the function to be constant and is therefore not considered.
\citet{Rigollet10} studied the two-arm case and obtained optimal minimax-regret rates for margin condition $\alpha\leq1$.
\edit{The rate optimal algorithm in this case (UCBogram) consists of segmenting the context space at the beginning and running separate MAB algorithms in parallel in each segment.
\citet{perchet2013} extended this to multiple arms and any $\alpha\geq0$ 
by proposing a another algorithm (ABSE) that
gradually refines the segmentations of the context space (hence avoiding pulling each arm in each of very many segments when the arm separation is strong) but still \textit{only} uses data \textit{within} each segment to estimate the reward functions in that segment.
Crucially, this hyperlocal approach will no longer be rate optimal when we impose smoothness, where we must use information from \textit{across} such segments to fully leverage reward smoothness.}

\edit{\citet{reeve2018k} also consider Lipschitz expected reward functions ($\beta=1$) but leverage a $k$-neareast neighbor regression algorithm in order to adapt to the underlying dimension of the support of covariates.
The regret bound is the same as \citet{Rigollet10,perchet2013} with $d$ replaced by the underlying dimension, which may be smaller than the ambient dimension. In particular, while they can leverage lower underlying dimension, when it exists, they cannot leverage higher-order differentiability.
\citet{slivkins2011contextual} considers a possibly infinite number of arms and assumes $\eta_a(x)$ is \emph{jointly} Lipschitz in $(a,x)$. When the number of arms is finite, the regret bound matches \citet{Rigollet10,perchet2013} (or, our bound with $\beta=1$) without margin a condition ($\alpha=0$), which \citet{slivkins2011contextual} does not impose.\label{reeve discussion}}

\edit{\label{minimax regret note}Like \citet{goldenshluger2013,Rigollet10,perchet2013}, our work focuses on computing the minimax regret rate, which is defined for a given class of bandit problem instances.
And, like \citet{Rigollet10,perchet2013}, the class of instances we consider is parameterized by a constant $\beta$ controlling the smoothness of expected reward functions and we compute the minimax regret rate for \emph{each} $\beta$. The minimax regret is defined as the infimum over policies of the supremum over instances in the class (see \cref{sec:minimaxreget}).
Although the infimizer (over policies) in the minimax regret cannot know the instance chosen by the inner supremum, it does know the class of instances available to it.
Both the above works and our work therefore compute an upper bound on the minimax regret by exhibiting a policy that \emph{depends} on the class of instances being considered in the supremum, and therefore on $\beta$ in our case.}

\edit{In addition to computing the minimax regret for each $\beta$,
an important supplementary question is \emph{adaptability} to $\beta$: does there exist a policy that does \emph{not} depend on $\beta$ yet achieves the minimax regret rate for \emph{each} $\beta$?
This question depends, of course, on \textit{first} computing the minimax regret rate for each $\beta$.
Since our paper and based on our work, \citet{gur2019smoothnessadaptive} answered this question negatively in general and positively if one further assumes a self-similarity condition on expected reward functions.
Under this assumption, they show that for adaptation it suffices to first explore arms evenly for $\tilde O(1)$ time, then use the collected data to estimate $\beta$ by $\hat\beta$, and then run our non-adaptive algorithm (\cref{alg}) with the smoothness parameter set to $\hat\beta$ if $\hat\beta>1$ or run the non-adaptive algorithm of \citet{perchet2013} with the smoothness parameter set to $\hat\beta$ if $\hat\beta\leq1$.\label{adaptive beta}}

\subsection{Notation}\label{sec:notation}
For any multiple index $r = (r_1, \dots, r_d) \in \mathbb{Z}^d_+$ and any $x = (x_1, \dots, x_d) \in \mathbb{R}^d$, define $|r| = \sum_{i=1}^d r_i$, $r! = r_1!\cdots r_d!$, $x^r = x_1^{r_1} \cdots x_d^{r_d}$, and the differential operator $D^r \coloneqq \frac{\partial^{r_1 + \dots r_d}}{\partial x_1^{r_1}\cdots\partial x_d^{r_d}}$.  We use $\|\cdot\|$ to represent the Euclidean norm, and $\leb[\cdot]$ the Lebesgue measure. We let $\mathcal{B}(x, h) = \{x'\in \mathbb{R}^d: \|x' - x\| \le h\}$ be the ball with center $x$ and radius $h > 0$, and $v_d=\pi^{d/2}/\Gamma(d/2+1)$ the volume of a unit ball in $\mathbb{R}^d$. For any $\beta > 0$, let $\floor*{\beta}=\sup\{i\in\mathbb Z:i<\beta\}$ be the maximal integer that is \emph{strictly} less than $\beta$,
and let $M_\beta$ be the cardinality of the set $\{r \in \mathbb{Z}^d_+: |r| \le  \floor*{\beta}\}$. For an event $A$, the indicator function $\ind(A)$ is equal to $1$ if $A$ is true and $0$ otherwise.
For two scalars $a, b \in \mathbb{R}$, $a \wedge b = \min\{a, b\}$ and $a \vee b = \max\{a, b\}$.
For a matrix $\mathcal{A}$, its minimum eigenvalue is denoted as $\lambda_{\min}(\mathcal{A})$.
For two functions $f_1(T) > 0$ and $f_2(T) > 0$, we use the standard notation for asymptotic order: $f_1(T) = O(f_2(T))$ represents $\limsup_{T \to \infty} \frac{f_1(T)}{f_2(T)} < \infty$, $f_1(T) = \Omega(f_2(T))$ represents $\liminf_{T \to \infty} \frac{f_1(T)}{f_2(T)} > 0$, and $f_1(T) = \Theta(f_2(T))$ represents simultaneously $f_1(T) = \Omega(f_2(T))$ and $f_1(T) = O(f_2(T))$. We use $\tilde{O}$, $\tilde{\Omega}$, $\tilde{\Theta}$ to represent the same order relationship up to polylogarithmic factors. For example, $f_1(T) = \tilde{O}(f_2(T))$ means $\limsup_{T \to \infty} \frac{f_1(T)}{\operatorname{polylog}(T)f_2(T)} < \infty$ for a polylogarithmic function $\operatorname{polylog}(T)$.

\subsection{Organization}

The rest of the paper is organized as follows. In  \cref{section: formulation}, we formally introduce the smooth nonparametric bandit problem and assumptions.
\edit{For a lucid exposition we focus on the setting of two arms in the main text and study the more general multi-arm setting in \cref{section: k arm}.}
We describes our proposed algorithm in \cref{section: algorithm}. In \cref{section: theory}, we analyze our algorithm theoretically: we derive an upper bound on the regret of our algorithm in \cref{section: ub}, and we prove a matching lower bound on the regret of any algorithm in \cref{section: lower bound}. We conclude our paper in \cref{section: conclusion}. While proof techniques are outlined, complete proof details are relegated to the appendix.

\section{The Smooth Contextual Bandit Problem} \label{section: formulation}
In this section, we formulate the smooth contextual bandit problem that we consider in this paper.
We break up this formulation into parts, explaining the significance or necessity of each part separately.
\edit{We focus on the two-armed smooth contextual bandit problem, letting $\mathcal A=\{-1,+1\}$. We extend the problem, our algorithm, and our analysis to multi-armed problems in \cref{section: k arm}.}

\subsection{Two-Armed Stochastic Contextual Bandits}

Consider the following two-armed contextual bandit problem. For $t=1,2,\dots$, nature draws $(X_t, Y_t(1), Y_t(-1))$ i.i.d. from a common distribution $\pr$ of $(X, Y(1), Y(-1))$, where $X\in \mathcal{X} \subseteq \mathbb{R}^d$ is the context (covariate),  and $Y(\pm 1) \in [0, 1]$ is random rewards corresponding to arm $\pm 1$.
At each time step $t$, the decision maker observes the context $X_t$, pulls an arm $A_t \in\{-1,1\}$ according to the observed context and history so far, and then obtains the reward $Y_t = Y_t(A_t)$ of the chosen arm.
Specifically, an admissible policy (allocation rule), $\pi = \{\pi_t\}$, is a sequence of \emph{random} functions $\pi_t: \mathcal{X} \to \{-1, 1\}$ such that, for each $t$, $\pi_t$ is conditionally independent of $\prns{X_1,A_1,Y_1(1),Y_1(-1),\dots}$, given $\prns{X_1,A_1, Y_1,\dots,X_{t-1},A_{t-1}, Y_{t-1}}$, where we let $A_t=\pi_t(X_t)$, $Y_t=Y_t(A_t)$.

For $x \in \mathcal{X}$, we denote the conditional expected reward functions as
$$\eta_{\pm 1}(x) = \expect[Y(\pm 1) \mid X = x],$$
and the conditional average treatment effect (CATE) of pulling arm $1$ versus arm $-1$ as
\[
\tau(x) = \expect[Y(1) \mid X = x] - \expect[Y(-1) \mid X = x] = \eta_1(x) - \eta_{-1}(x).
\]
Obviously, if we had full knowledge of the regression functions $\eta_{\pm 1}$ or the CATE function $\tau$, the optimal decision at each time step would be the oracle policy $\pi^*$ that always pulls the arm with higher expected reward given $X_t$ and regardless of history, namely,
\begin{align}\label{eq: oracle}
	\pi^*(x) =  \ind(\tau(x) \ge 0) - \ind(\tau(x) < 0)
    \in \operatorname{argmax}_{a \in \{-1,1\}} \eta_{a}(x).
\end{align}
However, since we do not know these functions, the oracle policy is infeasible in practice.
We measure the performance of a policy $\pi$ by its \textit{(expected cumulative) regret} compared to the oracle policy $\pi^*$ up to any time $T$, which quantifies how much the policy $\pi$ is inferior to the oracle policy $\pi^*$:
\begin{equation}
  \textstyle R_T(\pi) = \expect  \left[\sum_{t = 1}^T \left(Y_t(\pi^*(X_t)) - Y_t(\pi_t(X_t))\right)\right].
\end{equation}
The growth of this function in $T$ quantifies the quality of $\pi$.

\subsection{Smooth Rewards}

In this paper, we aim to construct a decision policy that achieves low regret \emph{without} strong parametric assumptions on the expected reward functions. We instead focus on expected reward functions restricted to a H\"older class of functions. This is the key restriction characterizing the nature of the bandit problem we consider.
\mybigskip
\begin{definition}[H\"older class of functions]
A function $\eta:\mathcal X\to[0,1]$ belongs to the $(\beta,L,\mathcal{X})$-H\"older class of functions if it is $\floor*{\beta}$-times continuously differentiable and for any $x, x' \in \mathcal{X}$,
    \begin{equation} \label{eq: holder}
    \left\vert \eta(x') - \sum_{|r| \le \floor*{\beta}} \frac{(x' - x)^r}{r!}D^r \eta(x) \right\vert  \le L\|x' - x\|^{\beta}.
    \end{equation}
\end{definition}
\mybigskip
Recall that $\floor*{\beta}$ is the largest integer \emph{strictly} smaller than $\beta$.
When $\beta\leq1$, \cref{eq: holder} reduces to H\"older continuity (\ie, $\abs{\eta(x)-\eta(x')}\leq L\|x' - x\|^{\beta}$), as considered in previous non-differentiable bandit literature \citep{Rigollet10,perchet2013}. When $\beta>1$, $\floor*{\beta}$ is the highest order of continuous derivatives. For example, when $\mathcal X$ is compact, $k$-times continuously differentiable functions are $(k,L, \mathcal{X})$-H\"older for some $L$.
Polynomials of bounded degree $k$ are $(\beta, 0, \mathcal{X})$-H\"older for all $\beta>k$.

\label{comment on beta geq 1}In this paper we focus on $\beta\geq1$, which crucially includes the smooth case ($\beta>1$).
\mybigskip
\begin{assumption}[Smooth Conditional Expected Rewards] \label{assump: smoothness}
For $a = \rpm 1$, $\eta_a$ is $(\beta,L,\mathcal{X})$-H\"older for $\beta\geq1$ and is also $(1,L_1,\mathcal{X})$-H\"older.
\end{assumption}

Given a function that is $(\beta,L,\mathcal X)$-H\"older on a compact $\mathcal X$ with $\beta\geq1$, there will \emph{always} exist a finite $L_1>0$ such that the function is also $(1,L_1,\mathcal X)$-H\"older (\ie, $L_1$-Lipschitz). Thus, assuming Lipschitzness in the second part of \cref{assump: smoothness} is actually \emph{not necessary} for characterizing the regret rate of our algorithm for any single, fixed instance, if we assume a compact $\mathcal X$ as we do below in \cref{assump: support}. However, from the perspective of characterizing the \emph{minimax} regret, where we take a supremum over instances, it is necessary, as the Lipschitz constant $L_1$ may be arbitrarily large in the $(\beta,L,\mathcal X)$-H\"older class of functions.

\subsection{Optimal Decision Region Regularity}

We next introduce a regularity condition on the context regions where each arm is optimal, namely,
$$
\mathcal{Q}_a=\{x \in \mathcal{X}: a\tau(x) \ge 0\}.
$$
When the expected rewards are not restricted parametrically as we imposed in the above, we \emph{must} use local information to estimate them since extrapolation is limited. In particular, in order to estimate $\eta_a(x)$ consistently at a given point $x$, we must have that the contexts of our data on outcomes from arm $a$ eventually become dense around the point $x$. To formalize this notion, we introduce the $(c_0,r_0)$-regularity condition:
\mybigskip
\begin{definition}[$(c_0,r_0)$-regularity Condition]\label{def: regularity}
    A Lebesgue-measurable set $\mathcal{S} \subseteq\mathbb{R}^d$ is called weakly $(c_0, r)$-regular at point $x\in \mathcal{S}$ if
    \[
    \leb[\mathcal{S} \cap \mathcal{B}(x,r)]\ge c_0\leb[\mathcal{B}(x,r)].
    \]
    If this condition holds for all $0 \le r \le r_0$, then set $\mathcal{S}$ is called strongly $(c_0, r_0)$-regular at $x$. Furthermore, if $\mathcal{S}$ is strongly $(c_0, r_0)$-regular at all $x \in \mathcal{S}$, then the set $\mathcal{S}$ is called a $(c_0, r_0)$-regular set.
\end{definition}
Essentially, \edit{if our data for arm $a$ became dense in the set $\mathcal S$ and if $\mathcal S$ were strongly $(c_0, r_0)$-regular at $x$, then sufficient data would be available within any small neighborhood around $x$ to estimate $\eta_a(x)$ well}. If $\mathcal S$ were not regular then, even if our data became dense in $\mathcal S$, there would be diminishing amounts of data available as we looked closer and closer near $x$. For example, the $\ell_q$ unit ball is regular for $q\geq1$ and irregular for $q<1$ because the points at its corners are too isolated from the rest of the set.

\begin{figure}[t!]\centering
~~~\begin{subfigure}[t]{0.4\textwidth}\includegraphics[width=\textwidth]{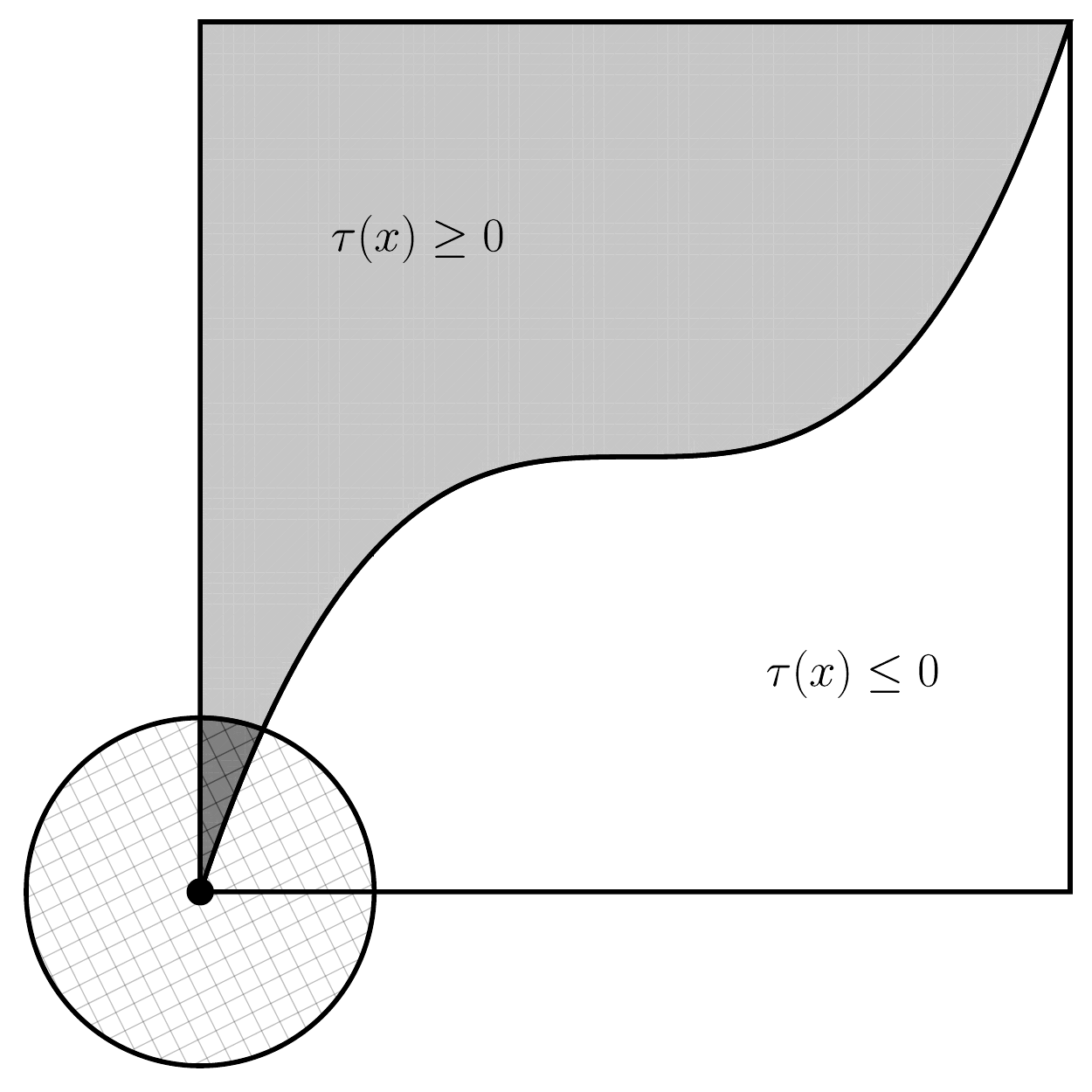}\caption{\cref{assump: decision-set} is satisfied: every ball centered in $\mathcal Q_a$ has at least $c_0=1/12$ of its volume intersecting $\mathcal Q_a$.}\label{fig: c0r0 good}\end{subfigure}%
\hfill\begin{subfigure}[t]{0.4\textwidth}\includegraphics[width=\textwidth]{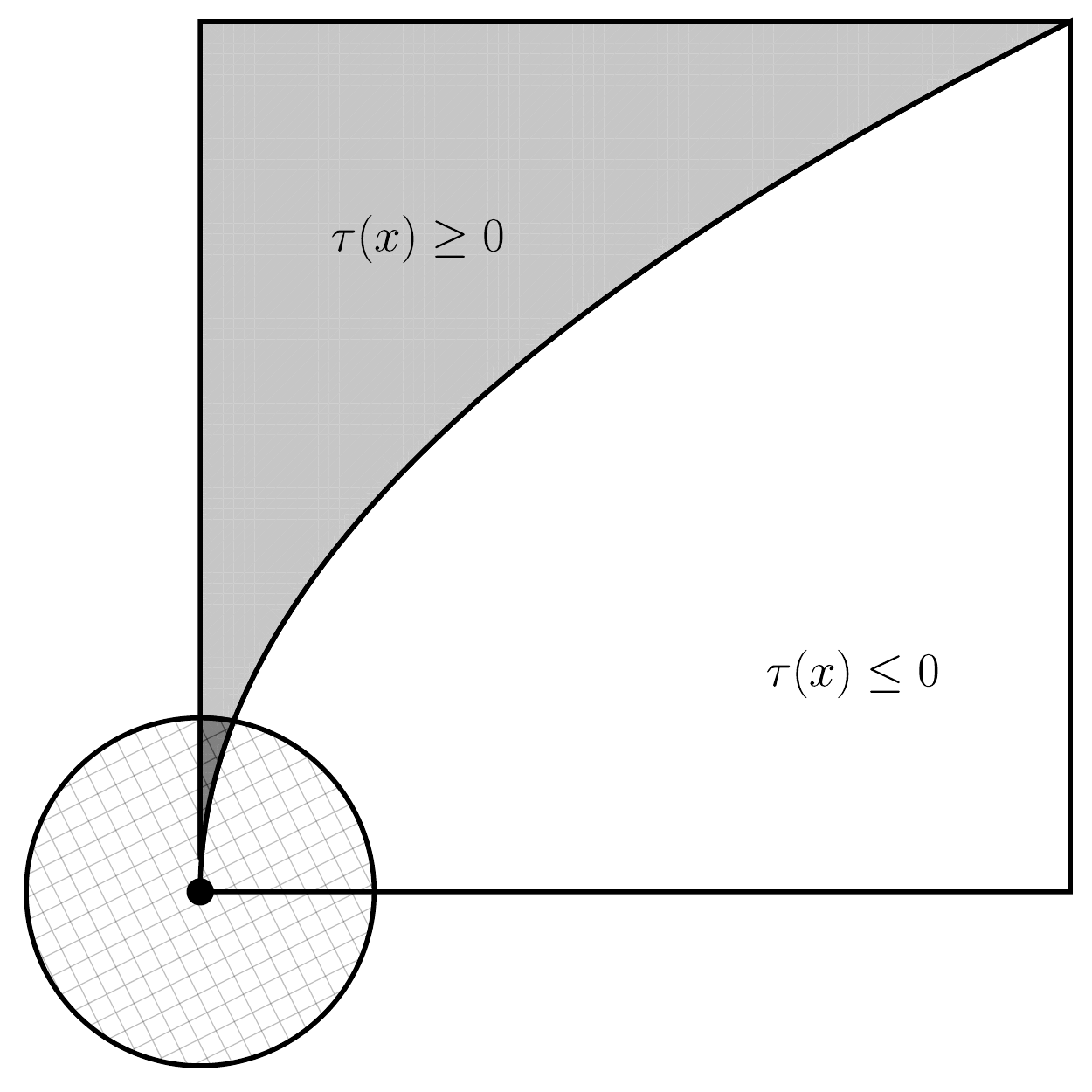}\caption{\cref{assump: decision-set} is violated: smaller balls centered at the corner have a vanishing fraction of their volume intersecting $\mathcal Q_a$.}\label{fig: c0r0 bad}\end{subfigure}~~~%
\vspace{\baselineskip}\caption{Illustration of \cref{assump: decision-set}: each optimal decision region must be regular in that the neighborhood of every point in the region must contain at least some constant fraction of the ball.}\label{fig: c0r0 illustration}%
\end{figure}

Naturally, we need enough data from arm $a$ around $x$ to estimate $\eta_a(x)$ accurately. Luckily, we need only worry about high-accuracy estimation for \emph{both} arms near the decision boundary, where it is hard to tell which of the arms is optimal. (Intuitively, away from the boundary, it is very easy to separate the arms with very few samples, as in the classic MAB case of \citealp{Lai:1985:AEA:2609660.2609757}.) But, we cannot rely on having enough data from arm $a$ in a whole ball around every point near the boundary, as that would require us to pull arm $a$ too often across the boundary, in $\mathcal Q_{-a}$, where it is not optimal. This would necessarily lead to high regret. Instead, we must be able to rely mostly on data from arm-$a$ pulls in $\mathcal Q_a$. Therefore, we must have that this set is regular. If, otherwise, there existed such a point $x\in\mathcal Q_a$ that is sufficiently isolated from the rest of $\mathcal Q_a$ then we cannot generate enough samples to learn $\eta_a(x)$ accurately enough without necessarily incurring high regret.
\mybigskip
\begin{assumption}[Optimal Decision Regions]\label{assump: decision-set}
For $a = \pm 1$, $\mathcal{Q}_a$ is a non-empty $(c_0,r_0)$-regular set.
\end{assumption}

An illustration of this condition is given in \cref{fig: c0r0 illustration}. We note that this condition is a refinement of the usual condition for nonparametric estimation, which simply requires the support $\mathcal X$ to be a regular set \citep{Tsybakov:2008:INE:1522486}. This refinement is necessary for the unique bandit setting we consider where we must worry about the costs of adaptive data collection and may not simply assume a good dataset is given.
Since the intersection of regular sets may not always be regular, it is insufficient to only assume the support $\mathcal{X}$ is regular and expected rewards are smooth in order to guarantee \cref{assump: decision-set}, as seen in \cref{fig: c0r0 bad}.

\label{assump 2 necessary discussion}\edit{\cref{assump: decision-set} is \textit{necessary} to guarantee the optimal minimax regret regime we study, bridging the previous non-differentiable and parametric regimes. In particular, we will show that for \emph{any} policy, there exist problem instances satisfying all assumption except \cref{assump: decision-set} for which the regret rate is \emph{higher} than the minimax regret rate for instances satisfying all assumptions (see \cref{lb2}).}

\subsection{Bounded Covariate Density}
While \cref{assump: decision-set} ensures there is sufficient volume around each point $x$ where we need to estimate $\eta_a(x)$, we also need to ensure that this translates to being able to collect sufficient data around each such point.
Toward this end, we make the assumption that the contexts have a density and that it is bounded away from zero and infinity.
\mybigskip

\begin{assumption}[Strong Density]\label{assump: support}
    The marginal distribution of $X$
    has density $\mu(x)$ with respect to Lebesgue measure and $\mu$ is bounded away from zero and infinity on its support $\mathcal X$:
    \[
    0<\mu_{\min}\le \mu(x) \le \mu_{\max}<\infty,\quad\forall x\in \mathcal{X}.
    \]
    Moreover, its support $\mathcal{X}$ is compact
    and
    $\mathcal{X} \subseteq [0, 1]^d$.
\end{assumption}
Note that restricting $\mathcal X$ to $[0,1]^d$ is without loss of generality, having assumed compactness. Scaling and shifting the covariates to be in $[0,1]$ will only affect the constants $L,\,L_1$ in \cref{assump: smoothness}.

Together, \cref{assump: decision-set,assump: support} imply a lower bound on the probability that each arm is optimal.
\mybigskip
\begin{lemma}\label{lemma: nomempty-opt}
Under \cref{assump: decision-set,assump: support}, we have
$\pr(X\in\mathcal{Q}_{a}) \ge p$ for $a=\pm1$, where
\[
p=\mu_{\min}c_0r_0^dv_d.
\]
\end{lemma}

\subsection{Margin Condition}

We further impose a margin condition commonly used in stochastic contextual bandits \citep{Rigollet10,goldenshluger2013} and classification \citep{mammen1999smooth,tsybakov2004optimal}, which determines how the estimation error of expected rewards translates into regret of decision-making.
\mybigskip
\begin{assumption}[Margin Condition]\label{assump: margin} The conditional average treatment effect function $\tau$ satisfies the  margin condition with parameters $\alpha\geq0$ and $\MarginL$:
	$$\mathbb{P}(0<|\tau(X)|\le t)\le \MarginL \; t^{\alpha}\quad\forall t>0.$$
\end{assumption}

The margin condition quantifies the concentration of contexts very near the decision boundary, where the optimal action transitions from one arm to the other. This measures the difficulty of determining which of the two arms is optimal.
When $\alpha$ is very small, the CATE function can be arbitrarily close to $0$ with high probability, so even very small estimation error of the CATE function may lead to suboptimal decisions.
In contrast, when $\alpha$ is very large, \edit{the probability that expected arm rewards are very close to one another but not equal is very low, or, in other words, the expected rewards for two arms are nicely separated on most of $\mathcal X$.}

\subsection{Minimax Regret}\label{sec:minimaxreget}

Having now defined the problem and our assumptions about the distribution $\mathbb P$ defining the problem instance, we can introduce the notion of \emph{minimax regret}. The minimax regret is the minimum over admissible policies $\pi$ of the maximum of the regret of $\pi$ over all problem instances $\mathbb P$
that fit our assumptions.
This describes the best-achievable behavior in the problem class we consider.

Formally, for $\beta\geq1$,
we let $\mathcal P(\beta,L_1,L,c_0,r_0,\mu_{\min},\mu_{\max},\MarginL,\alpha)$ be the set of all distributions $\mathbb P$ on $(X,Y(-1),Y(+1))\in[0,1]^d\times\mathbb R\times\mathbb R$ that satisfy \cref{assump: smoothness,assump: decision-set,assump: support,assump: margin} with these parameters.
For brevity, we write $\mathcal P$, implicitly considering the parameters as fixed.
Letting $\Pi$ denote all admissible policies, for some fixed parameters specifying a class $\mathcal P$, we then define the minimax regret as
$$
\mathcal R_T=\inf_{\pi\in\Pi}\sup_{\mathbb P\in\mathcal P}R_T(\pi).
$$
The minimax regret exactly characterizes how well we can hope to do in the given class of instances.
\edit{
	It can be thought of as a game against nature where nature plays second, after we choose a policy, but we know the set of plays available to nature (i.e., the instance class $\mathcal P$ with given parameters).
    Restricting the class is crucial for characterizing the dependence of regret on smoothness since the minimax regret against a single instance is always $0$ and the minimax regret against the class of instances with arbitrary $\beta$ is linear in $T$.
    The minimax regret therefore characterizes the best-achievable regret if one were \emph{only} told the smoothness parameter (and additional parameters above) but the instance might be adversarially bad in every other way.
}

\edit{Now we describe a general strategy for computing the minimax regret rate, which we will follow in this paper.} Suppose that, on the one hand,
 we can find a function $f(T)$ and an admissible policy $\hat\pi$ such that its regret for every instance $\mathbb P\in\mathcal P$ is bounded by the \emph{same} function, $R_T(\hat\pi)\leq f(T)$. Next, suppose that, on the other hand, we can show that there exists a function $f'(T)=\tilde\Omega(f(T))$ where for every admissible policy $\pi'$ there exists an instance $\mathbb P\in\mathcal P$ such that the regret is lower bounded by this \emph{same} function, $R_T(\pi')\geq f'(T)$. Then we will have shown two critical results:
(a) the minimax regret satisfies the rate $\mathcal R_T=\tilde\Theta(f(T))$ and (b) we have a specific algorithm $\hat\pi$ that can actually achieve this best-possible worst-case regret in rate, which also means the regret of $\hat\pi$ is known to be bounded in this rate for every single instance encountered.

In this paper, we will proceed exactly as in the above. First, in \cref{section: algorithm}, we will develop a novel algorithm that can adapt to every smoothness level. Then, in \cref{section: ub} we will prove a bound on its regret in every instance. Since this bound will depend only on the parameters of $\mathcal P$, we will have in fact established an upper bound on the minimax regret as above. In \cref{section: lower bound} we will find a bad instance for every policy that yields a matching (up to polylogs) lower bound on its regret, establishing a lower bound on the minimax regret. This will exactly yield the desired conclusion: a characterization of the minimax regret and the construction of a specific algorithm that achieves it.

\subsection{On the Relationship of Margin and Smoothness}\label{sec: margin-smoothness}

{\blockedit%
Before proceeding to develop a bandit algorithm for the smooth bandits problem and characterizing the minimax regret, we comment on the relationship between the smoothness of expected rewards and the margin assumption.
\cref{assump: smoothness} implies that the CATE function $\tau(x)$ is a member of the $(\beta,2L,\mathcal{X})$-H\"older class with $\beta\geq1$. Intuitively, when $\tau(x)$ is smooth, it cannot change too abruptly at the decision boundary $\tau(x) = 0$, so, if it either touches or crosses the decision boundary at all, the mass near it must be significant (small $\alpha$).

First, we present a direct corollary of Proposition 3.4 of \cite{Audibert05}, who study (offline) classification with a smooth conditional probability function.
\begin{proposition}\label{prop: margin no cross}
Suppose \cref{assump: smoothness,assump: support,assump: decision-set,assump: margin} hold with $\alpha>1$. Then for all $x\in\operatorname{interior}(\mathcal X)$ there exists $r>0,\,\sigma\in\{-1,1\}$ such that $\sigma\tau(x')\geq0$ for all $\|x'-x\|\leq r$.
\end{proposition}
Recall \cref{assump: smoothness,assump: support,assump: decision-set,assump: margin} specify the class of the bandit problem we consider, so \cref{prop: margin no cross} is a statement about the instances in this class.
\Cref{prop: margin no cross} shows that for a smooth bandit problem when $\alpha>1$, all interior points have a neighborhood where $\tau(x)$ is only nonnegative or only nonpositive, meaning $\tau(x)$ does not cross 0.
Notice that by continuity of $\tau$, this also implies that if any $x$ and $x'$ are in the same connected component of the interior (i.e., are connected by a interior path) then $\tau(x)\tau(x')\geq0$, so that there must exist an optimal policy $\pi^*(x)$ in \cref{eq: oracle} that is constant on connected components of the interior. However, $\tau(x)$ might still be arbitrarily close to zero, especially as we vary the instance in the class of instances $\mathcal P$ to compute the minimax regret, potentially making it difficult to distinguish the optimal arm and still requiring non-trivial regret. 

We next show this, however, does not happen when the margin is \emph{very} large.
\begin{proposition}\label{prop: margin sharp}
Suppose \cref{assump: smoothness,assump: support,assump: decision-set,assump: margin} hold with $\alpha>d$. Then there exists a positive constant $\tau_{\min}$ depending only on the parameters of \cref{assump: smoothness,assump: support,assump: decision-set,assump: margin} such that for any $x\in \mathcal{X}$, we have either $\tau(x) = 0$ or $\abs{\tau(x)}\geq\tau_{\min}$.
\end{proposition}
By continuity of $\tau(x)$, \cref{prop: margin sharp} implies that, on each connected component of $\mathcal X$, $\tau(x)$ has a constant sign (negative, zero, or positive). In particular, as it would contradict \cref{lemma: nomempty-opt}, \cref{prop: margin sharp} implies that there exist no smooth bandit instances with $\alpha>d$, $\mathcal X$ connected, and $\tau(x)$ not the constant zero function on $\mathcal X$, as such would require $\pr(X\in\mathcal Q_1)\wedge \pr(X\in\mathcal Q_{-1})=0$.

More crucially, \cref{prop: margin sharp} makes an implication on minimax regret when $\alpha>d$ since $\tau_{\min}$ is a \emph{uniform} bound (and in this sense the result is stronger than the statement corresponding to $\alpha>d$ in Proposition 3.4 of \citealp{Audibert05}).
Notice that $\abs{\tau(X)}\in\{0\}\cup[\tau_{\min},\infty)$ implies that \cref{assump: margin} holds for \emph{any} $\alpha\geq0$ (simply let $\gamma=\tau_{\min}^{-\alpha}$).
Recall that the class of instances $\mathcal P$ in \cref{sec:minimaxreget} is defined in terms of the parameters of \cref{assump: smoothness,assump: support,assump: decision-set,assump: margin}. Therefore, \cref{prop: margin sharp} shows that for any $\alpha'\geq\alpha>d$ (and $\gamma\geq\tau_{\min}^{-\alpha'}$ is sufficiently large), the minimax regret in the class of instances $\mathcal P$ is upper bounded by the minimax regret in the class $\mathcal P'$ where we set $\alpha$ to the \emph{larger} $\alpha'$.
More to the point, in the following, by exhibiting a feasible algorithm, we establish an upper bound on minimax regret of $\tilde O(1)$ whenever $\alpha\geq1+d/\beta$. \Cref{prop: margin sharp} shows that the same $\tilde O(1)$ bound applies even if just $\alpha>d\wedge (1+d/\beta)$.}

\section{\textsc{SmoothBandit}: A Low-Regret Algorithm for Any Smoothness Level} \label{section: algorithm}

In this section, we develop our algorithm, \textsc{SmoothBandit} (\cref{alg}). We first review local polynomial regression, which we use in our algorithm to estimate $\eta_a$.

\subsection{The Local Polynomial Regression Estimator}

A standard result of (offline) nonparametric regression is that the smoother a function is in terms of its H\"older parameter $\beta$, the faster it can be estimated. Appropriate convergence rates can, \eg, be achieved using local polynomial regression estimators that adjusts to different smoothness levels \citep{stone1982,stone1980}. In this section, we briefly review local polynomial regression and its statistical property in an \textit{offline} bandit setting. Its use in our \textit{online} algorithm is described in \cref{sec: alg-description}. More details about local polynomial regression can be found in \citet{Tsybakov:2008:INE:1522486,audibert2007}.

Consider an offline setting, where we have access to an exogenously collected i.i.d. sample, $S = \{(X_t, Y_t)\}_{t=1}^{n}$ drawn i.i.d. from $(X,Y)$, where $X$ has support $\mathcal X\subset\mathbb R^d$.
We can then estimate the regression function $\eta(x)=\Eb{Y\mid X=x}$ at every point $x$ using the
the following local polynomial estimator.
\mybigskip
\begin{definition} [Local Polynomial Regression Estimator]\label{def: local-polynomial}
	For any $x \in \mathcal{X}$, given a bandwidth $h > 0$, an integer $l \ge 0$,
    samples $S = \{(X_t, Y_t)\}_{t=1}^{n}$,
    and a degree-$l$ polynomial model ${\theta}(u; x, \vartheta, l) = \sum_{|r| \le l} {\vartheta}_r(S) (u - x)^r$,
    define the local polynomial estimate for $\eta(x)$ as \edit{
    $\hat{\eta}^{\text{LP}}(x; S,h, l) = {\theta}(x; x, \hat\vartheta_{x}, l)$}, where
	\begin{align}\label{eq: least-squares}\textstyle
	  \hat\vartheta_{x}\in\arg\min_\vartheta
	  \sum_{t: X_t \in \mathcal{B}(x, h)} \left(Y_t - {\theta}(X_t; x, \vartheta, l)\right)^2.
		\end{align}
  For concreteness, we define $\hat{\eta}^{\text{LP}}(x; S,h, l)=0$ if the minimizer is not unique.
\end{definition}

In words, the local polynomial regression estimator fits a polynomial by least squares to the data that is in the $h$-neighborhood of the query point $x$ and evaluates this fit at $x$ to predict $\eta(x)$.

Since \cref{eq: least-squares} is a least squares problem, the estimation accuracy of the local polynomial estimator $\hat{\eta}^{\text{LP}}(x; S,h, l)$ depends on the associated Gram matrix:
\begin{equation}\textstyle\label{eq: gram matrix}
\hat{\mathcal{A}}(x; S, h, l) = \{\hat{\mathcal{A}}_{r_1, r_2}(x; S, h)\}_{|r_1|, |r_2| \le l}, ~~ \text{where} ~~ \hat{\mathcal{A}}_{r_1, r_2}(x; S, h) = \sum_{t: X_t \in \mathcal{B}(x, h) } (\frac{X_t - x}{h})^{r_1 + r_2}.
\end{equation}
The following proposition illustrates (using the offline setting as an example) why our \cref{assump: smoothness,assump: support,assump: decision-set} are crucial in our problem. In particular, it shows that bounded density and regularity of the support of the data ensure a well-conditioned locally weighted Gram matrix.
Moreover, it shows how the bandwidth and polynomial degree should adapt to the smoothness level $\beta$. This proposition is a direct extension of Theorem 3.2 of \citet{audibert2007}. We include this result purely for motivation, while in our \emph{online} setting we will need to establish a more refined result that accounts for our adaptive data collection.
\mybigskip

\begin{proposition}\label{prop: ivertibility}

	Let $S$ be an i.i.d. sample of $(X,Y)$, where $\eta$ is $(\beta,L,\mathcal X)$-H\"older, $\mathcal X$ is compact and $(c_0,r_0)$-regular, and $X$ has a density bounded away from 0 and infinity on $\mathcal X$. Then there exist positive constants $\lambda_0,\, C_1,\, C_2$ such that, for any $x\in \mathcal X$, and $\epsilon > 0$, with probability at least $1 - C_1\exp\big\{-C_2n_{a}^{\frac{2\beta}{2\beta + d}}\varepsilon^2\big\}$, we have
	\[\textstyle
		\lambda_{\min}(\hat{\mathcal{A}}(x; S, n^{-\nicefrac{1}{(2\beta + d)}}, \floor*{\beta})) \ge \lambda_0, ~~~\text{and}~~~ \big\vert \hat{\eta}^{\text{LP}}(x; S, n^{-\nicefrac{1}{(2\beta + d)}}, \floor*{\beta}) - \eta(x)\big\vert \le \varepsilon .
	\]
\end{proposition}
In our online bandit setting, the samples for each arm are collected in an adaptive way, since both exploration and exploitation can depend on data already collected.
As a result, the distribution of the samples for each arm is considerably more complicated.
Thus, we will need to use the local polynomial estimator in a somewhat more sophisticated way and analyze it more carefully. See \cref{sec: alg-description,section: ub} for the details.

\subsection{Our Algorithm}\label{sec: alg-description}
\begin{algorithm}[t!]
    \caption{\textsc{SmoothBandit}}
    \label{alg}
    \begin{algorithmic}[1]
    \STATEx\textbf{Input:} Grid lattice $G$, epoch schedule $\{\mathcal{T}_{k}\}_{k = 1}^K$, H\"older smoothness constant $\beta$, regularity constant $c_0$, context dimension $d$.
    \STATE{Initialize $\mathcal{E}_{\pm 1,1} = \emptyset,\,\mathcal{R}_1 = \mathcal X$ (\edit{exploit nowhere, explore everywhere}).}
    \FOR{$t \in \mathcal{T}_1$}
                \STATE Pull $A_t = \pm 1$ randomly, equiprobably
    \ENDFOR
    \STATE {Log the samples $S_{\pm 1, 1} = \{(X_t, Y_t): t \in \mathcal{T}_1, A_t = \pm 1\}$}
    \FOR {$k = 2, 3, \dots, K$}
        \STATE Identify inestimable regions for local polynomial regression with bandwidth $H_{a , k-1}$ ($a = \pm 1$):
            \begin{align}
                \mathcal{D}_{a, k - 1} &= \bigcup\braces{\cube(x) :
                x \in \mathcal{R}_{k - 1} \cap {G} ,\begin{array}{l}\prns{\bigcup_{j = 1}^{k - 1}\exploit_{a, j} \cup \mathcal{R}_{k - 1}} \cap \mathcal{X}\text{ is not}\\\text{weakly $\prns{\frac{c_0}{2^d}, H_{a, k - 1}}$-regular at $x$}\end{array}}
                \label{eq: deletion-rule}
            \end{align}
        \STATE Set $N_{\pm1, k -1} = \abs{S_{\pm1, k - 1}},\,H_{\pm1, k - 1} = N_{\pm1, k - 1}^{-\nicefrac{1}{(2\beta + d)}}$
        \STATE Construct the CATE estimate for every $x \in G\cap\rand_{k - 1} \cap \mathcal{D}_{1, k - 1}^C \cap \mathcal{D}_{-1, k - 1}^C$
        \begin{equation}\label{eq: alg cate estimate}\hat{\tau}_{k-1}(x)=\hat{\eta}^{\text{LP}}(x; S_{+1,k-1},H_{+1,k-1}, \floor*{\beta})-\hat{\eta}^{\text{LP}}(x; S_{-1,k-1},H_{-1,k-1}, \floor*{\beta})
        \end{equation}
        \STATE Update decision regions: for $a = \pm 1$,
                    \begin{align}\textstyle
                    \mathcal{E}_{a,k}
                        &=
                        \bigcup\braces{
                        \cube(x):
                        x\in G\cap\mathcal{R}_{k-1} \cap \mathcal{D}_{1, k - 1}^C \cap \mathcal{D}_{-1, k - 1}^C,\,
                        a\hat{\tau}_{k-1}(x)> \epsilon_{k-1}
                        }
                        \cup \mathcal{D}_{-a, k - 1}, \label{eq: exploit} \\
                    \mathcal{R}_k
                        &=
                        \bigcup\braces{
                        \cube(x):
                        x\in G\cap\mathcal{R}_{k-1} \cap \mathcal{D}_{1, k - 1}^C \cap \mathcal{D}_{-1, k - 1}^C,\,
                        \abs{\hat{\tau}_{k-1}(x)}\leq \epsilon_{k-1}
                        }
                        \label{eq: rand}.
                    \end{align}
            \FOR{$t \in \mathcal{T}_k$}
                \STATE \textbf{if } $X_t \in \bigcup_{j = 1}^k \mathcal{E}_{+ 1, j}$ \textbf{ then } {pull $A_t = + 1$}
                \STATE \textbf{else if } $X_t \in \bigcup_{j = 1}^k \mathcal{E}_{- 1, j}$ \textbf{ then } {pull $A_t = - 1$}
                \STATE \textbf{else } pull $A_t = \pm 1$ randomly, equiprobably
            \ENDFOR
        \STATE Log samples $S_{\pm1, k} = \{(X_t, Y_t): t \in \mathcal{T}_k, A_t = \pm1\}$
    \ENDFOR
    \end{algorithmic}
\end{algorithm}
In this section we present our new algorithm for smooth contextual bandits, which uses local polynomial regression estimators that adjust to any smoothness level. The algorithm is summarized in \cref{alg}. Below we review its salient features. In what follows we assume a fixed horizon $T$, but can accommodate an unknown, variable $T$ using the well-known doubling trick (see \citealp{Auer1995GamblingIA}; \citealp[p.~17]{cesa2006prediction}).

{\blockedit\subsubsection{Algorithm Overview}\label{sec: algo overview}
We begin with a rough sketch of the overall structure of the algorithm. Specifics are given in \cref{alg} and in the sections below. Our algorithm makes a cover $\mathcal C$ of the covariate support using a grid of hypercubes, $\mathcal X=\bigcup_{S\in\mathcal C}S$, where $\mathcal C$ consists of the intersections of $\mathcal X$ with disjoint half-open-half-closed hypercubes with a finely tuned side length (see \cref{sec: grid}).
Our algorithm then proceeds in epochs of (roughly) geometrically increasing time lengths (see \cref{sec: epoch}). At the beginning of the $k$th epoch, each cube $S\in\mathcal C$ in the cover is assigned either to the \emph{random} exploration region, $\mathcal R_k$, or to one of two \emph{exploitation} regions, $\mathcal E_{+1,k},\mathcal E_{-1,k}$ (see \cref{sec: algo regions}). During the $k$th epoch, if $X_t$ falls in $\mathcal R_k$ we pull arms $\pm1$ with probability $1/2$ each and if $X_t$ falls in $\mathcal E_{a,k}$ we pull arm $a$. We start out with randomizing everywhere, $\mathcal R_0=\mathcal X$. Then, as we collect more data we peel hypercubes away from the randomization region and into the exploration region, where we declare one of the two arms is almost certainly optimal based on observations from the epoch that just concluded. There are two ways to infer that a hypercube should be moved to an exploitation region: either we have already declared one of the arms is almost certainly optimal in very many nearby hypercubes (see \cref{sec: algo cleaning}) or we have enough data near the center of the hypercube $x_0$ to fit a high-fidelity local polynomial regression estimate for both $\eta_{+1}(x_0),\eta_{-1}(x_0)$ (this involves data \emph{outside} the hypercube) and the difference is large enough to rule out (with high probability) that one arm appears better due only to estimation noise so we declare the apparently dominant arm is indeed dominant (see \cref{sec: algo cate est}). Thus, we maintain a plan of action of how we will act in each round $t$ depending on the observed context $X_t$, and at the end of each epoch we update this plan by declaring more context regions as exploitation regions where we only pull one of the two arms.
This structure is mimicked by our multi-arm extension where we maintain an active set of arms (subset of $\mathcal A$) for each hypercube (see \cref{section: k arm}).
}

\subsubsection{Grid Structure}\label{sec: grid}
\begin{figure}[t!]\centering%
\includegraphics[width=0.85\textwidth]{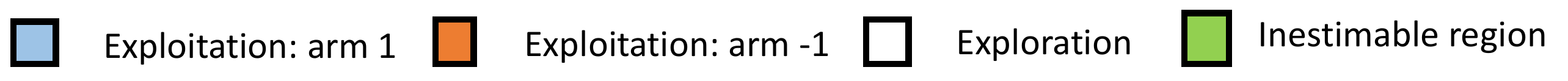}\\[1em]%
\begin{subfigure}[t]{0.3\textwidth}\includegraphics[width=\textwidth]{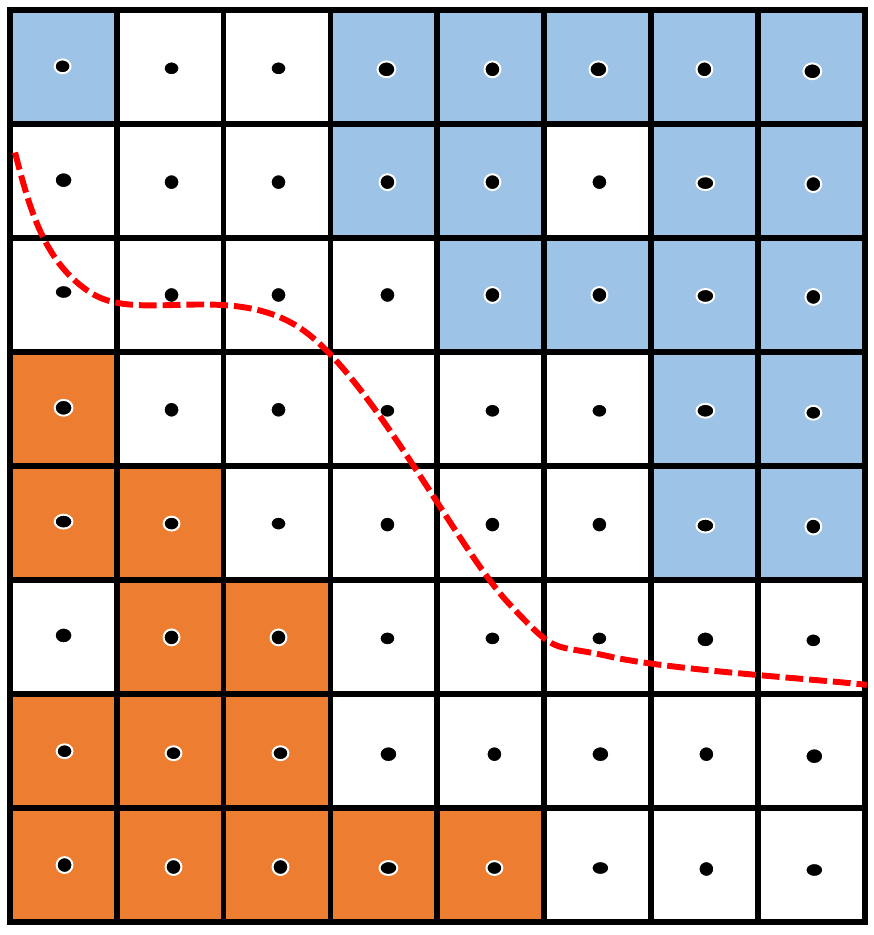}\caption{The decision regions at the end of $k - 1^{\text{th}}$ epoch: each hypercube is assigned with one action, either always pull arm $1$, always pull arm $-1$, or pull one of the arms at random equiprobably.}\label{fig: algorithm a}\end{subfigure}%
\hfill\begin{subfigure}[t]{0.3\textwidth}\includegraphics[width=\textwidth]{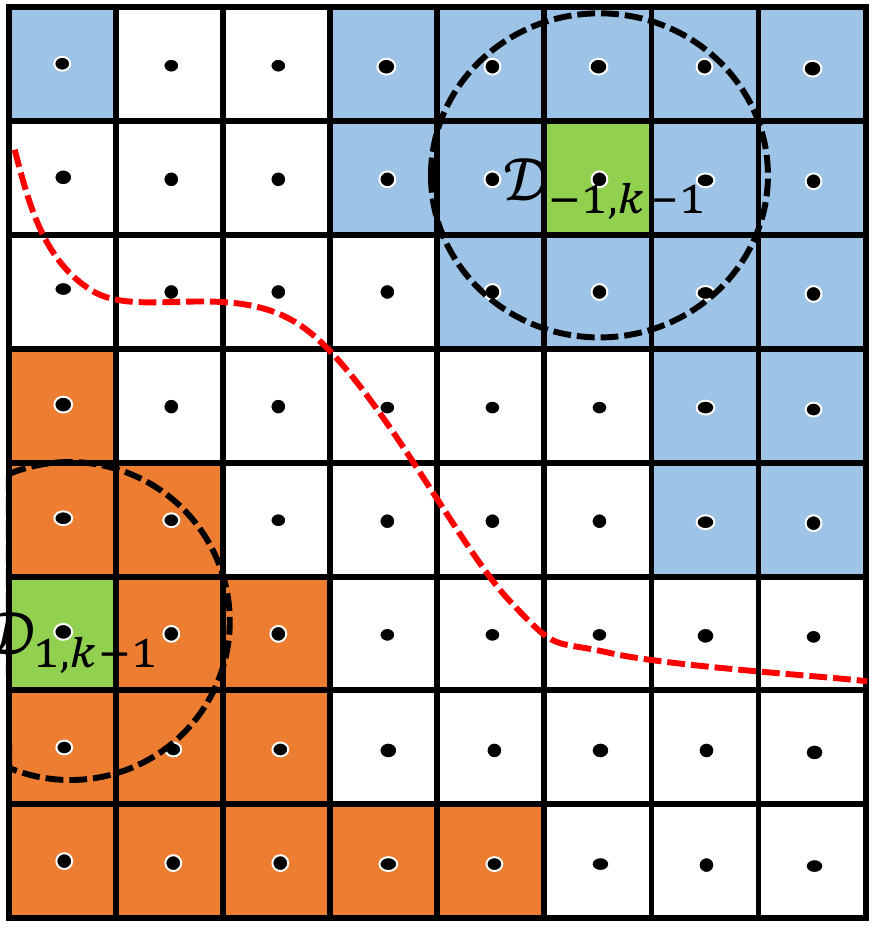}\caption{In epoch $k$, we cannot sufficiently accurately estimate $\eta_a$ in $\de_{a, k - 1}$ (green hypercubes) for lack of sufficient samples from arm $a$ in the local neighborhoods (black dashed circle).}\label{fig: algorithm b}\end{subfigure}%
\hfill\begin{subfigure}[t]{0.3\textwidth}\includegraphics[width=\textwidth]{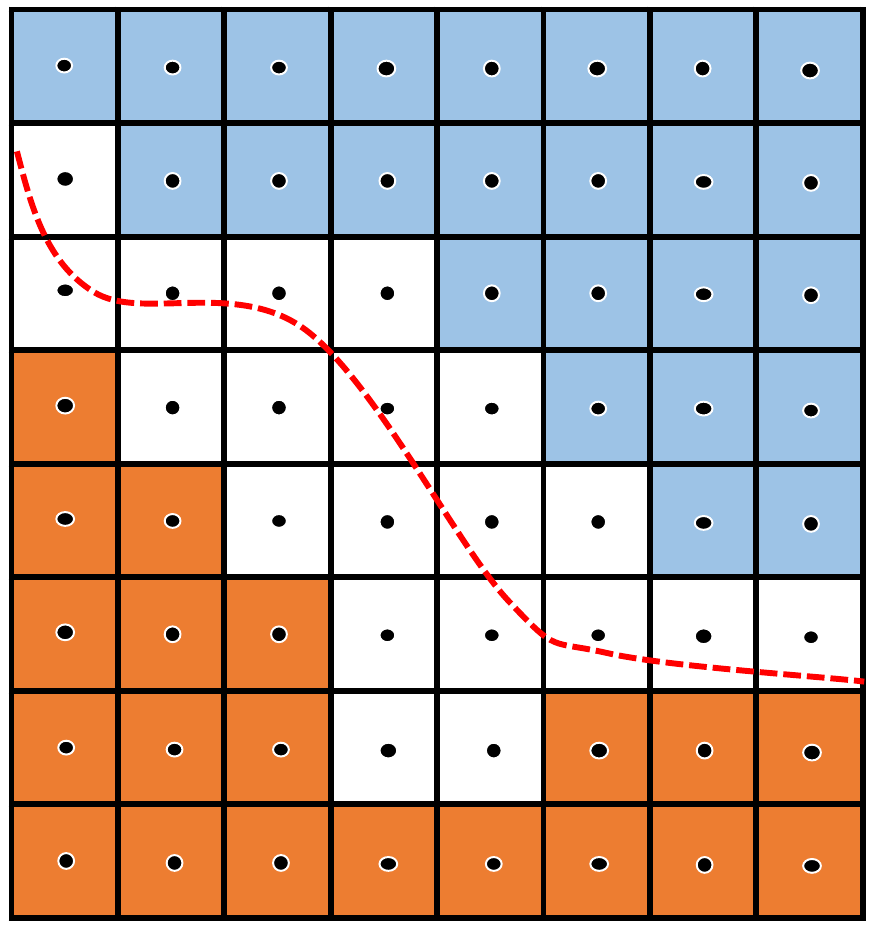}\caption{The decision regions at the end of $k^{\text{th}}$ epoch: previously-exploration regions are moved to exploitation either due to large estimated CATE where estimable or due to inestimability.
}\label{fig: algorithm c}\end{subfigure}%
\vspace{\baselineskip}\caption{Updating the decision regions. The black dots represent hypercube centers given in $G$. The red dashed curve is the true decision boundary.}
\label{fig: algorithm}
\end{figure}

Following \citep{stone1982} and similarly to previous nonparametric-response bandit literature \citep{Rigollet10,perchet2013}, we partition the context space into small hypercubes. For each time step, both our estimates of $\eta_a(x)$ and our policy $\pi_t(x)$ will be piecewise constant on these hypercubes.
Specifically, in each hypercube, we will either pull arm $+1$, pull arm $-1$, or equiprobably pull a random arm from among the two (see \cref{fig: algorithm}(c)).
Crucially, and differently from \citet{Rigollet10,perchet2013}, we use data from both \emph{inside and outside} each hypercube to define the estimates and action inside each hypercube.

We first define a grid lattice $G'$ on $[0,1]^d$: letting $\delta = T^{-\frac{\beta}{2\beta + d}}(\log T)^{-1}$,
\[
G'= \bigg\{\prns{\frac{2j_1 +1}{2}\delta,\dots, \frac{2j_d +1}{2}\delta}\ :\ j_i \in \{0, \dots, \lceil \delta^{-1}\rceil -1\},\, i= 1,\dots, d\bigg\}.
\]
For any $x\in \mathcal{X} \subseteq [0,1]^d$, we denote by $g(x)=\arg\min_{x'\in G'}\magd{x-x'}$ the closest point to $x$ in $G'$.
If there are multiple closest points to $x$, we choose $g(x)$ to be the one closest to $(0, 0, \cdots, 0)$.
All points that share the same closest grid point $g(x)$ belong to a hypercube with length $\delta$ and center $g(x)$. We denote this hypercube as $\cube(x) = \{x'\in \mathcal{X}: g(x') = g(x)\}$, and the collection of all such hypercubes overlapping with the covariate support as
$$
\mathcal{C} = \braces{\cube(x): x\in G}, ~~\text{where}~~ G=\braces{x\in G':\pr(\cube(x) \cap \mathcal{X}) > 0}
.$$
Note that the union of all cubes in $\mathcal C$, $\bigcup_{S\in\mathcal{C}}S$, must cover the covariate support $\mathcal{X} \subseteq [0,1]^d$.

\subsubsection{Epoch Structure}\label{sec: epoch}
Our algorithm then proceeds in an epoch structure, where the estimates and actions assigned to each hypercube is fixed for the duration of that epoch. For each epoch, we target a CATE-estimation error tolerance of $\epsilon_k=2^{-k}$.
With this target in mind, we set
the length of the $k^\text{th}$ epoch as follows:
\begin{align}\label{eq: sample-size}\textstyle
  n_k = \left\lceil \frac{4}{p}(\frac{\log(T\delta^{-d})}{C_0\epsilon_k^2})^{\frac{2\beta +d}{2\beta}}+\frac{2}{p^2}\log T \right\rceil,
\end{align}
where $p,\,C_0$ are positive constants given in \cref{lemma: nomempty-opt,lemma: eigenvalue}, respectively.
 We further denote the time index set associated with the $k^\text{th}$ epoch as $\mathcal{T}_k = \{t: \sum_{i=1}^{k-1} n_{i-1}+1\le t \le \min\{\sum_{i=1}^k n_i,T\} \}$.

In our algorithm, we continually maintain a growing region, composed of hypercubes, where we are near-certain which of the arms is optimal. In these regions we always pull the seemingly optimal arm.
In contrast, we randomize wherever we are not sure (denoted by the region $\mathcal R_k$ for epoch $k$).
The first epoch, $\mathcal T_1$, is a cold-start phase where, lacking any information, we simply pull each arm uniformly at random in every hypercube ($\mathcal R_1=\mathcal X$).
After that point, once we have some data, for each subsequent epoch, $k\geq2$, we add the hypercubes $\exploit_{a,k}\subseteq \mathcal R_{k-1}$ to the set of hypercubes where we just learned that arm $a$ is probably optimal, never removing any hypercube that was before added.
This means that, in epoch $k$, we are collecting data on arm $a$ exclusively
in the region $\bigcup_{j=1}^k\exploit_{a,j}\cup\mathcal R_{k}$.
We describe in detail how we determine which hypercubes, $\exploit_{a,k}$, to add to the exploitation region of each arm in each epoch in \cref{sec: algo cleaning,sec: algo regions}.

 The total number of epochs $K$ is the minimum integer such that $\sum_{k = 1}^K n_k \ge T$.
The following lemma shows that $K$ grows at most logarithimically with $T$ under the epoch schedule in \cref{eq: sample-size}.
\mybigskip
\begin{lemma} \label{lemma: k-bound}
	When $T\ge e^{C_0\vee 1} $,
	\begin{align*}\textstyle
	K &\le \ceil*{\frac{\beta}{(2\beta+d)\log 2}\log(T)}.
	\end{align*}
\end{lemma}

\subsubsection{Estimating CATE}\label{sec: algo cate est}
Next, we describe how we estimate the expected rewards, $\eta_{\pm1}(x)$, and CATE, $\tau(x) = \eta_{1}(x) - \eta_{-1}(x)$, which we use to determine the action we take in each hypercube in each epoch.
In particular, at the start of each $k^\text{th}$ epoch, $k\geq2$, we estimate each arm's expected reward $\eta_a(x)$ using the data for each arm from the last epoch, which we denote by $S_{a,k-1}$ as in \cref{alg}.
Our proposed estimate is the following piece-wise constant modification of the local polynomial regression estimate:
\begin{align}\label{eq: cate-estimator}
\hat\eta_{a,k-1}(x)&=\hat{\eta}^{\text{LP}}(g(x); S_{a, k - 1},H_{a,k-1}, \floor*{\beta}),\qquad\text{where}\\
H_{a, k-1} &= N_{a, k-1}^{-\nicefrac{1}{(2\beta + d)}},\qquad
N_{a, k -1} = \abs{S_{a, k - 1}}.\notag
\end{align}
Note that by construction $\hat\eta_{a,k-1}(x)=\hat\eta_{a,k-1}(x')$ whenever $x$ and $x'$ are belong to the same hypercube $g(x)=g(x')$.
Then our CATE estimate, $\hat\tau_{k-1}(x)$, is simply the difference of these for $a=\pm1$.
Since we only evaluate $\hat\tau_{k-1}(\cdot)$ at hypercube centers $x\in G$ in our \cref{alg},
we simply use $g(x)$ as the argument to the local polynomial estimates in \cref{eq: alg cate estimate}.
In particular, we only need to compute two local polynomial regression estimates at a subset of the (finitely-many) grid points. Note that some grid points may not even belong to $\mathcal X$ because their hypercubes may not be fully contained in $\mathcal X$; nevertheless, we can use these centers as representative as their $H_{\pm1,k-1}$ neighborhood will still contain sufficient data (\cref{lemma: h-size}).
Note also that the associated sample sizes, $N_{\pm 1, k -1}$, are random variables since they depend on how many samples in the $(k - 1)^\text{th}$ epoch fell in different decision regions and on the random decision regions themselves.

Similar to the non-differentiable bandit of \citet{Rigollet10}, our estimators, \cref{eq: cate-estimator}, are hypercube-wise constant.
That the estimate at the center of each hypercube is a good estimate for the whole hypercube is justified by the smoothness of $\eta_{\pm 1}$ and the error is controlled by the size of the hypercubes (see \cref{lemma: convergence} for details).

However, differently from \citet{Rigollet10}, our estimate at the center of each hypercube uses data from both \emph{inside and outside} the hypercube, instead of only inside. This is established by the next lemma. \edit{(Recall that our \cref{assump: smoothness} requires $\beta\geq1$.)}\label{new comment on beta geq 1}
\mybigskip
\begin{lemma}\label{lemma: h-size}
    There exists a positive constant $c_1$ such that
    \[
    \frac{H_{\pm1, k} }{\sqrt{d}\delta}  \ge c_1 T^{\beta - 1}\log(T)^{\frac{(2\beta - 1)(2\beta + d)}{2\beta}}.
    \]
    When $\beta \ge 1$, there exists $T_0 > 0$ such that for $T > T_0$, $H_{\pm 1, k} \ge \sqrt{d}\delta$ for $1 \le k \le K$.
\end{lemma}
\cref{lemma: h-size} shows that the bandwidth we use, \ie, the neighborhood of data used to construct the estimate, is \emph{much larger} than the hypercube size, where the estimate is used.
According to the nonparametric estimation literature \citep{stone1980,stone1982}, the proposed hypercube size and bandwidths (up to logarithmic factors) are crucial for achieving optimal nonparametric estimation accuracy for smooth functions. This means we indeed \emph{need} to leverage the more global information in order to leverage the {smoothness} of expected reward functions.
This also means that separating the problems into isolated MABs within each hypercube, as would be optimal for unsmooth rewards, is infeasible: we must use data across hypercubes to be efficient and so decisions in different hypercubes will be interdependent. In particular, our actions in one hypercube will affect how many samples we collect to learn rewards in other hypercubes.

\subsubsection{Screening Out Inestimable Regions and Accuracy Guarantees}\label{sec: algo cleaning}

Although using data across multiple hypercubes enables us to improve the estimation accuracy for smooth expected reward functions,
it also introduces complicated dependence between data collection in \emph{one} hypercube and algorithm decisions in \emph{other} hypercubes.
More concretely, the number of samples available to estimate $\eta_a$ in each hypercube, and correspondingly the accuracy of this estimate, depends on the arms we pull in \emph{other}, neighboring hypercubes.
Because in each epoch in each hypercube we either always exploit or randomly explore, this problem arises precisely when there is a hypercube that is surrounded by hypercubes where we are sure about the optimal arm (and therefore did not explore both arms) but in which we are not yet sure about the optimal arm (and therefore need to estimate both arm reward functions). (See \cref{fig: algorithm b,fig: algorithm c}.)
As a result, the local polynomial regression for estimating $\eta_a$ in this hypercube can be ill-conditioned and fail to ensure our accuracy target $\epsilon_k$. Worse yet, this problem will continue to persist at all future epochs because the nearby hypercubes will continue to exploit and the accuracy target will only become sharper.

Luckily, it turns out that whenever such a problem arises, we do not actually need to estimate $\eta_a$ in these hypercubes: the fact that the hypercube is surrounded by neighboring hypercubes where we are sure one arm is optimal means that the same arm is also optimal in this hypercube with high probability (see \cref{lemma: decision-set-characterization}). The only thing we need to do is to detect this issue correctly.
Specifically, we propose to use the rule in \cref{eq: deletion-rule} in order to screen out the inestimable regions.
This screening rule is motivated by \cref{prop: ivertibility,assump: decision-set}, which imply that the regularity property of the support of the samples $S_{a, k - 1}$ (\ie, $(\bigcup_{j = 1}^{k - 1}\exploit_{a, j} \cup \mathcal{R}_{k - 1}) \cap \mathcal{X}$) is critical for the conditioning of the local polynomial estimator.
We show in \cref{lemma: deletion-well-defined} that this screening procedure is well-defined: any hypercube in $\mathcal{C}$ can be classified into at most  one of $\mathcal{D}_{1, k}$ and $\mathcal{D}_{-1, k}$ but not both.
Moreover, although we check only \textit{weak} $(\frac{c}{2^d}, H_{a, k-1})$-regularity with respect to \textit{only} hypercube centers, \cref{lemma: delete-set} implies far stronger consequence for the proposed screening rule: $(\bigcup_{j = 1}^{k - 1}\exploit_{a, j} \cup \mathcal{R}_{k - 1}) \cap \mathcal{X}$ is not \textit{strongly} $(c_0, r_0)$-regular at \textit{any} point in $\mathcal{D}_{a, k - 1}$.

After removing these inestimable regions, we can show (\cref{thm: good-prob}) that the our \emph{uniform} estimation error anywhere in the remaining uncertain region from each epoch (\ie, $\mathcal{R}_{k} \cap D_{1, {k}}^C \cap D^C_{-1, {k}}$) is \emph{exponentially} shrinking:
\begin{align}\label{eq: est-bound}\textstyle
  \sup_{x\in \mathcal{R}_{k} \cap D_{1, {k}}^C \cap D^C_{-1, {k}}}\abs{\hat{\tau}_{k}(x)-\tau(x)} \le \epsilon_{k} ~~ \text{with probability} ~~  1 - {O}({T}^{-1}).
\end{align}

\subsubsection{Decision Regions}\label{sec: algo regions}
We start by randomizing everywhere, $\mathcal R_1=\mathcal X$, and in each subsequent epoch, we remove the hypercubes $\exploit_{-1,k},\exploit_{1,k}$ from the randomization region $\mathcal R_k$ and assign them to join the growing exploitation regions.
The set $\exploit_{a, k}$ is the union of two parts.
The first, $\big\{x\in \mathcal{R}_{k-1} \cap \mathcal{D}_{1, k - 1}^C \cap \mathcal{D}_{-1, k - 1}^C: a\hat{\tau}_{k-1}(x)> \epsilon_{k-1} \big\}$, is determined by $\hat{\tau}_{k - 1}$ and consists of the points where, as long as the event in \cref{eq: est-bound} holds, we are sure arm $a$ is optimal.
The second is $\de_{-a, k - 1}$ and, in contrast to the first, we cannot rely on the CATE estimator in order to determine that $a$ is optimal here. Nevertheless, we can show that $\de_{-a, k - 1} \cap \mathcal{X}  \subseteq \{x \in \mathcal{X}: a\tau(x) > 0\}$ under \cref{assump: decision-set} and as long as the event in \cref{eq: est-bound} holds (\cref{lemma: decision-set-characterization}).
This means that we can conclude that the arm $a$ is also optimal on $\de_{-a, k - 1}$ even though we cannot estimate CATE accurately there.

The remaining randomization region in each epoch, $\mathcal R_k$, consists of the subset of the previous randomization region where we cannot determine that either arm is optimal using either of the above criteria. In particular, the CATE estimate is below the accuracy target inside $\mathcal R_k$, $\abs{\hat\tau_{k-1}(x)}\leq\epsilon_{k-1}$, so, even when the event in \cref{eq: est-bound} holds, we cannot be sure which arm is optimal. Thus, we may as well pull each arm uniformly at random to provide maximum exploration for estimation in future epochs.
Moreover, the exploration cost is manageable since, as long as the event in \cref{eq: est-bound} holds: (1) the regret incurred from pulling sub-optimal arms at the randomization region shrinks exponentially since $\abs{\tau(x)}\le\abs{\hat\tau_{k-1}(x)}+\epsilon_{k-1}\leq2\epsilon_{k-1}$ for $x \in \rand_{k}$; and (2) the randomization region shrinks over the epochs, as \cref{assump: margin} implies that
$\pr(X \in \rand_{k} \cap \mathcal{X}, |\tau(X)| \ne 0)  \le  \mu(\{X: 0 < \abs{\tau(X)} \le 2\epsilon_{k-1}\}) \le \MarginL(2\epsilon_{k-1})^\alpha$.
In each epoch, we update the CATE estimates and the decision rule only where it is needed. We estimate CATE and design new decision regions (\ie, $\rand_{k}$ and $\exploit_{\pm 1, k}$) only within the region where we failed to learn the optimal arm with high confidence in previous epochs (\ie, $\rand_{k - 1}$), and we follow previous decision rules on regions where the optimal arm is already learned with high confidence (\ie, $\bigcup_{j = 1}^{k - 1} \exploit_{a, j}$).
In this way, we gradually refine the accuracy of CATE estimator in ambiguous regions, while making efficient use of the information learned in previous epochs.

{\blockedit\subsubsection{Comparison with (A)BSE when $\beta=1$} \label{section: large margin constant}
While our algorithm is most notable for tackling the case of $\beta>1$, it also handles the special case of $\beta=1$, which is exactly the intersection point with the previous literature that focus on $\beta\leq1$ \citep{Rigollet10,perchet2013}. Even when $\beta=1$, our algorithm is distinct from these. Notably, UCBograms in \citet{Rigollet10} and BSE in \citet{perchet2013}, which both run isolated MAB algorithms in each hypercube, can achieve minimax optimal regret only when $\alpha \le 1$ (and $\beta\leq1$). Since theses algorithms learn expected rewards using data only within each hypercube, they require pulling each arm at least once in each hypercube and thus necessarily incur regret of $\Omega(T^{\frac{d}{2 + d}})$ when $\beta = 1$, which is suboptimal when $\alpha > 1$ since the minimax regret rate in this case is $\Theta(T^{\frac{1 - \alpha + d}{2 + d}})$.
Indeed, when $\alpha > 1$, the expected rewards for two arms are relatively separated, and we can tell apart the optimal arms with relatively little data, so pulling each arm at least once in every hypercube may may be wasteful.
Addressing this is the central purpose of the ABSE algorithm in \citet{perchet2013}, which gradually refines the hypercubes (though still it can only handle $\beta\leq1$).

Our algorithm provides another way around this issue by using data across hypercubes \emph{even} in the special case of $\beta=1$, where our bandwidth is larger than the hypercube size in all but the last epoch. Then, whenever a particular hypercube has arms that are well-separated (as many must when $\alpha$ is large), we can still detect this even if we did not pull both arms in this hypercube.
For example, in the $k^{\text{th}}$ epoch, \cref{lemma: h-size} ensures that the learning radius $H_{\pm1, k}$ is much larger than the hypercube size $\delta$, so even if we have not pulled one of the arms in some hypercubes yet, 
we can still collect enough samples (with high probability) for both arms in their  neighborhood within the learning radius, so that we can construct CATE estimator $\hat \tau_{k -1}$ that achieves the target precision level $\epsilon_{k - 1}$ (see \cref{thm: good-prob} for the formal statement).
If the expected rewards for the two arms on some of these hypercubes are separated enough  so that $|\hat \tau_{k -1}(x)| > \epsilon_{k - 1}$ on them, then we can confidently push them into exploitation regions. As a result, we do not ``waste'' arm pulls in these hypercubes.
Importantly, our algorithm can determine optimal arms on hypercubes with well separated expected rewards in early epochs using relatively imprecise CATE estimators based on small samples, and do so on more difficult hypercubes with less separated expected rewards later on using more precise CATE estimators.
In this way, it carefully achieves the minimax optimal regret rate even when $\alpha > 1$ (see \cref{thm: regret}).
For $\beta=1$ the behavior of our algorithm is similar but different to ABSE in that both gradually refine the learning radius, but in ABSE the learning radius is set to be the same as the hypercube size, while in our algorithm the learning radius is different from the hypercube size.}

\subsubsection{Finite Running Time}

Finally, we remark that \cref{alg} can be run in finite time. First, we show that all decision regions are unions of hypercubes in $\mathcal{C}$, as shown in \cref{fig: algorithm}.
\mybigskip
\begin{lemma}\label{lemma: union-cube}
For $1 \le k \le K$, $\exploit_{\pm 1, k}$, $\de_{\pm 1, k}$ and $\rand_k$ are all unions of hypercubes in $\mathcal{C}$.
\end{lemma}
The number of hypercubes itself, $\abs{G}$, is of course finite.
To determine in what hypercube an arriving context falls, we need only divide each of its coordinate by $\delta$.

The remaining question is to compute which hypercubes belong to which decision region at the start of each epoch.
To compute $\de_{\pm 1, k}$, we need to compute the volume in the intersection of $\mathcal X$, a union of cubes, and a ball and compare it to a given constant. We need to do this at most once in each epoch for every hypercube. If $\mathcal X$ has a simple shape such as the unit hypercube, this can be done analytically. Alternatively, given a membership oracle for $\mathcal X$, we can compute this using rectangle quadrature integration.
In particular, we can easily allow for some slowly vanishing approximation error in the quadrature integration without deteriorating the regret rate of our algorithm.
Then, to compute $\exploit_{\pm 1, k}$ and $\rand_k$, we need only to compute $\hat\eta_{k,a}(x)$ at most once in each epoch at each lattice point $x\in G$. Computing this estimate requires constructing an $M_\beta\times M_\beta$ matrix given by averaging over the data within the bandwidth neighborhood and  then pseudo-inverting this matrix.

\section{Theoretical Guarantees: Upper and Lower Bounds on Minimax Regret}\label{section: theory}

We next provide two results that together characterize the minimax regret rate (up to polylogs): an upper bound on the regret of our algorithm and a matching lower bound on the regret of any other algorithm.

\subsection{Regret Upper Bound}\label{section: ub}
In this section, we derive an upper bound on the regret of our algorithm. The performance of our algorithm, as we will show in this section, crucially depends on two events: $\many_k$, the event that sufficiently \textit{many} samples for each arm are available for CATE estimation at the end of epoch $k$, and $\good_{k}$, the event that our estimator $\hat{\tau}_k$ has \textit{good} accuracy.
Concretely,
\begin{align} \notag
\many_k  &= \bigg\{
		{N_{1, k}\wedge N_{-1, k}}\ge \prns{\frac{\log(T\delta^{-d})}{C_0\epsilon_k^2}}^{\frac{2\beta +d}{2\beta}}
	\bigg\},\\\notag
	\good_{k}
		&= \bigg\{\sup_{x\in \mathcal{R}_{k} \cap D_{1, {k}}^C \cap D^C_{-1, {k}}}\abs{\hat{\tau}_{k}(x)-\tau(x)} \le \epsilon_{ k}\bigg\}.
\end{align}
For convenience, we also define $\ogood_k = \bigcap_{1 \le j \le k}\good_k$ and $\omany_k = \bigcap_{1 \le j \le k}\many_k$, where an empty intersection ($\ogood_0$ or $\omany_0$) is the whole event space (always true).

\subsubsection{Characterization of the decision regions.} The following lemma shows that these two events are critical for the effectiveness of the proposed decision rules, in that whenever they hold, we have the desired behavior described in \cref{sec: algo regions,sec: algo cleaning}.
\mybigskip
\begin{lemma}\label{lemma: decision-set-characterization}
	Fix any $k\ge 1$. Suppose \cref{assump: decision-set} holds and that
  \edit{$T \ge T_0 \vee \prns{\exp(1 \vee \frac{C_0(2\beta +d)}{4 (2r_0)^{2\beta}(2\beta+d + \beta d)})}$}
  with $T_0$ given in \cref{lemma: h-size} and $C_0$ given in \cref{lemma: eigenvalue}. Then, under the event $\ogood_{k-1} \cap \omany_{k - 1}$, we have for $a=\pm1$:
	\begin{enumerate}[label=\roman*.,ref=\roman*]
		\item $\rand_{k} \cap \mathcal{X} \subseteq \{x \in \mathcal{X}: \abs{\tau(x)} \le 2\epsilon_{k - 1}\}$, \label{lemma: decision-set-characterization1}
		\item $(\bigcup_{j=1}^k \mathcal{E}_{a,j})\cap \mathcal{X} \subseteq \{x \in \mathcal{X}:a\tau(x)>0\}$,\label{lemma: decision-set-characterization2}
		\item $\mathcal{Q}_a\subseteq \big((\bigcup_{j=1}^{k} \mathcal{E}_{a,j})\cup \mathcal{R}_k\big) \cap \mathcal{X}$, and\label{lemma: decision-set-characterization3}
		\item $\de_{a, k}\cap \mathcal{X} \subseteq \{x \in \mathcal{X}: a\tau(x) < 0\}$.\label{lemma: decision-set-characterization4}
	\end{enumerate}
\end{lemma}

In \cref{lemma: decision-set-characterization}, statement \ref{lemma: decision-set-characterization1} means that we cannot identify the optimal arm on the randomization region $\rand_{k}$. Statement \ref{lemma: decision-set-characterization2} says that pulling arm $a$ on the exploitation region $\bigcup_{j=1}^k \mathcal{E}_{a,j}$ is optimal.
Statement \ref{lemma: decision-set-characterization3} shows that the support of the sample $S_{a, k}$ (\ie, $\big((\bigcup_{j=1}^{k} \mathcal{E}_{a,j})\cup \mathcal{R}_k\big) \cap \mathcal{X}$) always contains the region where arm $a$ is optimal, $\mathcal{Q}_a$.
Statement \ref{lemma: decision-set-characterization4} says that the optimal arm on $\de_{a, k}$ is $-a$, which justifies why we put $\de_{a, k}$ into $\exploit_{-a, k}$ in \cref{eq: exploit}.
Recall that on $\de_{a, k}$, the support of the sample $S_{a, k}$ is insufficiently regular and thus we cannot hope to obtain good estimates there.
Fortunately, statement \ref{lemma: decision-set-characterization4} guarantees that accurate decision making is still \textit{possible} on $\de_{a, k}$ even though accurate CATE estimation is \textit{impossible}.

Statement \ref{lemma: decision-set-characterization3} in \cref{lemma: decision-set-characterization} is crucial. On the one hand, it is critical in guaranteeing that sufficient samples can be collected for both arms for future epochs (see also the discussion following \cref{thm: good-prob}).
On the other hand, it leads to statement \ref{lemma: decision-set-characterization4}, which enables us to make correct decisions in the inestimable regions. The argument is roughly as follows. Given statement \ref{lemma: decision-set-characterization3}, if statement \ref{lemma: decision-set-characterization4} didn't hold, \ie, if there were any $x_0 \in \de_{a, k}\cap \mathcal{X}$ such that $x_0 \in \mathcal{Q}_a = \{x \in \mathcal X: a\tau(x) \ge 0\}$, then by the regularity of $\mathcal{Q}_a$ imposed by \cref{assump: decision-set}, $\big((\bigcup_{j=1}^{k} \mathcal{E}_{a,j})\cup \mathcal{R}_k\big) \cap \mathcal{X}$ would be sufficiently regular at $g(x_0)$, which violates the construction of $\de_{a, k}$ in \cref{eq: deletion-rule}.

\subsubsection{A preliminary regret analysis.}
Based on \cref{lemma: decision-set-characterization}, we can  decompose the regret according to $ \ogood_{k - 1}\cap\omany_{k - 1}$. Let $\hat\pi$ denote our algorithm, \cref{alg}. Then:
	\begin{align*}\textstyle
		R_T(\hat{\pi})
			&=  \sum_{k=1}^K \sum_{t\in\mathcal{T}_k} \expect [Y_{t}(\pi^*(X_t)) - Y_{t}(A_t)] \\
			&\le \sum_{k=1}^K\sum_{t\in\mathcal{T}_k} \expect[ Y_{t}(\pi^*(X_t)) - Y_{t}(A_t) \mid \ogood_{k - 1} \cap \omany_{k - 1}]
			 + \sum_{k=1}^K\sum_{t\in\mathcal{T}_k} \pr(\ogood_{k - 1}^C \cup \omany_{k - 1}^C).
	\end{align*}
We can further decompose the regret in the $k^\text{th}$ epoch given $ \ogood_{k - 1} \cap \omany_{k - 1}$ into the regret due to exploitation in $\bigcup_{j = 1}^k \exploit_{1, j}\cup \exploit_{-1, j}$ and the regret due to exploration in $\rand_k$:
	\begin{align*}
		&\sum_{t\in\mathcal{T}_k} \expect[ Y_{t}(\pi^*(X_t)) - Y_{t}(A_t) \mid \ogood_{k - 1} \cap \omany_{k - 1}] \\
		&\qquad\le \sum_{t\in\mathcal{T}_k} \expect\big[ Y_{t}(\pi^*(X_t)) - Y_{t}(A_t) \mid \ogood_{k - 1} \cap \omany_{k - 1}, X_t \in ({\textstyle\bigcup_{j = 1}^k} \exploit_{1, j}\cup \exploit_{-1, j} )\big] \\
		&\qquad\phantom{\le}+ \sum_{t\in\mathcal{T}_k} \expect\big[\abs{\tau(X_t)}\mid \ogood_{k - 1} \cap \omany_{k - 1}, X_t \in \rand_k\big]\pr\big(X_t \in \rand_k \mid \ogood_{k - 1} \cap \omany_{k - 1}\big).
	\end{align*}
\cref{lemma: decision-set-characterization} statement \ref{lemma: decision-set-characterization2} implies that the proposed algorithm always pulls the optimal arm on the exploitation region. Therefore, the first term on the right-hand side, \ie, the regret due to exploitation, is equal to $0$. Moreover,
	\begin{align*}
	&\sum_{t\in\mathcal{T}_k} \expect\big[\abs{\tau(X_t)}\mid \ogood_{k - 1} \cap \omany_{k - 1}, X_t \in \rand_k\big]\pr\big(X_t \in \rand_k \mid \ogood_{k - 1} \cap \omany_{k - 1}\big)  \\
	&\qquad\le \sum_{t\in\mathcal{T}_k} 2\epsilon_{k-1}\mathbb{P}(0 < \abs{\tau(X_t)}\le 2\epsilon_{k-1} \mid  \ogood_{k - 1} \cap \omany_{k - 1}) \le  \MarginL2^{1+\alpha}\epsilon_{k-1}^{1+\alpha} n_k,
	\end{align*}
where the first inequality follows from \cref{lemma: decision-set-characterization} statement \ref{lemma: decision-set-characterization1}, and the second inequality follows from the margin condition of \cref{assump: margin}.

Therefore, the total regret is bounded as follows:
\begin{align}\label{eq: regret-intermediate}
R_T(\hat{\pi})
	&\le \sum_{k = 1}^K \MarginL2^{1+\alpha}\epsilon_{k-1}^{1+\alpha} n_k  + \sum_{k=1}^K n_k \pr(\ogood_{k - 1}^C \cup \omany_{k - 1}^C) \nonumber \\
	&\le \edit{=O(
        T^{\frac{\beta+d-\alpha\beta}{2\beta+d}}\log^{1+\frac{d}{2\beta}}(T)+\log^2(T))} + \sum_{k=1}^K n_k \pr(\ogood_{k - 1}^C \cup \omany_{k - 1}^C),
\end{align}
where the \edit{${O}(\cdot)$} term depends only on the parameters of \cref{assump: smoothness,assump: support,assump: margin,assump: decision-set} and not on the particular instance.
Thus, if we can prove that $\pr(\ogood_{k - 1}^C \cup \omany_{k - 1}^C)$ is small enough for all $k$, then we can (uniformly) bound the cumulative regret $R_T(\hat{\pi})$ of our proposed algorithm.

\subsubsection{Bounding $\pr(\ogood_{k - 1}^C \cup \omany_{k - 1}^C)$.} The analysis in \cref{eq: regret-intermediate} shows that the cumulative regret of the proposed algorithm depends on the probability of $\ogood_{k - 1}^C \cup \omany_{k - 1}^C$, \ie, that the CATE estimator may not be accurate enough or that the total sample size for one arm is not sufficient in any epoch prior to the $k^\text{th}$ epoch.

To bound this probability, we need to analyze the distribution of the samples for each arm.
The sample distributions in each epoch can be distorted by decisions in previous epochs. Since a well-behaved density is crucial for nonparametric estimation, we must make sure that such distortions do not undermine our CATE estimation.
\mybigskip
\begin{lemma}\label{lemma: strong-density}
For any $1 \le k \le K$ and $a = \pm 1$,
$S_{a, k} = \{(X_t, Y_t): t \in \mathcal{T}_{k}, A_t = a\}$ are conditionally i.i.d. samples, given $\mathcal{F}_{k - 1}\cup\mathcal A_k$, where $\mathcal{F}_{k - 1} = \{(X_t, A_t, Y_t): t \in \bigcup_{k' = 1}^{k - 1}\mathcal{T}_{k'}\}$, $\mathcal A_k=\{A_t: t \in \mathcal{T}_k\}$.

Now suppose \cref{assump: decision-set,assump: support} hold, let $C_0$ be defined as in \cref{lemma: eigenvalue} below for any given $\beta,\,L_1, c_0, r_0, \mu_{\min}$, and suppose
\edit{$T \ge T_0 \vee \prns{\exp(1 \vee \frac{C_0(2\beta +d)}{4 (2r_0)^{2\beta}(2\beta+d + \beta d)})} $}.
Then, for $a=\pm1$, under the event $\ogood_{k - 1}\cap \omany_{k - 1}$, the (common) conditional density of any of $\{X_t: A_t = a, t \in \mathcal{T}_k\}$ with respect to Lebesgue measure, given $\mathcal{F}_{k - 1}\cup\mathcal A_k$, which we denote by $\mu_{a, k}$,
satisfies the following conditions:
	\begin{enumerate}
		\item $\frac{1}{2}\mu_{\min}\le \mu_{a,k}(x)\le \frac{2\mu_{\max}}{p}$ for any $x \in \big((\bigcup_{j=1}^{k} \mathcal{E}_{a,j})\cup \mathcal{R}_k\big) \cap \mathcal{X}$.
		\item $ \mu_{a,k}(x)=0$ for any $x \in \prns{\bigcup_{j=1}^{k} \mathcal{E}_{-a,j}} \cap \mathcal{X}$.
	\end{enumerate}
\end{lemma}
\Cref{lemma: strong-density} shows that in the $k^\text{th}$ epoch, samples for each arm are i.i.d given the history, and it satisfies a strong density condition on the support of each sample, $\big((\bigcup_{j=1}^{k} \mathcal{E}_{a,j})\cup \mathcal{R}_k\big) \cap \mathcal{X}$.
Furthermore, this distribution support set is sufficiently regular with respect to points in $\mathcal{R}_{k} \cap \mathcal{D}_{1, k}^C \cap \mathcal{D}_{-1, k}^C$, according to the screening rule given in \cref{eq: deletion-rule}.
Together, this strong density condition and support set regularity condition guarantee that we can estimate CATE using local polynomial estimators well on $\rand_k$ in the $(k + 1)^\text{th}$ epoch,  after we remove the inestimable regions.

In particular, the following lemma shows that the local polynomial estimator is well-conditioned with high probability, which echoes the classic result in the offline setting (\cref{prop: ivertibility}).
\mybigskip
\begin{lemma}\label{lemma: eigenvalue}
Suppose the conditions of \cref{lemma: strong-density} hold.
Let $1\le k \le K-1,\,a = \pm 1,\,n_{\pm 1, k}$ be given.
Consider the Gram matrices of the local polynomial regression estimators in \cref{eq: cate-estimator}, \ie, $\hat{\mathcal{A}}(x; S_{a, k}, H_{a,k}, \floor*{\beta})$ as defined in \cref{eq: gram matrix}.
Then, given $N_{\pm1,k}=n_{\pm1,k}$ and $\omany_{k - 1} \cap \ogood_{k - 1}$, these satisfy the following with conditional probability at least
$1 - 2M_{\beta}^2\exp\big\{-C_0\big(4(1+ L_1\sqrt{d})^2\big)n_{a, k}^{\nicefrac{2\beta}{(2\beta + d)}}\big\}$:
\[
	\lambda_{\min}(\hat{\mathcal{A}}(x; S_{a, k}, H_{a,k}, \floor*{\beta})) \ge \lambda_0 > 0, ~~~~ \forall x \in \rand_{k} \cap \de_{1, k}^C \cap \de_{-1, k}^C,
\]
 where
\begin{align*}
	\lambda_0 &=
	\frac{1}{4}\mu_{\min}
	\inf_{\substack{W \in \mathbb{R}^d,\, S \subset \mathbb{R}^d\,:\, \magd{W}=1 \\ S \subseteq \mathcal{B}(0,1) ~\text{is compact},~\leb(S)=  \nicefrac{c_0v_d}{2^d}}} \int_S \bigg(\sum_{|s|\le \floor*{\beta}}W_s u^s\bigg)^2d u, \\
		C_0 &= \frac{3p \lambda_0^2}{4(1+ L_1\sqrt{d})^2}\min\bigg\{\frac{1}{12M_{\beta}^4\mu_{\max}v_d+2p\lambda_0 M_{\beta}^2}, \frac{1}{108 M_{\beta}v_d\mu_{\max}+6\sqrt{M_{\beta}}p\lambda_0}, \\&\qquad\qquad\qquad\qquad\qquad\qquad\qquad
		\frac{1}{108 M_{\beta}L^2v_d\mu_{\max} + 6\sqrt{M_{\beta}}L(2v_d\mu_{\max}+p)\lambda_0}\bigg\}.
\end{align*}
\end{lemma}
In \cref{lemma: eigenvalue}, $\lambda_0$ is positive because the unit shell is compact and, for fixed $W$, the infimum over $S$ is continuous in $W$ and positive as the integrand can be zero only a measure-zero set while $S$ has positive measure. The constant $C_0$ dictates the epoch schedule $\{\mathcal T_k\}_{k = 1}^K$ of our proposed algorithm (see \cref{section: algorithm}).
Note that we can also use any positive constant no larger than $C_0$ in our algorithm without deteriorating the regret rate.

In the following theorem, we show that $\pr(\ogood_{k - 1}^C \cup \omany_{k - 1}^C)$ is indeed very small for large $T$, so its contribution to the cumulative regret bound in \cref{eq: regret-intermediate} is negligible.
\mybigskip
\begin{theorem}\label{thm: good-prob}
	When
  \edit{$T \ge T_0 \vee \prns{\exp(1 \vee \frac{C_0(2\beta +d)}{4 (2r_0)^{2\beta}(2\beta+d + \beta d)} \vee  \frac{36{M}_{\beta}L^2v^2_d\mu^2_{\max}C_0(2\beta + d)}{p^2\lambda^2_0 (2\beta + d + \beta d)})}$},
  if we assume \cref{assump: smoothness,assump: decision-set,assump: support},
	then for any $1\le k\le K-1$,
	\begin{align*}
		&\qquad \pr (\good_{k}^C \mid \ogood_{k-1},  \omany_k )
			\le \frac{8+4M_{\beta}^2}{T}, \quad \pr(\many_k^C \mid \ogood_{k - 1}, \omany_{k-1})
			\le \frac{2}{T},\quad \pr(\ogood_k^C \cup \omany_{k}^C) \le \frac{(10 + 4M_{\beta}^2)k}{T}.
	\end{align*}
\end{theorem}
Here the upper bound on $\pr\big(\good_{k}^C \mid \ogood_{k-1},  \omany_k \big)$ is derived from the uniform convergence of local polynomial regression estimators \citep{stone1982} given a well-conditioned Gram matrices (which we ensure in \cref{lemma: eigenvalue}) and sufficiently many samples for each arm (ensured by $\many_k$) whose sample distribution satisfies strong density condition (which we ensure in \cref{lemma: strong-density}).
The upper bound on $\pr(\many_k^C \mid \ogood_{k - 1}, \omany_{k-1})$ arises from
\cref{lemma: nomempty-opt}
 and \cref{lemma: decision-set-characterization} statement \ref{lemma: decision-set-characterization3}, since they imply that $\pr\big(X \in (\bigcup_{j=1}^{k} \mathcal{E}_{a,j})\cup \mathcal{R}_k\big) \ge\pr(X\in \mathcal Q_a)\ge p$ for $a = \pm 1$. As a result, at least a constant fraction of $n_k$ many samples will accumulate for each arm, so that $\many_{k}$ holds with high probability
 as the $n_k$ proposed in \cref{eq: sample-size} is sufficiently large
 .
The upper bound on $\pr(\ogood_k^C \cup \omany_{k}^C)$ follow from the first two upper bounds by induction.

\subsubsection{Regret Upper Bound.} Given \cref{thm: good-prob} and \cref{eq: regret-intermediate}, we are now prepared to derive the final upper bound on our regret.
\mybigskip
\begin{theorem}\label{thm: regret}
Suppose \cref{assump: smoothness,assump: margin,assump: decision-set,assump: support} hold.
	Then,
	\begin{align*}
		R_T(\pi)
        \edit{=O(
        T^{\frac{\beta+d-\alpha\beta}{2\beta+d}}\log^{1+\frac{d}{2\beta}}(T)+\log^2(T))}
        =\Tilde{O}(T^{\frac{\beta+d-\alpha\beta}{2\beta+d}}\edit{+1}),
	\end{align*}
    where the \edit{$O(\cdot)$ and} $\Tilde{O}(\cdot)$ terms only depend on the parameters of \cref{assump: smoothness,assump: margin,assump: decision-set,assump: support}. (An explicit form is given in the proof.)
\end{theorem}
\proof{Proof sketch.}
	\cref{thm: good-prob} states that for $2 \le k \le K$,
	\[
		n_k\pr(\ogood_{k - 1}^C \cup \omany_{k - 1}^C) \le n_k \frac{(10 + 4M_{\beta}^2)(k - 1)}{T} \le (10 + 4M_{\beta}^2)(k - 1).
	\]
	Furthermore, \cref{lemma: k-bound} implies that,
	\begin{align*}
		K &\le \lceil \frac{\beta}{(2\beta+d)\log 2}\log T \rceil.
	\end{align*}
	Thus
	\begin{align*}
	\sum_{k=1}^K n_k \pr(\ogood_{k - 1}^C \cup \omany_{k - 1}^C)
		&\le (5 + 2M_{\beta}^2)K^2 \\
		&\le (5 + 4M_{\beta}^2)\frac{\beta^2 \log^2 T}{(2\beta + d)^2\log^2 2} = \tilde{O}(1).
	\end{align*}
	The final conclusion follows from \cref{eq: regret-intermediate}.
\endproof
A complete and detailed proof is given in the supplement.
\mybigskip
\begin{corollary}
Let any problem parameters be given. Then, for the corresponding class of contextual bandit problems $\mathcal P$, the minimax regret satisfies
$$\mathcal R_T=\Tilde{O}(T^{\frac{\beta+d-\alpha\beta}{2\beta+d}}).
$$
\end{corollary}

\subsection{Regret Lower Bound} \label{section: lower bound}
In this section, we prove a matching lower bound (up to polylogarithmic factors) for the regret rate in \cref{thm: regret}. This means that there does not exist any other algorithm that can achieve a lower rate of regret for all smooth bandit instances in a given smoothness class. Thus, our algorithm achieves the minimax-optimal regret rate.
\begin{theorem}[Regret Lower Bound]\label{lb}
Fix any positive parameters $\alpha,\beta,d, L, L_1$ satisfying $\alpha\beta\le d$.
For any admissible policy $\pi$ and $T$,
there exists a contextual bandit instance
satisfying \cref{assump: support,assump: smoothness,assump: margin,assump: decision-set} with the provided parameters such that
\begin{equation} \label{eq: lower}
  \sup_{\mathbb{P}\in \mathcal{P}} R_T(\pi) = \Omega(T^{\frac{\beta+d-\alpha\beta}{2\beta+d}}),
\end{equation}
where the $\Omega(\cdot)$ term only depends on the parameters of the class $\mathcal{P}$ and not on $\pi$.
Hence, we also have $\mathcal R_T = \Omega(T^{\frac{\beta+d-\alpha\beta}{2\beta+d}})$.
\end{theorem}
\proof{Proof sketch.}
Define the \textit{inferior sampling rate} of a given policy $\pi$ as the expected number of times that $\pi$ disagrees with the oracle policy $\pi^*$ (for a given instance $\mathbb P$), \ie,
\begin{equation*}
  \textstyle I_T(\pi) = \expect  \left[\sum_{t = 1}^T \ind\left(\pi^*(X_t) \neq \pi_t(X_t)\right)\right].
\end{equation*}

Lemma 3.1 in \cite{Rigollet10} relates $R_T(\pi)$ to $ I_T(\pi)$: under \cref{assump: margin},
\begin{align}\label{eq: inferior-regret}
	I_T(\pi) = O(T^{\frac{1}{1+\alpha}} R_T(\pi)^{\frac{\alpha}{1+\alpha}}).
\end{align}
Note the implicit dependence of $I_T(\pi),\,R_T(\pi)$ on the instance $\mathbb P$.

We then construct a finite class, $\mathcal{H}$, of contextual bandit instances with smooth expected rewards and show, first, that $\mathcal H\subseteq\mathcal{P}$, \ie, that our construction fits the provided parameters (\edit{in particular, our construction is fundamentally different from that in \citealp{Rigollet10}, as their construction approach is only suitable for non-differentiable functions}), and, second, that
\begin{align}\label{eq: inferior-rate}
\sup_{\mathbb{P}\in \mathcal{P}}	I_T(\pi) \geq \sup_{\mathbb{P}\in \mathcal{H}}	I_T(\pi)
\geq
\frac{1}{\abs{\mathcal{H}}}\sum_{\mathbb P\in\mathcal H}    I_T(\pi)
= \Omega(T^{1-\frac{\alpha\beta}{2\beta+d}}).
\end{align}
We arrive at the final conclusion by combining \cref{eq: inferior-regret,eq: inferior-rate}.
\endproof

A complete and detailed proof is given in the supplement.

Note that in \cref{lb}, we allow $\alpha,\,\beta,\,d,\,L,\,L_1$ to be given. The proof then constructs an example with appropriate values for the rest of the parameters, $c_0,\,r_0,\,\mu_{\max},\,\mu_{\min},\,\MarginL$, for which the class of bandit problems $\mathcal P$ satisfies the above lower bound. This shows that the rate given in \cref{thm: regret} is tight (for the regime $\alpha\beta\leq d$).

\edit{We can furthermore show that our non-standard
\cref{assump: decision-set} is \textit{necessary} for achieving the minimax regret rate $\tilde{\Theta}(T^{\frac{\beta + d - \alpha\beta}{2\beta + d}}\edit{+1})$.
In particular, we will prove that there exists a class of problem instances satisfying \cref{assump: smoothness,assump: support,assump: margin} but not necessarily \cref{assump: decision-set} such that the corresponding regret lower bound is \emph{higher} in order.
}
\edit{
\begin{theorem}[Regret Lower Bound, without \cref{assump: decision-set}]\label{lb2}
Under the conditions of \cref{lb}, for any admissible policy $\pi$ and $T$,
there exists a contextual bandit instance
satisfying \cref{assump: smoothness,assump: margin,assump: support} (but not necessarily \cref{assump: decision-set}) with the provided parameters such that
\begin{align}\label{eq: lower2}
\frac{
R_T(\pi)}{T^{\frac{\beta+d-\alpha\beta}{2\beta+d}}} = \Omega\left(T^{\frac{\alpha(1 + \alpha)\beta^2 d}{2(2\beta + d)(\alpha\beta^2  + 2\beta d + \alpha\beta d + d^2)}}\right) \to \infty  ~~~ \text{ as } ~~ T \to \infty.
\end{align}
\end{theorem}
\proof{Proof sketch.}
  Similar to the proof of \cref{lb},
  we can construct another finite class
  $\mathcal H'$
   of bandit problems that satisfy \cref{assump: smoothness,assump: margin,assump: support} but  not \cref{assump: decision-set} with the provided parameters, and show that for $\Delta = \frac{1}{2}(\beta + \frac{\beta d}{\alpha \beta + d})$,
  \begin{align}\label{eq: inferior-rate2}
  \sup_{\mathbb{P}\in \mathcal{H}'}	I_T(\pi)
  \geq
  \frac{1}{\abs{\mathcal{H}'}}\sum_{\mathbb P\in\mathcal H'}    I_T(\pi)
  = \Omega(T^{1-\frac{\alpha\Delta}{2\Delta+d}}).
  \end{align}
  The conclusion \cref{eq: lower2} then follows from \cref{eq: inferior-regret,eq: inferior-rate2}.
\endproof
}

\section{Conclusions} \label{section: conclusion}

In this paper, we defined and solved the smooth-response contextual bandit problem. We proposed a rate-optimal algorithm that interpolates between using global and local reward information according to the underlying smoothness structure.
Our results connect disparate results for contextual bandits and bridge the gap between linear-response and non-differentiable bandits, and contribute to revealing the whole landscape of contextual bandit regret and its interplay with the inherent complexity of the underlying learning problem.

\bibliographystyle{plainnat}
\bibliography{bandit}

\newpage

\gdef\thesection{\Alph{section}} %
\makeatletter
\renewcommand\@seccntformat[1]{Appendix \csname the#1\endcsname.\hspace{0.5em}}
\makeatother

\appendix
\begin{algorithm}[H]
    \caption{\textsc{SmoothBandit (Multiple Arms)}}
    \label{alg-multi-arm}
    \edit{
    \begin{algorithmic}[1]
    \STATEx\textbf{Input:} Grid lattice $G$ and hypercubes $\mathcal C$, epoch schedule $\{\mathcal{T}_{k}\}_{k = 1}^K$, H\"older smoothness constant $\beta$, regularity constant $c_0$, context dimension $d$.
    \STATE{Initialize $\mathcal A_{j, 1}  = \mathcal A$ for any $j \in \{1, \dots, |\mathcal C|\}$ and $\mathcal R_1(\mathcal A) = \mathcal X$}
    \FOR{$t \in \mathcal{T}_1$}
                \STATE Pull $A_t \in \mathcal A$ randomly, equiprobably
    \ENDFOR
    \STATE {Log the samples $S_{a, 1} = \{(X_t, Y_t): t \in \mathcal{T}_1, A_t = a\}$ for $a \in \mathcal A$}
    \FOR {$k = 2, 3, \dots, K$}
        \STATE Identify inestimable regions for local polynomial regression with bandwidth $H_{a , k-1}$ ($a \in \mathcal A$):
            \begin{align*}
                \mathcal{D}_{\text{Irr}, k - 1}(a) &= \bigcup\braces{\cube(x) :
                x \in \left(\bigcup_{\mathcal A_0: a \in \mathcal{A}_0}\mathcal{R}_{k - 1}(\mathcal A_0)\right)\cap {G} ,\begin{array}{l}\prns{\bigcup_{\mathcal A_0: a \in \mathcal{A}_0}\mathcal{R}_{k - 1}(\mathcal A_0)} \cap \mathcal{X}\text{ is not}\\\text{weakly $\prns{\frac{c_0}{2^d}, H_{a, k - 1}}$-regular at $x$}\end{array}}
                \label{eq: deletion-rule}
            \end{align*}
        \STATE Set $N_{a, k -1} = \abs{S_{a, k - 1}},\,H_{a, k - 1} = N_{a, k - 1}^{-\nicefrac{1}{(2\beta + d)}}$ for $a \in \mathcal A$
        \STATE Construct the local polynomial estimate for every $a \in \mathcal A$ and  $x\in  {G} \cap \left(\bigcup_{\mathcal A_0: a \in \mathcal{A}_0}\mathcal{R}_{k - 1}(\mathcal A_0)\right) \cap \mathcal{D}_{\text{Irr}, k - 1}^C(a)$:
        \begin{equation*}\label{eq: alg cate estimate}
        \hat{\eta}_{a, k-1}(x)=\hat{\eta}^{\text{LP}}(x; S_{a,k-1},H_{a,k-1}, \floor*{\beta})
        \end{equation*}
        and identify the region where arm $a$ is estimated to be suboptimal:
        \begin{align*}
        \mathcal D_{\text{Est}, k}(a) &=
    \bigcup\bigg\{
    \cube(x): \exists a' \in \mathcal A,
    x\in  \left(\bigcup_{\mathcal A_0: a, a'\in \mathcal{A}_0}\mathcal{R}_{k - 1}(\mathcal A_0)\right)\cap {G} \cap \mathcal{D}_{Irr, k - 1}^C(a) \cap \mathcal{D}_{Irr, k - 1}^C(a'), \nonumber \\
    &\qquad\qquad\qquad\qquad\qquad\qquad\qquad\qquad\qquad\qquad\qquad\qquad\hat{\eta}_{a', k-1}(x) - \hat{\eta}_{a, k-1}(x)> \epsilon_{k-1}
    \bigg\}
        \end{align*}
        \STATE Update active arm sets: for $j \in \{1, \dots, |\mathcal C|\}$,
        \begin{align*}
        \mathcal A_{j, k} = \mathcal A_{j, k-1} - \{a \in \mathcal A_{j, k - 1}: \cube_j \in \mathcal{D}_{\text{Irr}, k - 1}(a) \cup \mathcal D_{\text{Est}, k}(a)\},
        \end{align*}
        \STATE Update decision regions: for $\mathcal A_0 \subseteq \mathcal A$,
        \begin{align*}
        \mathcal R_{k}(\mathcal A_0) = \bigcup_{j = 1}^{|\mathcal C|} \{\cube_j(x): \mathcal A_{j, k} = \mathcal A_0\}.
        \end{align*}
            \FOR{$t \in \mathcal{T}_k$ and $\mathcal A_0 \subseteq \mathcal A$}
                \STATE \textbf{if } $X_t \in \mathcal R_{k}(\mathcal A_0)$ \textbf{ then} pull $A_t \in \mathcal A_0$ randomly, equiprobably
            \ENDFOR
        \STATE Log samples $S_{a, k} = \{(X_t, Y_t): t \in \mathcal{T}_k, A_t = a\}$ for $a \in \mathcal A$
    \ENDFOR
    \end{algorithmic}
    }
\end{algorithm}

\section{Extention to Multi-armed Setting} \label[appendix]{section: k arm}
\bedit
In this section, we consider the extension of the smooth contextual bandit to multiple arms, i.e., $A \in \mathcal A$ with $|\mathcal A|\geq2$.
In the following subsections, we discuss in order the extension of the problem and assumptions, the extension of our algorithm, and the extension of the regret upper bound.

\subsection{Extending the Problem}

We first extend \cref{assump: support,assump: smoothness,assump: decision-set,assump: margin} to the multi-armed setting as follows.
\begin{assumption}\label{assump: multi-arm}
Suppose the following conditions hold:
\begin{enumerate}[label=\roman*.,ref=\roman*]
\item $\eta_a(\cdot) = \expect[Y(a) \mid X = \cdot]$ is $(\beta,L,\mathcal{X})$-H\"older for $\beta\geq1$ and is also $(1,L_1,\mathcal{X})$-H\"older for every $a \in \mathcal A$.
\item \label{cond: multi-arm-regularity} $\mathcal Q_a = \{x \in \mathcal X: \eta_a(x) \ge \eta_{a'}(x) ~\forall a' \in\mathcal A\}$ is a non-empty $(c_0,r_0)$-regular set for every $a \in \mathcal A$.
\item \Cref{assump: support} (strong density condition) holds with parameters $\mu_{\min}$ and $\mu_{\max}$.
\item For parameters $\alpha\geq0$ and $\MarginL > 0$, we have
\[
\pr(0 < |\eta_{a}(X) - \eta_{a'}(X)| \le t) \le \gamma t^\alpha
\]
for any $t > 0$, $a, a' \in \mathcal A$.
\end{enumerate}
\end{assumption}
Note that when $|\mathcal A| = 2$, each condition above reduces to its counterpart in \cref{section: formulation}, that is, \cref{assump: support,assump: smoothness,assump: decision-set,assump: margin}, in order respectively. By following the proof of \cref{lemma: nomempty-opt}, we can again show that $\min_{a \in \mathcal A}\pr(X\in\mathcal{Q}_{a}) \ge p = \mu_{\min}c_0r_0^dv_d > 0$. In other words \cref{assump: multi-arm} implies that no arm is strictly suboptimal everywhere.

\subsection{Extending the Algorithm} To extend our algorithm to multi-armed setting, we follow the same grid structure in \cref{sec: grid} and the epoch structure in \cref{sec: epoch}, and employ the hypercube-wise constant local polynomial estimator in \cref{sec: algo cate est} to estimate the conditional reward function $\eta_a$ for each $a \in \mathcal A$. We only need to slightly change the epoch schedule by setting the length of the $k^{\text{th}}$ epoch according to the number of arms $|\mathcal A|$:
\[
	n_k = \left\lceil \frac{2|\mathcal A|}{p}\prns{\frac{\log(T\delta^{-d})}{C_{\mathcal{A}}\epsilon_k^2}}^{\frac{2\beta +d}{2\beta}}+\frac{|\mathcal A|^2}{2p^2}\log T \right\rceil
\]
where
\begin{align*}
C_{\mathcal{A}} &= \frac{3p \lambda_0^2}{4(1+ L_1\sqrt{d})^2}\min\bigg\{\frac{1}{6|\mathcal A|M_{\beta}^4\mu_{\max}v_d+2p\lambda_0 M_{\beta}^2}, \frac{1}{54|\mathcal A| M_{\beta}v_d\mu_{\max}+6\sqrt{M_{\beta}}p\lambda_0}, \\&\qquad\qquad\qquad\qquad\qquad\qquad\qquad
		\frac{1}{54|\mathcal A| M_{\beta}L^2v_d\mu_{\max} + 6\sqrt{M_{\beta}}L(|\mathcal A|v_d\mu_{\max}+p)\lambda_0}\bigg\}.
\end{align*}
Still, we can use any positive constant no larger than $C_{\mathcal A}$ in our algorithm without deteriorating the regret rate.

We mainly need to revise the decision rules in \cref{alg}. Indeed, when only two arms are available, eliminating one arm (either due to estimated suboptimal reward or inestimability screening) unambiguously suggests that the other arm should be pulled. However, this no longer applies when there are multiple arms.
To tackle with this problem, we associate each region with an active arm set consisting of all arms that we still pull with positive probability in this region.
Our algorithm (\cref{alg-multi-arm}) gradually eliminates arms from each active arm set, and explores the remaining arms with equal probability.
When only a single arm remains in the active arm set of a region, we exploit this arm in this region until the end of the time horizon.

More concretely, we number the hypercubes in $\mathcal C$ (see \cref{sec: grid}) as $\cube_j$ for $j = 1, \dots, |\mathcal C|$. At the $k$th epoch, we equip hypercube $\cube_j$ with an active arm set $\mathcal A_{j, k}$ where each arm within $\mathcal A_{j, k}$ is pulled with equal probability at the $k$th epoch. For example, when $\mathcal A_{j, k} = |3|$, the decision for $X \in \cube_j$ is $\pi_k(X) = a$ with probability $1/3$ for any $a \in \mathcal A_{j, k}$. When $|\mathcal A_{j, k}| = 1$, we exploit the unique arm in $\mathcal A_{j, k}$ when $X \in \cube_j$. In the first epoch, we start with $\mathcal A_{j,1}=\mathcal A$ for all $j = 1, \dots, |\mathcal C|$ so we explore all arms in all regions.

For any $\mathcal A_0 \subseteq \mathcal A$, we denote the region where the active arm set is exactly $\mathcal A_0$ in the $(k - 1)$th epoch by:
\[
\mathcal R_{k - 1}(\mathcal A_0) = \bigcup_{j = 1}^{|\mathcal C|} \{\cube_j(x): \mathcal A_{j, k - 1} = \mathcal A_0\}.
\]
Then, for each $a \in \mathcal A$, $\prns{\bigcup_{\mathcal A_0: a \in \mathcal{A}_0}\mathcal{R}_{k - 1}(\mathcal A_0)}\cap \mathcal X$ is the region where we can collect samples for arm $a$ in the $(k - 1)^{\text{th}}$ epoch, and we denote these samples as $S_{a, k - 1} = \{(X_t, Y_t): t \in \mathcal{T}_{k - 1}, A_t = a\}$ with size $N_{a, k -1}=|S_{a, k - 1}|$.

Similar to the two-arm setting (\cref{sec: algo cleaning}), we may not be able to estimate the conditional reward function for a certain arm because of irregularity. The following set is the irregular region for arm $a \in \mathcal A$:
\begin{align*}
                \mathcal{D}_{\text{Irr}, k - 1}(a) &= \bigcup\braces{\cube(x) ~:~
                x \in \left({\bigcup}_{\mathcal A_0: a \in \mathcal{A}_0}\mathcal{R}_{k - 1}(\mathcal A_0)\right)\cap {G} ,~~\begin{array}{l}\prns{{\bigcup}_{\mathcal A_0: a \in \mathcal{A}_0}\mathcal{R}_{k - 1}(\mathcal A_0)} \cap \mathcal{X}\text{ is not}\\\text{weakly $\prns{\frac{c_0}{2^d}, H_{a, k - 1}}$-regular at $x$}\end{array}}.
\end{align*}
We will show that although we cannot estimate $\eta_a$ reliably on $\mathcal D_{\text{Irr}, k - 1}$ due to irregularity, we can ensure that arm $a$ is strictly suboptimal on $\mathcal D_{\text{Irr}, k - 1}$, and thus can safely remove arm $a$ from the corresponding active arm sets (\cref{lemma: decision-set-characterization-multi-arm} statement \ref{decision-set-4}).

The other scenario where we can safely remove an arm is when the local polynomial estimates confidently determine that the arm is suboptimal.
For each $a\in\mathcal A$, we next define the region in which there exists an $a'\in\mathcal A$ such that $\eta_a$ and $\eta_{a'}$ can both be reliably estimated based on samples from the $(k-1)$th epoch and
arm $a$ is estimated to be outperformed by arm $a'$
so that arm $a$ can be eliminated:
\begin{align*}
\mathcal D_{\text{Est}, k}(a) &=
    \bigcup\bigg\{
    \cube(x): \exists a' \in \mathcal A,
    x\in  \left(\bigcup_{\mathcal A_0: a, a'\in \mathcal{A}_0}\mathcal{R}_{k - 1}(\mathcal A_0)\right)\cap {G} \cap \mathcal{D}_{Irr, k - 1}^C(a) \cap \mathcal{D}_{Irr, k - 1}^C(a'), \nonumber \\
    &\qquad\qquad\qquad\qquad\qquad\qquad\qquad\qquad\qquad\qquad\qquad\qquad\hat{\eta}_{a', k-1}(x) - \hat{\eta}_{a, k-1}(x)> \epsilon_{k-1}
    \bigg\}
\end{align*}
It follows that we can remove arm $a$ from the active arm sets of all hypercubes in $\mathcal{D}_{\text{Irr}, k - 1}(a)$ and $\mathcal D_{\text{Est}, k}(a)$. So, for $j = 1, \dots, |\mathcal C|$, we set
\begin{align}\label{eq: multi-arm-update}
\mathcal A_{j, k} = \mathcal A_{j, k-1} - \{a \in \mathcal A_{j, k - 1}: \cube_j \in \mathcal{D}_{\text{Irr}, k - 1}(a) \cup \mathcal D_{\text{Est}, k}(a)\}.
\end{align}
The complete procedure is summarized in \cref{alg-multi-arm}.

\subsection{Extending the Regret Upper Bound}
In order to upper bound the regret of our algorithm in multi-armed setting, we need to extend \cref{lemma: decision-set-characterization,lemma: strong-density,lemma: eigenvalue,thm: good-prob,thm: regret} to the multi-armed setting.
The proofs for all results in this section are given in \cref{sec: proof-multi-arm}.

The performance of \cref{alg-multi-arm} again depends on the following two crucial events:
\begin{align*} \notag
\many_k  &= \bigg\{
		\min_{a \in \mathcal{A}}{N_{a, k}}\ge \prns{\frac{\log(T\delta^{-d})}{C_{\mathcal A}\epsilon_k^2}}^{\frac{2\beta +d}{2\beta}}
	\bigg\},\\\notag
	\good_{k}
		&= \bigg\{
		\abs{\hat{\eta}_{k, a}(x)-\eta_{k, a}(x)} \le \epsilon_{k}/2 \text{ for any } x \in \prns{\bigcup_{\mathcal A_0: a \in \mathcal A_0} \mathcal R_k(\mathcal A_0)} \cap \mathcal D_{\text{Irr}, k - 1}^C(a),  a \in \mathcal{A}_0, \text{ and any }\mathcal A_0 \subseteq \mathcal A \bigg\},
\end{align*}
where $\many_k$ represents that sufficiently \textit{many} samples for \emph{all} arms are available, and $\good_{k}$ represents \textit{good} accuracy of the estimated conditional reward function for \emph{every} arm uniformly over the region where it is estimable. We again define $\ogood_k = \bigcap_{1 \le j \le k}\good_k$ and $\omany_k = \bigcap_{1 \le j \le k}\many_k$, where an empty intersection ($\ogood_0$ or $\omany_0$) is the whole event space (always true).

The following lemma extends \cref{lemma: decision-set-characterization} and characterizes the decision regions of our algorithm in multi-armed setting, under the events above.
\begin{lemma}\label{lemma: decision-set-characterization-multi-arm}
  Suppose $T \ge T_0 \vee \prns{\exp(1 \vee \frac{C_{\mathcal A}(2\beta +d)}{4 (2r_0)^{2\beta}(2\beta+d + \beta d)})}$ and \cref{assump: multi-arm} Statement \ref{cond: multi-arm-regularity} holds.
If the event $\ogood_{k-1} \cap \omany_{k - 1}$ holds, then for any $\mathcal A_0 \subseteq \mathcal A$ and $a \in \mathcal A$,
\begin{enumerate}[label=\roman*.,ref=\roman*]
\item \label{decision-set-1} $D_{\text{Est}, k}(a) \cap \mathcal X \subseteq \braces{x \in \mathcal X: \eta_{a}(x) < \max_{a' \in \mathcal A}\eta_{a'}(x)}$;
\item \label{decision-set-2} $\mathcal R_k(\mathcal A_0) \cap \mathcal{X}  \subseteq \braces{x \in \mathcal X: \argmax_{a \in \mathcal A} \eta_{a}(x) \subseteq \mathcal A_0}$;
\item \label{decision-set-3} $\mathcal R_k(\mathcal A_0) \cap \mathcal{X}  \subseteq\braces{x \in \mathcal X: \max_{a \in \mathcal A_0} \eta_{a}(x) - \min_{a \in \mathcal A_0} \eta_{a}(x) \le 2\epsilon_{k - 1}}$;
\item \label{decision-set-4} $\mathcal Q_a \subseteq \bigcup_{\mathcal A_0: a \in \mathcal A_0} \mathcal R_k(\mathcal A_0) \cap \mathcal X$;
\item \label{decision-set-5} $D_{\text{Irr}, k}(a) \cap \mathcal X \subseteq \braces{x \in \mathcal X: \eta_{a}(x) < \max_{a' \in \mathcal A}\eta_{a'}(x)}$.
\end{enumerate}
\end{lemma}
Here statement \ref{decision-set-1} means that arms removed according to local polynomial estimates are strictly suboptimal.
Statement \ref{decision-set-2} implies that optimal arms on $\mathcal R_k(\mathcal A_0)$ are never removed by mistake, while statement \ref{decision-set-3} means that arms that are being randomized (including the optimal arms) have close expected rewards relative to the tolerance level $\epsilon_k$, which justifies our decision to explore them randomly. Statement \ref{decision-set-4} states that the support of sample for arm $a$ in the $k^{\text{th}}$ epoch
 (i.e., ${\bigcup}_{\mathcal A_0: a \in \mathcal A_0} \mathcal R_k(\mathcal A_0)$) always contains the region where arm $a$ is optimal. Statement \ref{decision-set-5} shows that
the arms  removed because of irregularity are also always strictly suboptimal.
Statements \ref{decision-set-1} and \ref{decision-set-5} justify the updating rule of removing arm $a$ in hypercubes in $D_{\text{Est}, k}(a)$ and $D_{\text{Irr},k}(a)$ (see \cref{eq: multi-arm-update}). These statements guarantee low regret of our algorithm provided $\omany_k$ and $\ogood_k$ hold for $k = 1, \dots, K$.

To prove that $\omany_k$ and $\ogood_k$ happen with high probability we next further extend
\cref{lemma: strong-density,lemma: eigenvalue} to multi-armed setting to ensure that local polynomial estimates are well-behaved.

\begin{lemma}\label{lemma: strong-density-multi-arm}
For any $1 \le k \le K$ and $a \in \mathcal A$,
$S_{a, k} = \{(X_t, Y_t): t \in \mathcal{T}_{k}, A_t = a\}$ are conditionally i.i.d. samples, given $\mathcal{F}_{k - 1}\cup\mathcal A_k$, where $\mathcal{F}_{k - 1} = \{(X_t, A_t, Y_t): t \in \bigcup_{k' = 1}^{k - 1}\mathcal{T}_{k'}\}$, $\mathcal A_k=\{A_t: t \in \mathcal{T}_k\}$.

Now suppose \cref{assump: multi-arm} holds
and
$T \ge T_0 \vee \prns{\exp(1 \vee \frac{C_{\mathcal A}(2\beta +d)}{4 (2r_0)^{2\beta}(2\beta+d + \beta d)})} $.
Then, for any $a \in \mathcal A$, under the event $\ogood_{k - 1}\cap \omany_{k - 1}$, the (common) conditional density of any of $\{X_t: A_t = a, t \in \mathcal{T}_k\}$ with respect to Lebesgue measure, given $\mathcal{F}_{k - 1}\cup\mathcal A_k$, which we denote by $\mu_{a, k}$,
satisfies the following conditions:
    \begin{enumerate}
        \item $\frac{1}{|\mathcal A|}\mu_{\min}\le \mu_{a,k}(x)\le \frac{|\mathcal A|\mu_{\max}}{p}$ for any
        $x \in {\bigcup}_{\mathcal A_0: a \in \mathcal{A}_0}\mathcal{R}_{k - 1}(\mathcal A_0) \cap \mathcal X$;
        \item $ \mu_{a,k}(x)=0$ for any $x \in {\bigcup}_{\mathcal A_0: a \not\in \mathcal{A}_0}\mathcal{R}_{k - 1}(\mathcal A_0) \cap \mathcal X$.
    \end{enumerate}
\end{lemma}

\begin{lemma}\label{lemma: eigenvalue-multi-arm}
Suppose the conditions of \cref{lemma: strong-density-multi-arm} hold.
Let $1\le k \le K-1,\,\,n_{\pm a, k}$ for $a \in \mathcal A$ be given.
Consider the Gram matrices of the local polynomial regression estimators in \cref{eq: cate-estimator}, \ie, $\hat{\mathcal{A}}(x; S_{a, k}, H_{a,k}, \floor*{\beta})$ as defined in \cref{eq: gram matrix}.
Then, for any $a \in \mathcal A$, given $N_{a,k}=n_{a,k}$ and $\omany_{k - 1} \cap \ogood_{k - 1}$, these satisfy the following with conditional probability at least
$1 - 2M_{\beta}^2\exp\big\{-C_{\mathcal A}\big(4(1+ L_1\sqrt{d})^2\big)n_{a, k}^{\nicefrac{2\beta}{(2\beta + d)}}\big\}$:
\[
    \lambda_{\min}(\hat{\mathcal{A}}(x; S_{a, k}, H_{a,k}, \floor*{\beta})) \ge \lambda_0 > 0, ~~~~ \forall x \in \prns{\bigcup_{\mathcal A_0: a \in \mathcal A_0} \mathcal R_k(\mathcal A_0)} \cap \mathcal D_{\mathrm{Irr}, k - 1}^C(a),
\]
 where
\begin{align*}
    \lambda_0 &=
    \frac{1}{4}\mu_{\min}
    \inf_{\substack{W \in \mathbb{R}^d,\, S \subset \mathbb{R}^d\,:\, \magd{W}=1 \\ S \subseteq \mathcal{B}(0,1) ~\text{is compact},~\leb(S)=  \nicefrac{c_0v_d}{2^d}}} \int_S \bigg(\sum_{|s|\le \floor*{\beta}}W_s u^s\bigg)^2d u.
\end{align*}
\end{lemma}

This allows us to next establish an analogue of \cref{thm: good-prob}, showing that $\omany_k$ and $\ogood_k$ happen with high probability.

\begin{theorem}\label{thm: good-prob-multi-arm}
    When
  $T \ge T_0 \vee \prns{\exp(1 \vee \frac{C_{\mathcal A}(2\beta +d)}{4 (2r_0)^{2\beta}(2\beta+d + \beta d)} \vee  \frac{36{M}_{\beta}L^2v^2_d\mu^2_{\max}C_{\mathcal A}(2\beta + d)}{p^2\lambda^2_0 (2\beta + d + \beta d)})}$,
  if we assume \cref{assump: multi-arm},
    then for any $1\le k\le K-1$,
    \begin{align*}
        &\qquad \pr (\good_{k}^C \mid \ogood_{k-1},  \omany_k )
            \le \frac{(4+2M_{\beta}^2)|\mathcal A|}{T}, \quad \pr(\many_k^C \mid \ogood_{k - 1}, \omany_{k-1})
            \le \frac{|\mathcal A|}{T},\quad \pr(\ogood_k^C \cup \omany_{k}^C) \le \frac{|\mathcal A|(5 + 2M_{\beta}^2)k}{T}.
    \end{align*}
\end{theorem}

\eedit\edit{
Finally, as a result, we can prove that the regret of our algorithm scales with the optimal rate $\Tilde{O}(T^{\frac{\beta+d-\alpha\beta}{2\beta+d}})$, and an additional factor growing with the number of arms $|\mathcal A|$.
\begin{theorem}\label{thm: regret-multi-arm}
Suppose
$T\ge T_0 \vee \prns{\exp(1 \vee C_{\mathcal{A}} \vee \frac{C_{\mathcal{A}}(2\beta +d)}{4 (2r_0)^{2\beta}(2\beta+d + \beta d)} \vee  \frac{36{M}_{\beta}L^2v^2_d\mu^2_{\max}C_{\mathcal{A}}(2\beta + d)}{p^2\lambda^2_0 (2\beta + d + \beta d)})}$ and \cref{assump: multi-arm} holds.
	Then
	\begin{align*}
		R_T(\pi)
        =\Tilde{O}(|\mathcal A|^3 T^{\frac{\beta+d-\alpha\beta}{2\beta+d}}),
	\end{align*}
    where the $\Tilde{O}(\cdot)$ term only depends on the parameters of \cref{assump: multi-arm}.
\end{theorem}}

\section[appendix]{Omitted Proofs}
\subsection[appendix]{Supporting Lemmas}

\begin{lemma}\label{lemma: epsilon-size}
	Under the assumptions in \cref{lemma: k-bound},
	\begin{align*}
	\epsilon_K^{-1} &= 2^K \le 2T^{\frac{\beta}{2\beta + d}}, \\
	\epsilon_k &\ge \frac{1}{2}\delta.
	\end{align*}
\end{lemma}
\proof{Proof.}
	\cref{lemma: k-bound} implies that
		\[
	2^K \le 2^{\frac{\beta}{(2\beta+d)\log 2}\log(T) + 1} = 2T^{\frac{\beta}{2\beta + d}}.
	\]
	It follows that for $1 \le k \le K$,
	\begin{align*}
		2^{-k} \ge 2^{-K} \ge \frac{1}{2}T^{-\frac{\beta}{2\beta + d}} \ge \frac{1}{2}\delta.
	\end{align*}
\endproof

\begin{lemma}\label{lemma: superset-regularity}
	For two sets $A \subseteq B$, and a point $x \in A$, if $A$ is $(c_0, r_0)$-regular at $x$, then $B$ is also $(c_0, r_0)$-regular at $x$.
\end{lemma}
\proof{Proof.}
	This is obvious according to the definition of $(c_0, r_0)$-regularity (Definition \ref{def: regularity}).
\endproof

\begin{lemma}\label{lemma: h-size2}
Given $\omany_{k}$, if
\edit{$T\ge \exp(1 \vee \frac{C_0(2\beta +d)}{4 (2r_0)^{2\beta}(2\beta+d + \beta d)}) $},
for $1 \le k' \le k \le K$.
\end{lemma}
\proof{Proof.}
	Given $\omany_{k}$, when \edit{$T\ge \exp(1 \vee \frac{C_0(2\beta +d)}{4 (2r_0)^{2\beta}(2\beta+d + \beta d)})$},
\[
\delta^{-d}T \ge \exp(\frac{C_0}{4 (2r_0)^{2\beta}}) \ge \exp(\epsilon_{k'}^2\frac{C}{ (2r_0)^{2\beta}})
\]
which implies that $H_{\pm 1, k'} = N_{\pm 1, k'}^{-\frac{1}{2\beta + d}} \le (\frac{\log(T\delta^{-d})}{C\epsilon_{k'}^2})^{-\frac{1}{2\beta}} \le 2r_0$.
\endproof

\begin{lemma}\label{lemma: delete-set}
	For any $1 \le k \le K$, if $H_{ 1, k} = N_{ 1, k}^{-\frac{1}{2\beta + d}} \in [\sqrt{d}\delta, 2r_0]$, then  $\big((\bigcup_{j=1}^{k} \mathcal{E}_{1,j})\cup  {\mathcal{R}}_{k}\big) \cap \mathcal{X}$ is not strongly $(c_0, r_0)$-regular at any $x \in D_{1, k} \cap \mathcal{X}$. Similarly, if $H_{- 1, k} = N_{- 1, k}^{-\frac{1}{2\beta + d}} \in [\sqrt{d}\delta, 2r_0]$, then $\big((\bigcup_{j=1}^{k} \mathcal{E}_{-1,j})\cup  {\mathcal{R}}_{k}\big) \cap \mathcal{X}$ is not strongly $(c_0, r_0)$-regular at any $x \in D_{-1, k} \cap \mathcal{X}$.
\end{lemma}
\proof{Proof.}
	We prove the first statement about $D_{1, k}$ by contradiction. Suppose there exists a point $x \in D_{1, k} \cap \mathcal{X}$ such that $\big((\bigcup_{j=1}^{k} \mathcal{E}_{1,j})\cup  {\mathcal{R}}_{k}\big) \cap \mathcal{X}$ is strongly $(c_0, r_0)$-regular at $x$. Since $\|g(x) - x\| \le \frac{1}{2}\sqrt{d}\delta \le \frac{1}{2}H_{1, k}$,
	\[
		\ball(x, \frac{H_{1, k}}{2}) \subseteq \ball(x, H_{1, k} -  \frac{1}{2}\sqrt{d}\delta) \subseteq \ball({g(x)}, H_{1, k}).
	\]
	Thus,
	\begin{align*}
		\leb[\mathcal{B}(g(x),H_{1,k})\cap (\bigcup_{j=1}^{k} \mathcal{E}_{1,j}\cup \mathcal{R}_k)]
			&\ge \leb[\mathcal{B}(x,\frac{H_{1, k}}{2})\cap (\bigcup_{j=1}^{k} \mathcal{E}_{1,j}\cup \mathcal{R}_k)] \\
			&\ge c_0v_d2^{-d}H_{1,k}^d = \frac{c_0}{2^d}\leb[\mathcal{B}(g(x),H_{1,k})],
	\end{align*}
	where $v_d$ is the volume of a unit ball in $\mathbb{R}^d$.  Here the second inequality uses the assumption that $\big((\bigcup_{j=1}^{k} \mathcal{E}_{1,j})\cup  {\mathcal{R}}_{k}\big) \cap \mathcal{X}$ is $(c_0, r_0)$-regular at $x$ and that $H_{a, k}/2 \le r_0$.

	This means that if there exists any point $x \in D_{1, k}$ such that $\big((\bigcup_{j=1}^{k} \mathcal{E}_{1,j})\cup  {\mathcal{R}}_{k}\big) \cap \mathcal{X}$  is $(c_0, r_0)$-regular at $x$, then $\big((\bigcup_{j=1}^{k} \mathcal{E}_{1,j})\cup  {\mathcal{R}}_{k}\big) \cap \mathcal{X}$  must be weakly $(\frac{c_0}{2^d}, H_{1, k})$-regular at $g(x)$, the center of the hypercube that $x$ belongs to.  By construction, $\big((\bigcup_{j=1}^{k} \mathcal{E}_{1,j})\cup  {\mathcal{R}}_{k}\big) \cap \mathcal{X}$  is not weakly $(\frac{c_0}{2^d}, H_{1, k})$-regular at any center of any hypercube in $D_{1, k}$ (\cref{eq: deletion-rule}). Thus the first statement about $D_{1, k}$ holds. The second one about $D_{-1, k}$ can be proved analogously.
\endproof

\begin{lemma}\label{lemma: deletion-well-defined}
Under the conditions in \cref{lemma: decision-set-characterization}, and suppose $H_{\pm 1, k}  \in [\sqrt{d}\delta, 2r_0]$, any hypercube in $\mathcal{C}$ at most belongs to  one of $\de_{1, k}$ and $\de_{-1, k}$.
\end{lemma}
\proof{Proof.}
We prove this by contradiction. Suppose there exists a $\cube \in \mathcal{C}$, such that for $(\bigcup_{j = 1}^{k}\exploit_{1, j} \cup \mathcal{R}_{k}) \cap \mathcal{X}$ is not weakly $(\frac{c_0}{2^d}, H_{1, k})$-regular at the center of $\cube$, and  $(\bigcup_{j = 1}^{k}\exploit_{-1, j} \cup \mathcal{R}_{k}) \cap \mathcal{X}$ is not weakly $(\frac{c_0}{2^d}, H_{-1, k})$-regular at the center of $\cube$ either.

However, since $\cube \cap \mathcal{X} \ne \emptyset$, there must exist $x \in \cube \cap \mathcal{X}$ such that either $\tau(x) \ge 0$ or $\tau(x) \le 0$. According to \cref{lemma: decision-set-characterization} statement  \ref{lemma: decision-set-characterization3} and \cref{lemma: superset-regularity}, either $(\bigcup_{j = 1}^{k - 1}\exploit_{1, j} \cup \mathcal{R}_{k - 1}) \cap \mathcal{X}$ or $(\bigcup_{j = 1}^{k - 1}\exploit_{-1, j} \cup \mathcal{R}_{k - 1}) \cap \mathcal{X}$ is $(c_0, r_0)$-regular at $x $. Without loss of generality, we suppose this for $(\bigcup_{j = 1}^{k - 1}\exploit_{1, j} \cup \mathcal{R}_{k - 1}) \cap \mathcal{X}$. Then the proof in \cref{lemma: delete-set} implies
that $(\bigcup_{j = 1}^{k - 1}\exploit_{1, j} \cup \mathcal{R}_{k - 1}) \cap \mathcal{X}$ must be $(\frac{c_0}{2^d}, H_{1, k})$-regular at the center of $\cube$. Thus contradiction arises.
\endproof

\begin{lemma}\label{lemma: convergence}
	For $\forall 1\le k \le K-1$, and integers $n_{\pm 1,k}$ that satisfy $n_{\pm 1,k}\ge (\frac{6\sqrt{M}L v_d\mu_{\max}}{p\lambda_0 \epsilon_k})^{\frac{2\beta+d}{\beta}}$, $n_{1,k}+n_{-1,k} = n_k$,  if we assume the Assumption \cref{assump: smoothness,assump: decision-set,assump: support} and that
  \edit{$T\ge T_0 \vee \prns{\exp(1 \vee \frac{C_0(2\beta +d)}{4 (2r_0)^{2\beta}(2\beta+d + \beta d)})} $},
  then the estimator $\hat{\tau}_k$ based on samples in the $k^\text{th}$ epoch satisfies that
	\begin{align*}
	&\mathbb{P}(\sup_{x\in \mathcal{R}_k \cap \de_{1, k}^C \cap \de^C_{-1, k}}|\hat{\tau}_k(x)-\tau(x)|\ge \epsilon_k \mid \ogood_{k-1}, \omany_{k - 1}, N_{\pm 1,k} = n_{\pm 1,k})\\ \le & \delta^{-d}{(8 + 4M_{\beta}^2)} \exp(-C_0 n_{a,k}^{\frac{2\beta}{2\beta+d}}\epsilon_k^2).
	\end{align*}
	where $C_0$ and $\lambda_0$ are given in \cref{lemma: eigenvalue}.
\end{lemma}
\proof{Proof.}
	In the following proof, we condition on $ \ogood_{k-1}, \omany_{k - 1}, N_{\pm 1,k} = n_{\pm 1,k}$, and $\mathcal{F}_{k - 1}$. According to \cref{lemma: strong-density}, the samples $S_{a, k} = \{(X_t, Y_t): A_t = a, t \in\mathcal{T}_k \}$ are i.i.d whose conditional density for $X_t$ is $\mu_{a, k}$: $\frac{1}{2}\mu_{\min}\le \mu_{a,k}(x)\le \frac{2\mu_{\max}}{p}$ for any $x \in \big((\bigcup_{j=1}^{k} \mathcal{E}_{a,j})\cup \mathcal{R}_k\big) \cap \mathcal{X}$, and $ \mu_{a,k}(x)=0$ for any $x \in \big(\bigcup_{j=1}^{k} \mathcal{E}_{-a,j}\big) \cap \mathcal{X}$. Moreover, recall that $\eta_a(x) = \expect[Y_t \mid X_t = x, A_t = a]$ for any $x \in \mathcal{X}$ and $a  = \pm 1$.  Here the purpose of conditioning on $\ogood_{k - 1}, \omany_{k - 1}, \mathcal{F}_{k - 1}$ is merely to guarantee the strong density condition for $\mu_{a, k}$ by \cref{lemma: strong-density}. In what follows, $K(x)$ denotes the kernel function $\ind(\|x\| \le 1)$.

	\textbf{Step I: Characterize the estimation error for a fixed point on the grid.} We first fix $x_0 \in G \cap \rand_k \cap D_{1, k}^C \cap D^C_{-1, k}$. To estimate the CATE, we first use local polynomial regression of order $\beta$ based on samples $S_{a, k} = \{(X_t, Y_t): A_t = a, t \in\mathcal{T}_k \}$ to estimate the conditional expected reward $\eta_{a}(x_0)$:
	\begin{align*}
		\hat{\eta}_{a,k}(x_0)
		&= e_1^T\bigg(\frac{1}{n_{a,k}h_{a,k}^d} \sum_{t\in \mathcal{T}_{a,k}}K(\frac{X_t-x_0}{h_{a,k}}) U(\frac{X_t-x_0}{h_{a,k}})U^T(\frac{X_t-x_0}{h_{a,k}})\bigg)^{-1} \\
		&\qquad \times  \bigg(\frac{1}{n_{a,k}h_{a,k}^d} \sum_{t\in \mathcal{T}_{a,k}}K(\frac{X_t-x_0}{h_{a,k}})U(\frac{X_t-x_0}{h_{a,k}})Y_t\bigg).
	\end{align*}
	where $U(u) = (u^r)_{|r| \le \beta} $ is a vector-valued function from $\mathbb{R}^d$ to $\mathbb{R}^M$, $h_{a,k} = n_{a,k}^{-\frac{1}{2\beta+d}}$ is the bandwidth, $e_1$ is a $M \times 1$ vector whose all elements are $0$ except the first one. Recall that $M_{\beta} = |\{r: |r| \le \floor*{\beta}\}|$.

	According to Proposition 1.12 in \cite{Tsybakov:2008:INE:1522486}, the true conditional expected reward $\eta_{a}$ can be written in the following way:
	\begin{align*}
		\eta_a(x_0) = \eta_{a, \floor*{\beta}}(x_0;x_0)
			&= e_1^T\bigg(\frac{1}{n_{a,k}h_{a,k}^d}\sum_{t\in \mathcal{T}_{a,k}}K(\frac{X_t-x_0}{h_{a,k}}) U(\frac{X_t-x_0}{h_{a,k}})U^T(\frac{X_t-x_0}{h_{a,k}})\bigg)^{-1} \\
			&\qquad \times  \bigg(\frac{1}{n_{a,k}h_{a,k}^d} \sum_{t\in \mathcal{T}_{a,k}}K(\frac{X_t-x_0}{h_{a,k}})U(\frac{X_t-x_0}{h_{a,k}})\eta_{a, \floor*{\beta}}(X_t;x_0)\bigg).
	\end{align*}
	where
	\[
		\eta_{a, \floor*{\beta}}(x;x_0) = \sum_{|r| \le \floor*{\beta}} \frac{(x - x_0)^r}{r!}D^r \eta_a(x_0).
	\]

	After denoting
	\[
		\hat{\mathcal{A}}_{a,k}(x_0) = \frac{1}{n_{a,k}h_{a,k}^d} \sum_{t\in \mathcal{T}_{a,k}}K(\frac{X_t-x_0}{h_{a,k}}) U(\frac{X_t-x_0}{h_{a,k}})U^T(\frac{X_t-x_0}{h_{a,k}}),
	\]
	the estimation error for $\hat{\eta}_{a,k}(x_0)$ has the following upper bound:
	\begin{align*}
		\vert \hat{\eta}_{a,k}(x_0)  - \eta_a(x_0) \vert
			&\le \bigg\|\bigg(\frac{1}{n_{a,k}h_{a,k}^d} \sum_{t\in \mathcal{T}_{a,k}}K(\frac{X_t-x_0}{h_{a,k}}) U(\frac{X_t-x_0}{h_{a,k}})U^T(\frac{X_t-x_0}{h_{a,k}})\bigg)^{-1}\bigg\| \\
			&\times \bigg\|\frac{1}{n_{a,k}h_{a,k}^d} \sum_{t\in \mathcal{T}_{a,k}}K(\frac{X_t-x_0}{h_{a,k}})U(\frac{X_t-x_0}{h_{a,k}})(Y_t - \eta_{a, \floor*{\beta}}(X_t;x_0))\bigg\| \\
			&\le \frac{\sqrt{M_{\beta}}}{\lambda_{\min}(\hat{\mathcal{A}}_{a,k}(x_0))}\bigg\vert \frac{1}{n_{a,k}h_{a,k}^d} \sum_{t\in \mathcal{T}_{a,k}} \big(Y_t - \eta_{a, \floor*{\beta}}(X_t;x_0))K(\frac{X_t-x_0}{h_{a,k}})\big) \bigg\vert \\
			&\le \frac{\sqrt{M_{\beta}}}{\lambda_{\min}(\hat{\mathcal{A}}_{a,k}(x_0))}\bigg\vert \frac{1}{n_{a,k}h_{a,k}^d} \sum_{t\in \mathcal{T}_{a,k}} \big(Y_t - \eta_{a}(X_t)\big)K(\frac{X_t-x_0}{h_{a,k}}) \bigg\vert \\
			&+ \frac{\sqrt{M_{\beta}}}{\lambda_{\min}(\hat{\mathcal{A}}_{a,k}(x_0))}\bigg\vert \frac{1}{n_{a,k}h_{a,k}^d} \sum_{t\in \mathcal{T}_{a,k}} \big(\eta_{a}(X_t) - \eta_{a, \floor*{\beta}}(X_t;x_0)\big)K(\frac{X_t-x_0}{h_{a,k}}) \bigg\vert,
	\end{align*}
	where the second inequality follows from the fact that  $U(\frac{X_t-x_0}{h_{a,k}})K(\frac{X_t-x_0}{h_{a,k}})$ is a $M_{\beta} \times 1$ vector whose elements are bounded by $K(\frac{X_t-x_0}{h_{a,k}})$.

	We further denote
	\begin{align*}
		\Gamma_1
			&= \frac{1}{n_{a,k}h_{a,k}^d} \sum_{t\in \mathcal{T}_{a,k}} \big(Y_t - \eta_{a}(X_t)\big)K(\frac{X_t-x_0}{h_{a,k}}) \\
		\Gamma_2
			&= \frac{1}{n_{a,k}h_{a,k}^d} \sum_{t\in \mathcal{T}_{a,k}} \big(\eta_{a}(X_t) - \eta_{a, \floor*{\beta}}(X_t;x_0)\big)K(\frac{X_t-x_0}{h_{a,k}})
	\end{align*}
	Then
	\begin{align}\label{eq: est-error}
		\vert \hat{\eta}_{a,k}(x_0)  - \eta_a(x_0) \vert  \le \frac{\sqrt{M_\beta}}{\lambda_{\min}(\hat{\mathcal{A}}_{a,k}(x_0))}(\Gamma_1 + \Gamma_2)
	\end{align}

	\textbf{Step II:  lower bound for $\lambda_{\min}(\hat{\mathcal{A}}_{a,k}(x_0))$.}

	According to \cref{lemma: eigenvalue},  with high probability $1 - 2M_{\beta}^2\exp\big\{-C_0\big(4(1+ L_1\sqrt{d})^2\big)n_{a, k}^{\frac{2\beta}{2\beta + d}}\big\}$,
	\begin{align}\label{eq: eigen}
		\lambda_{\min}(\hat{\mathcal{A}}_{a,k}(x_0)) \ge \lambda_0.
	\end{align}

	\textbf{Step III:  upper bound $\Gamma_1$ and $\Gamma_2$.}

	We first bound
		\begin{align*}
	\Gamma_1
	&= \frac{1}{n_{a,k}h_{a,k}^d} \sum_{t\in \mathcal{T}_{a,k}} \big(Y_t - \eta_{a}(X_t)\big)K(\frac{X_t-x_0}{h_{a,k}}) = \frac{1}{n_{a,k}} \sum_{t\in \mathcal{T}_{a,k}} \hat{Z}_t
	\end{align*}
	where
	\[
		\hat{Z}_t = \frac{1}{h_{a,k}^d}(Y_t-\eta_a(X_i))K(\frac{X_t-x_0}{h_{a,k}}).
	\]
	It is easy to prove that
	\begin{align*}
		|\hat{Z}_t| &\le h_{a,k}^{-d}, \\
		\expect( \hat{Z}_t \mid A_t = a, \ogood_{k-1}, \omany_{k - 1},  &N_{a,k} = n_{a,k}, \mathcal{F}_{k-1})
			= 0, \\
		\expect(\hat{Z}_t^2  \mid A_t = a, \ogood_{k-1}, \omany_{k - 1},  &N_{a,k} = n_{a,k}, \mathcal{F}_{k-1})
			\le \frac{2v_d\mu_{\max}}{ph_{a,k}^d},
	\end{align*}
	where the last inequality uses the fact that $|Y_t-\eta_a(X_i)| \le 1$.
	By Bernstein's inequality, for $\forall \epsilon>0$,
	\begin{align}\label{eq: gamma-1}
	&\mathbb{P}(|\Gamma_1|\ge \epsilon \mid \ogood_{k-1}, \omany_{k - 1},  N_{a,k} = n_{a,k}, \mathcal{F}_{k-1}) \nonumber \\
	=&\mathbb{P}(|\sum_{t \in \mathcal{T}_{a,k}}\hat{Z}_t|\ge n_{a,k}\epsilon \mid \ogood_{k-1}, \omany_{k - 1},  N_{a,k} = n_{a,k}, \mathcal{F}_{k-1}) \nonumber\\
	\le & 2\exp(-\frac{3p n_{a,k}h_{a,k}^d\epsilon^2}{12v_d\mu_{\max}+2p\epsilon}).
	\end{align}

	We now bound
	\begin{align*}
		\Gamma_2
			&= \frac{1}{n_{a,k}h_{a,k}^d} \sum_{t\in \mathcal{T}_{a,k}} \big(\eta_{a}(X_t) - \eta_{a, \floor*{\beta}}(X_t;x_0)\big)K(\frac{X_t-x_0}{h_{a,k}})
			=  \frac{1}{n_{a, k}}\sum_{t\in \mathcal{T}_{a,k}}\Tilde{Z}_t,
	\end{align*}
	where
	\[
		\Tilde{Z}_t = \frac{1}{h_{a,k}^d}(\eta_a(X_t)-\eta_{a, \floor*{\beta}}(X_t;x_0))K(\frac{X_t-x_0}{h_{a,k}})
	\]
	Note that by the definition of H\"older class (Assumption \ref{assump: smoothness}),
	\[
		\bigg\vert \big(\eta_{a}(X_t) - \eta_{a, \floor*{\beta}}(X_t;x_0)\big)K(\frac{X_t-x_0}{h_{a,k}}) \bigg\vert \le  LK(\frac{X_t-x_0}{h_{a,k}})  \|X_t - x_0\|^\beta \le Lh_{a, k}^\beta.
	\]
	It follows that
	\begin{align*}
		|\expect(\Tilde{Z}_t|   A_t = a, \ogood_{k-1},\omany_{k - 1},  N_{a,k} = n_{a,k}, \mathcal{F}_{k-1})|  &\le \frac{2}{p}L v_d\mu_{\max}h_{a,k}^{\beta} \\
		|\Tilde{Z}_t-\expect( \Tilde{Z}_t| A_t = a, \ogood_{k-1},\omany_{k - 1},  N_{a,k} = n_{a,k}, \mathcal{F}_{k-1})| &\le \frac{2}{p}L v_d\mu_{\max}h_{a,k}^{\beta} + L h_{a,k}^{\beta-d} \\
		\expect(\Tilde{Z}_t^2|  A_t = a, \ogood_{k-1}, \omany_{k - 1},  N_{a,k} = n_{a,k}, \mathcal{F}_{k-1}) &\le \frac{2}{p}L^2 v_d\mu_{\max}h_{a,k}^{2\beta-d}.
	\end{align*}
	By Bernstein's inequality, for $\forall \epsilon > 0$,
	\begin{align} \label{eq: gamma-2}
	&\mathbb{P}(\vert \Gamma_2 \vert  \ge \epsilon + \frac{2}{p}L v_d\mu_{\max}h_{a,k}^{\beta} \mid  \ogood_{k-1}, \omany_{k - 1}, N_{a,k} = n_{a,k}, \mathcal{F}_{k-1}) \nonumber \\
	= & \mathbb{P}(\vert \frac{1}{n_{a,k}} \sum_{t \in \mathcal{T}_{a,k}}\Tilde{Z}_t\vert  \ge \epsilon + \frac{2}{p}L v_d\mu_{\max}h_{a,k}^{\beta} \mid  \ogood_{k-1}, \omany_{k - 1},  N_{a,k} = n_{a,k}, \mathcal{F}_{k-1}) \nonumber \\
	\le &  \mathbb{P}(\frac{1}{n_{a,k}}|\sum_{t \in \mathcal{T}_{a,k}}\Tilde{Z}_t- \expect(\Tilde{Z}_t \mid \ogood_{k-1}, N_{a,k} = n_{a,k}, \mathcal{F}_{k-1})|\ge \epsilon \mid \ogood_{k-1}, \omany_{k - 1},  N_{a,k} = n_{a,k}, \mathcal{F}_{k-1}) \nonumber \\
	 \le &2\exp(-\frac{3p n_{a,k}h_{a,k}^d\epsilon^2}{12L^2v_d\mu_{\max}h_{a,k}^{2\beta}+2L(2v_d\mu_{\max}+p)h_{a,k}^{\beta}\epsilon}).
	\end{align}

	\textbf{Step IV: error bound for a fixed point}

	Plug \cref{eq: eigen,eq: gamma-1,eq: gamma-2} with $\epsilon = \frac{\lambda_0}{3\sqrt{M_{\beta}}}\epsilon_k$ into \cref{eq: est-error}, we can get that
	\begin{align*}
		&\pr\bigg(\big\vert \hat{\eta}_{a,k}(x_0)  - \eta_a(x_0) \big\vert  \ge \frac{\sqrt{M_{\beta}}}{\lambda_0}(\frac{2\lambda_0}{3\sqrt{M_{\beta}}}\epsilon_k + \frac{2}{p}L v_d\mu_{\max}h_{a,k}^{\beta}) \mid \ogood_{k-1}, \omany_{k - 1},  N_{a,k} = n_{a,k}, \mathcal{F}_{k-1} \bigg) \\
		\le& 2M^2_{\beta}\exp(-\frac{3p n_{a,k}h_{a,k}^d\lambda_0^2}{12M^4_{\beta}\mu_{\max}v_d+2p\lambda_0 M^2_{\beta}}) +  2\exp(-\frac{p n_{a,k}h_{a,k}^d\lambda_0^2\epsilon_k^2}{36 Mv_d\mu_{\max}+2\sqrt{M_{\beta}}p\epsilon_k\lambda_0}) \\
		+&  2\exp(-\frac{p n_{a,k}h_{a,k}^d\lambda_0^2\epsilon_k^2}{36 M_\beta L^2v_d\mu_{\max}h_{a,k}^{2\beta}+2\sqrt{M_{\beta}}L(2v_d\mu_{\max}+p)h_{a,k}^{\beta}\epsilon_k\lambda_0})
	\end{align*}

	Note that $\frac{2}{p}L v_d\mu_{\max}h_{a,k}^{\beta} \le \frac{\lambda_0}{3\sqrt{M_{\beta}}}\epsilon_k$ under the assumption that $n_{\pm 1,k}\ge (\frac{6\sqrt{M_{\beta}}L v_d\mu_{\max}}{p\lambda_0 \epsilon_k})^{\frac{2\beta+d}{\beta}}$. In other words,
	\[
		\frac{\sqrt{M_{\beta}}}{\lambda_0}(\frac{2\lambda_0}{3\sqrt{M_{\beta}}}\epsilon_k + \frac{2}{p}L v_d\mu_{\max}h_{a,k}^{\beta}) \le \epsilon_k.
	\]
	After denoting
	\begin{align*}
	C = \min\bigg\{\frac{3p \lambda_0^2}{12M^4_{\beta}\mu_{\max}v_d+2p\lambda_0 M^2_{\beta}}, \frac{p \lambda_0^2}{36 M_{\beta}v_d\mu_{\max}+2\sqrt{M_{\beta}}p\lambda_0}, \\
	\frac{p \lambda_0^2}{36 M_{\beta}L^2v_d\mu_{\max} + 2\sqrt{M_{\beta}}L(2v_d\mu_{\max}+p)\lambda_0}\bigg\},
	\end{align*}
	it is easy to verify that
	\begin{align*}
		&2M^2_{\beta}\exp\big\{-C_0\big(4(1+ L_1\sqrt{d})^2\big)n_{a, k}^{\frac{2\beta}{2\beta + d}}\big\} \le  2M^2_{\beta}\exp(-Cn_{a, k}h_{a, k}^d\epsilon_k^2), \\
		&\exp(-\frac{3p n_{a,k}h_{a,k}^d\lambda_0^2\epsilon_k^2}{108 M_{\beta}v_d\mu_{\max}+6\sqrt{M_{\beta}}p\epsilon_k\lambda_0}) \le \exp(-Cn_{a, k}h_{a, k}^d\epsilon_k^2), \\
		&\exp(-\frac{3p n_{a,k}h_{a,k}^d\lambda_0^2\epsilon_k^2}{108 M_{\beta}L^2v_d\mu_{\max}h_{a,k}^{2\beta}+6\sqrt{M_{\beta}}L(2v_d\mu_{\max}+p)h_{a,k}^{\beta}\epsilon_k\lambda_0}) \\
		&\le \exp(-Cn_{a, k}h_{a, k}^{d - \beta}\epsilon_k^2)
		\le \exp(-Cn_{a, k}h_{a, k}^d\epsilon_k^2).
	\end{align*}
	Therefore,
	\begin{align*}
		\pr\bigg(\big\vert \hat{\eta}_{a,k}(x_0)  - \eta_a(x_0) \big\vert  \ge \epsilon_k\mid \ogood_{k-1}, \omany_{k - 1},  N_{a,k} = n_{a,k}, \mathcal{F}_{k-1} \bigg)
			&\le (2M^2_{\beta} + 4)\exp(-Cn_{a, k}h_{a, k}^d\epsilon_k^2) \\
			&= (4+2M^2_{\beta})\exp(-C n_{a,k}^{\frac{2\beta}{2\beta+d}}\epsilon_k^2)
	\end{align*}
	Moreover, marginalizing over the history $\mathcal{F}_{k-1}$ gives
	\begin{align*}
	\pr\bigg(\big\vert \hat{\eta}_{a,k}(x_0)  - \eta_a(x_0) \big\vert  \ge \epsilon_k\mid \ogood_{k-1}, \omany_{k - 1},  N_{a,k} = n_{a,k} \bigg) \le   (4+2M^2_{\beta})\exp(-C n_{a,k}^{\frac{2\beta}{2\beta+d}}\epsilon_k^2)
	\end{align*}

	\textbf{Step V:  uniform convergence}

	Since $|G \cap \rand_k \cap D_{1, k}^C \cap D^C_{-1, k}| \le |G| = \delta^{-d}$, we can take union bound:
	\begin{align*}
	&\mathbb{P}(\sup_{x\in G \cap \rand_k \cap D_{1, k}^C \cap D^C_{-1, k}}|\hat{\eta}_{a,k}(x)-\eta_a(x)|\ge \epsilon_k| \ogood_{k-1}, \omany_{k - 1},  N_{a,k} = n_{a,k})\\
		\le &\delta^{-d}\mathbb{P}(|\hat{\eta}_{a,k}(x_0)-\eta_a(x_0)|\ge \epsilon_k| \ogood_{k-1}, \omany_{k - 1},  N_{a,k} = n_{a,k})\\
		\le & \delta^{-d}{(4+2M_{\beta}^2)} \exp(-C n_{a,k}^{\frac{2\beta}{2\beta+d}}\epsilon_k^2).
	\end{align*}

	Moreover, since $\eta_a$ is Lipschitz, we know that for any $x \in \mathcal{X}$,
	\begin{align*}
		|\hat{\eta}_a(x) - \eta_a(x)|
			&= |\hat{\eta}_a(g(x)) - \eta_a(x)|  \\
			&\le |\hat{\eta}_a(g(x)) - \eta_a(g(x))|  + | \eta_a(g(x)) - \eta_a(x)| \\
			&\le |\hat{\eta}_a(g(x)) - \eta_a(g(x))|  + L_1 \|g(x) - x\| \\
			&\le |\hat{\eta}_a(g(x)) - \eta_a(g(x))|  + \frac{1}{2}L_1 \sqrt{d}\delta.
	\end{align*}
	Therefore,
	\begin{align*}
	&\mathbb{P}(\sup_{x\in  \rand_k \cap D_{1, k}^C \cap D^C_{-1, k}}|\hat{\eta}_{a,k}(x)-\eta_a(x)|\ge (1+ L_1\sqrt{d})\epsilon_k| \ogood_{k-1}, \omany_{k - 1}, N_{a,k} = n_{a,k}) \\
	\le & \mathbb{P}(\sup_{x\in  \rand_k \cap D_{1, k}^C \cap D^C_{-1, k}}|\hat{\eta}_{a,k}(x)-\eta_a(x)|\ge \epsilon_k +\frac{1}{2} L_1\sqrt{d}\delta \mid \ogood_{k-1}, \omany_{k - 1},N_{a,k} = n_{a,k}) \\
	\le &  \mathbb{P}(\sup_{x\in G \cap \rand_k \cap D_{1, k}^C \cap D^C_{-1, k}}|\hat{\eta}_{a,k}(x)-\eta_a(x)|\ge \epsilon_k| \ogood_{k-1}, \omany_{k - 1}, N_{a,k} = n_{a,k})   \\
	\le & \delta^{-d}{(4+2M_\beta^2)} \exp(-C n_{a,k}^{\frac{2\beta}{2\beta+d}}\epsilon_k^2).
	\end{align*}
	where the first inequality uses the fact that $\epsilon_k = 2^{-k} \ge \frac{1}{2}\delta $ for $1 \le k \le K$ (\cref{lemma: epsilon-size}).

	Taking union bound over $a = \pm 1$ and replacing $(1 + L_1\sqrt{d})\epsilon_k$ with $\epsilon_k$ gives the final conclusion.
\endproof

\subsection[appendix]{Proofs for \cref{section: formulation}}

\proof{Proof of \cref{lemma: nomempty-opt}.}
By assumption there exists some $x^* \in \mathcal{Q}_1$. Then
\begin{align*}
\pr(X\in\mathcal{Q}_1)
    &\ge \mu_{\min}\leb(\mathcal{Q}_1) \\
    &\ge \mu_{\min}\leb(\mathcal{Q}_1\cap \mathcal{B}(x^*, r_0)) \\
    &\ge  \mu_{\min}c_0\leb(\mathcal{B}(x^*, r_0)) \\
    &= \mu_{\min}c_0r_0^dv_d.
\end{align*}
Analogously, we can prove the same result for $\pr(\mathcal{Q}_{-1})$.
\endproof

{\blockedit
\proof{Proof of \cref{prop: margin sharp}}

If $\mathcal{X}_{\ne 0}:=\{x\in\mathcal X: |\tau(x)|>0\}$ is empty, the statement is trivially true.
In what follows, we assume $\mathcal{X}_{\ne 0} \ne \emptyset$.

We first show that $\mathcal{X}_{\ne 0}$ is closed by contradiction.
Consider a sequence $x_1,x_2,\cdots\in \mathcal{X}_{\ne 0}$ with limit point $\lim_{n\rightarrow \infty}x_n=x \notin \mathcal{X}_{\ne 0}$, i.e., $\tau(x)=0$.
By \cref{assump: smoothness}, $\abs{\tau(x)}$ is $2L_1$-Lipschitz, so for every $n\in \mathbb{N}_+$,
\begin{align*}
  \pr(0<|\tau(X)|\le  |2\tau(x_n)|) &\ge  \pr(|\tau(X) - \tau(x_n)|\le |\tau(x_n)|) \\
  &\ge  \pr\prns{\|X-x_n\| \le \frac{|\tau(x_n)|}{2 L_1} \wedge r_0}\\
  &\ge  c_0 \mu_{\min} v_d \prns{\frac{|\tau(x_n)|}{2 L_1} \wedge r_0}^d.
\end{align*}
By \cref{assump: margin},
\begin{align*}
\pr(0<|\tau(X)|\le 2|\tau(x_n)|) \le \gamma (2|\tau(x_n)|)^{\alpha},
\end{align*}
and we conclude that
$$\gamma (2|\tau(x_n)|)^{\alpha}\ge c_0\mu_{\min} v_d \prns{\frac{|\tau(x_n)|}{2 L_1} \wedge r_0}^d.$$
By continuity $\lim_{n\rightarrow \infty}|\tau(x_n)|=|\tau(x)| = 0$, but this implies $\alpha \le d$, which contradicts our assumption that $\alpha>d$.
Thus, $\mathcal{X}_{\ne 0}$ is closed.

Because $\tau(x)$ is continuous and $\mathcal{X}_{\ne 0}$ compact, there exists $x_0\in\mathcal{X}_{\ne 0}$ such that $\abs{\tau(x)}\geq\abs{\tau(x_0)}>0$ for all $x\in\mathcal{X}_{\ne 0}$. We next show that $t_0=\abs{\tau(x_0)}$ is lower bounded by a constant depending only on our assumption parameters.

By \cref{assump: smoothness}, $\abs{\tau(x)}$ is $2L_1$-Lipschitz. Hence,
for $t\geq t_0$,
\begin{align*}
\pr(0<\abs{\tau(X)}\leq t)
&  =\pr(0<\abs{\tau(X)}-\abs{\tau(x_0)}\leq t-t_0)\\
&\geq\pr\prns{\|X-x_0\|\leq\frac{t-t_0}{2L_1}\wedge r_0}\\
&\geq c_0\mu_{\min}v_d\prns{\frac{t-t_0}{2L_1}\wedge r_0}^d.
\end{align*}
By \cref{assump: margin}, for any $t\geq0$,
$$
\pr(0<\abs{\tau(X)}\leq t)\leq\gamma t^\alpha.
$$
We conclude that for $t\in[t_0,\,t_0+2L_1r_0]$ we have
$$
t_0\geq t-\kappa t^{\alpha/d}, ~~ \kappa=\frac{2L_1\gamma^{1/d} }{(c_0\mu_{\min}v_d)^{1/d}}
$$
Since the inequality also trivially holds for $t\in[0,\,t_0]$ and $2L_1r_0\leq t_0+2L_1r_0$, this inequality holds for $t\in[0,\,2L_1r_0]$.
Therefore, $t_0 \ge \tau_{\min} \coloneqq \max_{0 \le t \le 2L_1r_0} (t - \kappa t^{\alpha/d})$.
Since $\alpha>d$, $\tau_{\min}$ is positive and only depends on $\kappa,\alpha/d,2L_1r_0$.
\endproof}

\subsection[appendix]{Proofs for Section \ref{section: algorithm}}

\proof{Proof of \cref{lemma: k-bound}.}
	By definition, $K$ is the smallest integer such that $\sum_{j=1}^K n_j \ge T$.

	Note that if $T\ge e^{C_0}$, for an integer $K_0$,
	\[
	\sum_{j=1}^{K_0} n_j
	\ge \sum_{j=1}^{K_0} \frac{4}{p} (\frac{\log T}{C_0})^{\frac{2\beta+d}{2\beta}}2^{\frac{2\beta+d}{\beta}j}
	\ge \frac{4}{p} (2^{K_0\frac{2\beta+d}{\beta}} - 1).
	\]
	If $K_0$ is the smallest integer such that $\frac{4}{p} (2^{K_0\frac{2\beta+d}{\beta}} - 1) \ge T$, then by the definition of $K$, we have
	\[
	K \le K_0 = \lceil\frac{\beta}{(2\beta+d)\log 2}\log(\frac{pT}{4} + 1)\rceil \le \lceil\frac{\beta}{(2\beta+d)\log 2}\log(T)\rceil.
	\]
\endproof

\proof{Proof of \cref{lemma: h-size}.}
Note that $H_{\pm 1, k} = (\frac{1}{N_{\pm 1, k}})^{\frac{1}{2\beta + d}} \ge (\frac{1}{n_k})^{\frac{1}{2\beta + d}} \ge (\frac{1}{n_K})^{\frac{1}{2\beta + d}}$, so we only need to prove the statement for $(\frac{1}{n_K})^{\frac{1}{2\beta + d}}$:
\begin{align*}
\frac{\frac{1}{n_K}}{(\sqrt{d}\delta )^{2\beta + d}}
	&= \frac{1}{{d}^{\frac{2\beta + d}{2}}}\frac{\delta ^{-(2\beta + d)}}{ \frac{4}{p}(\frac{\log(T\delta^{-d})}{C_0\epsilon_K^2})^{\frac{2\beta +d}{2\beta}}+\frac{2}{p^2}\log T} \\
	&\ge \frac{1}{{d}^{\frac{2\beta + d}{2}}}\frac{T^\beta (\log T)^{2\beta + d}}{\frac{4}{p}[\frac{2\beta + d + \beta d}{(2\beta + d)C_0}]^{\frac{2\beta +d}{2\beta}}\log(T)^{\frac{2\beta +d}{2\beta}}T + \frac{2}{p^2}\log T} \\
	&=  \frac{1}{{d}^{\frac{2\beta + d}{2}}}\frac{T^{\beta - 1}(\log T)^{\frac{(2\beta - 1)(2\beta + d)}{2\beta}}}{\frac{4}{p}[\frac{2\beta + d + \beta d}{(2\beta + d)C_0}]^{\frac{2\beta +d}{2\beta}} +  \frac{2}{p^2}\frac{1}{T(\log T)^{\frac{d}{2\beta}}}}.
\end{align*}
Thus there exists $c_1 > 0$ such that
\[
\frac{H_{1, k} }{\sqrt{d}\delta}  \ge c_1 T^{\beta - 1}\log(T)^{\frac{(2\beta - 1)(2\beta + d)}{2\beta}}.
\]
Since $c_0 T^{\beta - 1}\log(T)^{\frac{(2\beta - 1)(2\beta + d)}{2\beta}} \to \infty$ when $T \to \infty$ and $\beta \ge 1$, so there exists $T_0$ such that $\frac{H_{1, k} }{\sqrt{d}\delta} \ge 1$ for $T \ge T_0$.
\endproof

\proof{Proof of \cref{lemma: union-cube}.}
	We prove this by induction. When $k = 1$, this is trivially true because $\exploit_{\pm 1, 1} = \mathcal{D}_{\pm 1, 1} = \emptyset$ and $\rand_1$ is the union of all hypercubes in $\mathcal{T}$. Suppose that this statement is also true for $1 \le k \le k_0$.

	For $k = k_0 + 1$ and $a = \pm 1$,
	\begin{align*}
		\rand_k
			&= \big\{x\in \mathcal{R}_{k-1} \cap \mathcal{D}_{1, k - 1}^C \cap \mathcal{D}_{-1, k - 1}^C: |\hat{\tau}_{k-1}(x)| \le \epsilon_{k-1}\big\}, \\
		\mathcal{E}_{a,k}
			&= \big\{x\in \mathcal{R}_{k-1} \cap \mathcal{D}_{1, k - 1}^C \cap \mathcal{D}_{-1, k - 1}^C: a\hat{\tau}_{k-1}(x)> \epsilon_{k-1} \big\}  \cup \mathcal{D}_{-a, k - 1}.
	\end{align*}
	Obviously $\mathcal{R}_k$ can be written as unions of hypercubes in $\mathcal{C}$ because $\mathcal{R}_{k-1} \cap \mathcal{D}_{1, k - 1}^C \cap \mathcal{D}_{-1, k - 1}^C$ can be written as unions of hypercubes according to the induction assumption, and $\hat{\tau}_{k - 1}$ is constant within each hypercube in $\mathcal{C}$ (\cref{eq: cate-estimator}). Similarly, $\mathcal{E}_{a,k}$ can be written as unions of hypercubes in $\mathcal{C}$.

	Moreover, by the definition of $\de_{a, k}$, it can be also written  as unions of hypercubes in $\mathcal{C}$ (\cref{eq: deletion-rule}).
\endproof

\subsection[appendix]{Proofs for \cref{section: ub}}
\proof{Proof of \cref{lemma: decision-set-characterization}.}
We will prove all statements for $a = 1$, and those for $a = -1$ can be proved analogously. We prove the statements by induction. For $k=1$, $\bigcup_{j=1}^k\mathcal{E}_{1,j} = \emptyset\subseteq\{x \in \mathcal{X}:\tau(x)>0\}$, $\bigcup_{j=1}^k\mathcal{E}_{-1,j}= \emptyset\subseteq\{x \in \mathcal{X}:\tau(x)<0\}$, $\rand_{k} \cap \mathcal{X} = \mathcal{X}$, $\{x \in \mathcal{X}: a\tau(x) \le 2\epsilon_{k - 1}\} = \mathcal{X}$, and $\de_{\pm1 , k} = \emptyset$ since $\bigcup_{j=1}^k\mathcal{E}_{\pm1,j}\cup \rand_k = \mathcal{X}$ is strongly $(c_0, r_0)$ regular at any $x \in \mathcal{X}$ according to assumption \ref{assump: decision-set}. So statements 1-4 hold for $k = 1$.

Assume that statements 1-4 hold for $k\le k_0$. We only need to prove that the statements also holds for $k_0 + 1$.

\textbf{Statement 1. } Statement 1 follows from the following fact: under the event $\ogood_{k_0}$,
\begin{align*}
\rand_{k_0 + 1}\cap \mathcal{X}
	&=  \big\{x\in \mathcal{R}_{k_0} \cap \mathcal{D}_{1, k_0}^C  \cap \mathcal{D}_{-1, k_0}^C \cap \mathcal{X}: |\hat{\tau}_{k_0}(x)| \le \epsilon_{k_0}\big\} \\
	&\subseteq \{x\in \mathcal{R}_{k_0} \cap \mathcal{X}: | \tau(x)|\le 2\epsilon_{k_0}\}.
\end{align*}

\textbf{Statement 2. } According to the decision updating rule \cref{eq: exploit},
\begin{align*}
\big(\bigcup_{j=1}^{k_0+1} \mathcal{E}_{1,j}\big) \cap \mathcal{X}
&= \bigg(\big(\bigcup_{j=1}^{k_0} \mathcal{E}_{1,j} \big) \cap \mathcal{X} \bigg) \cup \{x\in \mathcal{R}_{k_0} \cap D_{1, k_0}^C \cap D^C_{-1, k_0} \cap \mathcal{X}: \\
&\qquad\qquad\qquad\qquad\qquad\qquad \hat{\tau}_{k_0 }(x)>\epsilon_{k_0 }\}\cup \big(\mathcal{D}_{-1,k_0} \cap \mathcal{X}).
\end{align*}
According to induction assumption, $\big(\bigcup_{j=1}^{k_0} \mathcal{E}_{1,j} \big) \cap \mathcal{X}  \subseteq\{x \in \mathcal{X}: \tau(x) > 0\}$ and $D_{-1, k_0} \cap \mathcal{X} \subseteq \{x \in \mathcal{X}: \tau(x) > 0\}$. $\ogood_{k_0}$ implies that $\{x\in \mathcal{R}_{k_0} \cap D_{1, k_0}^C \cap D^C_{-1, k_0} \cap \mathcal{X}: \hat{\tau}_{k_0}(x)>\epsilon_{k_0}\}\subseteq\{x \in \mathcal{X}: \tau(x) > 0\}$.

\textbf{Statement 3.}
For $k\ge 2$,  according to statement 2,
$$  ((\bigcup_{j=1}^{k} \mathcal{E}_{1,j})\cup \mathcal{R}_k)^C \cap \mathcal{X} = \bigcup_{j=1}^{k} \mathcal{E}_{-1,j}\cap \mathcal{X} \subseteq \{x \in \mathcal{X}:\tau(x)<0\},$$
which implies $ \{x \in \mathcal{X}:\tau(x)\ge 0\} \subseteq \big((\bigcup_{j=1}^{k} \mathcal{E}_{1,j})\cup \mathcal{R}_k\big) \cap \mathcal{X}$.

\textbf{Statement 4.} We prove $D_{-1, k_0 + 1} \cap \mathcal{X} \subseteq \{x \in \mathcal{X}: \tau(x) > 0\}$ by showing that for any $x \in \mathcal{X}$ such that $\tau(x) \le 0$, $x \not\in D_{-1, k_0 + 1}$. Recall that $D_{-1, k_0 + 1} \subseteq \mathcal{R}_{k_0 + 1} \subseteq \bigcup_{j = 1}^{k_0 + 1}\mathcal{E}_{-1, j} \cup \mathcal{R}_{k_0 + 1}$.  According to statement 2 and statement 3,
\[
\bigg((\bigcup_{j=1}^{k_0 + 1} \mathcal{E}_{-1,j})\cup  {\mathcal{R}}_{k_0 + 1}\bigg)^C \cap \mathcal{X} = \big(\bigcup_{j=1}^{k_0 + 1} \mathcal{E}_{1,j}\big) \cap \mathcal{X} \subseteq \{x \in \mathcal{X}: \tau(x) > 0\},
\]
which implies that
\[
\{x \in \mathcal{X}:\tau(x)\le 0\}\subseteq \bigg((\bigcup_{j=1}^{k_0 + 1} \mathcal{E}_{-1,j})\cup  {\mathcal{R}}_{k_0 + 1}\bigg) \cap \mathcal{X}.
\]
If there exists any point $x\in \mathcal{X}$ such that $\tau(x) \le 0$, then by Assumption  \ref{assump: decision-set} and \cref{lemma: superset-regularity}, $\bigg((\bigcup_{j=1}^{k_0 + 1} \mathcal{E}_{-1,j})\cup  {\mathcal{R}}_{k_0 + 1}\bigg)\cap \mathcal{X}$ is strongly $(c_0, r_0)$-regular at $x$.
Since
\edit{$T\ge T_0 \vee \prns{\exp(1 \vee \frac{C_0(2\beta +d)}{4 (2r_0)^{2\beta}(2\beta+d + \beta d)})} $},
\cref{lemma: h-size} and \cref{lemma: h-size2} implies that $H_{\pm 1, k'} = N_{\pm 1, k'}^{-\frac{1}{2\beta + d}} \in [\sqrt{d}\delta, 2r_0]$ for $1 \le k' \le k_0$. Thus we can use \cref{lemma: delete-set}, which implies that $x \not\in D_{-1, k_0 + 1}$.  Therefore, $D_{-1, k_0 + 1} \cap \mathcal{X} \subseteq \{x \in \mathcal{X}: \tau(x) > 0\}$.

\endproof

\proof{Proof of \cref{lemma: strong-density}.}
	When $k  = 1$, the conclusions hold trivially since $\exploit_{\pm 1} = \emptyset, R_1 = \mathcal{X}$, \ie, we pull each arm with prob. $1/2$ for all samples in the first stage. This implies that $\mu_{\pm1, 1} = \mu(x)$ for $x \in \mathcal{X}$. Then the conclusion follows from Assumption \ref{assump: support}. Denote $\mathcal{T}_{a, k} = \{t \in \mathcal{T}_k: A_t = a\}$.

	When $k \ge 2$, $\{(X_t, Y_t), t\in \mathcal{T}_{a, k}\}$ are obviously i.i.d conditionally on $\mathcal{F}_{k-1}$, $\ogood_{k - 1}$ and $\omany_{k-1}$, since $A_t$ only depends on $X_t$ and $\mathcal{F}_{k - 1}$. This implies that $\{X_t:  t\in \mathcal{T}_{a, k}\}$ are i.i.d conditionally on $\mathcal{F}_{k-1}$, $\ogood_{k - 1}$, $\omany_{k - 1}$. Furthermore, $X_t {\perp\!\!\!\perp} N_{\pm 1,k}|A_t$, thus $\{X_t:  t\in \mathcal{T}_{a, k}\}$ are i.i.d given $\mathcal{F}_{k-1}$, $\ogood_{k - 1}$, $\omany_{k - 1}$, $N_{\pm 1, k} = n_{\pm 1, k}$. This also implies that for any $x \in \mathcal{X}$,
	\begin{align*}
		\mu_{a,k}(x)
			&= \mu_{X_t \mid A_t=a, \ogood_{k-1}, \omany_{k-1}, N_{\pm 1, k} = n_{\pm 1, k}, \mathcal{F}_{k-1}}(x) \\
			&= \mu_{X_t \mid A_t=a, \ogood_{k-1}, \omany_{k-1}, \mathcal{F}_{k-1}}(x)
	\end{align*}

	For any $x\in \mathcal{X},$ obviously $\mu_{X_t| \ogood_{k-1}, \omany_{k - 1}, \mathcal{F}_{k-1}}(x) = \mu_{X_t}(x)$ for $t \in \mathcal{T}_k$.  Thus
	\begin{align}\label{eq: density-bound}
		&\mu_{X_t|A_t=a, \ogood_{k-1}, \omany_{k - 1}, \mathcal{F}_{k-1}}(x) \nonumber \\
		& = \frac{\mathbb{P}(A_t=a \mid X_t=x, \ogood_{k-1}, \omany_{k-1}, \mathcal{F}_{k-1})}{\mathbb{P}(A_t=a \mid \ogood_{k-1},\omany_{k - 1}, \mathcal{F}_{k-1})}\mu_{X_t \mid \ogood_{k-1}, \omany_{k -1}, \mathcal{F}_{k-1}}(x) \nonumber \\ & = \frac{\mathbb{P}(A_t=a \mid X_t=x, \ogood_{k-1}, \omany_{k -1}, \mathcal{F}_{k-1})}{\mathbb{P}(A_t=a \mid \ogood_{k-1}, \omany_{k -1}, \mathcal{F}_{k-1})}\mu_{X_t}(x).
	\end{align}

	For $\forall x \in \big(\bigcup_{j=1}^{k} \mathcal{E}_{-a,j}\big) \cap \mathcal{X}$, our algorithm ensures that
	\[
		\mathbb{P}(A_t=a|X_t=x, \ogood_{k-1}, \omany_{k - 1},\mathcal{F}_{k-1})=0.
	\]
	Therefore, for $\forall x \in \big(\bigcup_{j=1}^{k} \mathcal{E}_{-a,j}\big) \cap \mathcal{X}$, $\mu_{a,k}(x)=0$, which proves statement 2.

	For any $x\in \big((\bigcup_{j=1}^{k} \mathcal{E}_{a,j})\cup \mathcal{R}_k\big) \cap \mathcal{X}$, our algorithm ensures that
	\[
		 \mathbb{P}(A_t=a|X_t=x, \ogood_{k-1}, \omany_{k - 1}, \mathcal{F}_{k-1}) \in \{\frac{1}{2}, 1\}.
	\]
	Plus, since
  \edit{
  $T\ge T_0 \vee \prns{\exp(1 \vee \frac{C_0(2\beta +d)}{4 (2r_0)^{2\beta}(2\beta+d + \beta d)})} $,}
  \cref{lemma: decision-set-characterization} implies that $\{x:a\tau(x)\ge 0\}\subseteq \big((\bigcup_{j=1}^{k} \mathcal{E}_{a,j})\cup \mathcal{R}_k\big) \cap \mathcal{X}$. By Assumption \ref{assump: decision-set},
	\begin{align*}
	&\mathbb{P}(A_t=a| \ogood_{k-1}, \omany_{k - 1}, \mathcal{F}_{k-1})\\
	= & \mathbb{P}(A_t=a|X_t\in \big((\bigcup_{j=1}^{k} \mathcal{E}_{a,j})\cup \mathcal{R}_k\big) \cap \mathcal{X}, \ogood_{k-1}, \omany_{k - 1}, \mathcal{F}_{k-1})  \\
	& \times \mathbb{P}(X_t \in \big((\bigcup_{j=1}^{k} \mathcal{E}_{a,j})\cup \mathcal{R}_k\big)\cap\mathcal{X} \mid \ogood_{k-1}, \omany_{k - 1}, \mathcal{F}_{k-1})
	 \\ \ge & \frac{1}{2}\pr(a\tau(x)\ge 0) \ge \frac{p}{2},
	\end{align*}
	where the last inequality follows from \cref{lemma: nomempty-opt}.

	Then it follows from \cref{assump: support,eq: density-bound} that for $\forall x\in \big((\bigcup_{j=1}^{k} \mathcal{E}_{a,j})\cup \mathcal{R}_k\big) \cap \mathcal{X}$,
	\[
		\frac{1}{2}\mu_{\min}\le \mu_{a,k}(x)\le \frac{2\mu_{\max}}{p},
	\]
	which proves statement 1.
\endproof

\proof{Proof of \cref{lemma: eigenvalue}.}
	In the following proof, we condition on $ \ogood_{k-1}, \omany_{k - 1}, N_{\pm 1,k} = n_{\pm 1,k}$, and $\mathcal{F}_{k - 1}$. According to \cref{lemma: strong-density}, the samples $S_{a, k} = \{(X_t, Y_t): A_t = a, t \in\mathcal{T}_k \} = \{(X_t, Y_t): t \in\mathcal{T}_{k, a} \}$ are i.i.d whose conditional density for $X_t$ is $\mu_{a, k}$: $\frac{1}{2}\mu_{\min}\le \mu_{a,k}(x)\le \frac{2\mu_{\max}}{p}$ for any $x \in \big((\bigcup_{j=1}^{k} \mathcal{E}_{a,j})\cup \mathcal{R}_k\big) \cap \mathcal{X}$, and $ \mu_{a,k}(x)=0$ for any $x \in \big(\bigcup_{j=1}^{k} \mathcal{E}_{-a,j}\big) \cap \mathcal{X}$. Moreover, recall that $\eta_a(x) = \expect[Y_t \mid X_t = x, A_t = a]$ for any $x \in \mathcal{X}$ and $a  = \pm 1$.  Here the purpose of conditioning on $\ogood_{k - 1}, \omany_{k - 1}, \mathcal{F}_{k - 1}$ is merely to guarantee the strong density condition for $\mu_{a, k}$. We fix $x_0 \in G \cap \rand_k \cap D_{1, k}^C \cap D^C_{-1, k}$.

	Recall that
	\[
		\hat{\mathcal{A}}_{a,k}(x_0; S_{a, k}) = \frac{1}{n_{a,k}h_{a,k}^d} \sum_{t\in \mathcal{T}_{a,k}}K(\frac{X_t-x_0}{h_{a,k}}) U(\frac{X_t-x_0}{h_{a,k}})U^T(\frac{X_t-x_0}{h_{a,k}}).
	\]
	where $U(u) = (u^r)_{|r| \le \beta} $ is a vector-valued function from $\mathbb{R}^d$ to $\mathbb{R}^M$, $h_{a,k} = n_{a,k}^{-\frac{1}{2\beta+d}}$ is the bandwidth, $e_1$ is a $M \times 1$ vector whose all elements are $0$ except the first one.

	Note that the $(r_1, r_2)$-th entry of $\hat{\mathcal{A}}_{a,k}(x_0)$ is
	\[
		(\hat{\mathcal{A}}_{a,k}(x_0))_{r_1,r_2} =\frac{1}{n_{a,k}h_{a,k}^d}\sum_{t \in \mathcal{T}_{a,k}}(\frac{X_t-x_0}{h_{a,k}})^{r_1+r_2}K(\frac{X_t-x_0}{h_{a,k}})
	\]
	whose conditional expectation is
	\begin{align*}
		&\qquad\qquad \expect\big[(\hat{\mathcal{A}}_{a,k}(x_0))_{r_1,r_2} \mid \ogood_{k-1}, \omany_{k - 1}, N_{a,k} = n_{a,k}, \mathcal{F}_{k-1}\big]  \\
			& =\frac{1}{h_{a,k}^d} \int_{\|\frac{x-x_0}{h_{a,k}}\| \le 1, x \in (\bigcup_{j=1}^{k} \mathcal{E}_{a,j})\cup \mathcal{R}_k} (\frac{x-x_0}{h_{a,k}})^{r_1+r_2} \mu_{a, k}(x)dx \\
			&= \int_{||u||\le 1, x_0+u h_{a,k}\in (\bigcup_{j=1}^{k} \mathcal{E}_{a,j})\cup \mathcal{R}_k}u^{s_1+s_2}\mu_{a,k}(x_0+h_{a,k}u)d u \\
			& \coloneqq (\mathcal{A}_{a,k}(x_0))_{r_1,r_2}
	\end{align*}
	It follows that
	\begin{align}\label{eq: eigen-diff}
		\lambda_{\min}(\hat{\mathcal{A}}_{a,k}(x_0)) &\ge \min_{||W||=1}W^T \mathcal{A}_{a,k}(x_0) W + \min_{||W||=1} W^T(\hat{\mathcal{A}}_{a,k}(x_0)-\mathcal{A}_{a,k}(x_0))W \nonumber \\
	& \ge \min_{||W||=1}W^T \mathcal{A}_{a,k}(x_0) W - \sum_{|r_1|,|r_2|\le \floor*{\beta}}  |(\hat{\mathcal{A}}_{a,k}(x_0))_{r_1,r_2}-(\mathcal{A}_{a,k}(x_0))_{r_1,r_2}|.
	\end{align}
	We first derive lower bound for $\min_{||W||=1}W^T \mathcal{A}_{a,k}(x_0) W$:
	\begin{align*}
		W^T \mathcal{A}_{a,k}(x_0) W
			&= \int_{B_{a,k}(x_0)}  \bigg(\sum_{|r|\le \floor*{\beta}}W_ru^r\bigg)^2\mu_{a,k}(x_0+h_{a,k}u)du \\
			&\ge \frac{1}{2}\mu_{\min}\int_{B_{a,k}(x_0)} \bigg(\sum_{|r|\le \floor*{\beta}}W_ru^r\bigg)^2du,
	\end{align*}
	where $B_{a,k}(x_0) = \{u\in \mathbb{R}^d: ||u||\le 1, x_0+u h_{a,k}\in \big((\bigcup_{j=1}^{k} \mathcal{E}_{a,j})\cup \mathcal{R}_k\big) \cap \mathcal{X}\}$.

	Note that
	\begin{align*}
		\leb[B_{a,k}(x_0)]
			&= h_{a,k}^{-d}\leb[\mathcal{B}(x_0,h_{a,k})\cap \big(\big((\bigcup_{j=1}^{k} \mathcal{E}_{a,j})\cup \mathcal{R}_k\big) \cap \mathcal{X}\big)] \\
			&\ge h_{a,k}^{-d}\frac{c_0}{2^d}\leb[\mathcal{B}(x_0,h_{a,k})] = h_{a,k}^{-d}\frac{c_0}{2^d}v_dh^d_{a,k} = \frac{c_0}{2^d}v_d
	\end{align*}
	where the first equality holds because of change of variable, and the inequality holds since $x_0 \not\in D_{\pm 1, k}$ and our construction of $D_{\pm 1, k}$ guarantees that $(\bigcup_{j=1}^{k} \mathcal{E}_{a,j})\cup \mathcal{R}_k$ is weakly $(\frac{c_0}{2^d}, h_{a, k})$-regular at $x_0 \in G_K$.

	Thus
	\begin{align}\label{eq: eigen-pop}
			&\lambda_{\min}( \mathcal{A}_{a,k}(x_0)) =  \min_{\|W\| = 1} W^T \mathcal{A}_{a,k}(x_0) W \nonumber \\
			\ge&  \frac{1}{2}\mu_{\min}\min_{||W||=1;S \subseteq \mathcal{B}(0,1), \leb(S)= \frac{c_0}{2^d}v_d}\int_S \bigg(\sum_{|s|\le \floor*{\beta}}W_s u^s\bigg)^2d u = 2\lambda_0.
	\end{align}

	Now we derive upper bound for $\sum_{|r_1|,|r_2|\le \floor*{\beta}}  |(\hat{\mathcal{A}}_{a,k}(x_0))_{r_1,r_2}-(\mathcal{A}_{a,k}(x_0))_{r_1,r_2}|$.

	For $t \in S_{a, k}$ and $|r_1|, |r_2| \le \beta$, denote $Z_t(r_1, r_2) =  \frac{1}{h_{a,k}^d}(\frac{X_t-x_0}{h_{a,k}})^{r_1+r_2}K(\frac{X_t-x_0}{h_{a,k}})$. Obviously, $|Z_t(r_1, r_2)|\le \frac{1}{h^d_{a,k}}$, and
	\begin{align*}
	&\qquad \expect( Z_t^2(r_1, r_2)| A_t = a, \ogood_{k-1}, \omany_{k - 1}, N_{a,k} = n_{a,k}, \mathcal{F}_{k-1})   \\
		&= \frac{1}{h_{a,k}^{2d}}\int_{\mathcal{X}}(\frac{x-x_0}{h_{a,k}})^{2(r_1+r_2)}K^2(\frac{x-x_0}{h_{a,k}})\mu_{a,k} (x)dx\\
		& \le \frac{2\mu_{\max}}{p h_{a,k}^d}\int_{B_{a,k}(x_0)}u^{2(r_1+r_2)}du
		\le \frac{2\mu_{\max}}{p h_{a,k}^d}\leb[B_{a,k}(x_0)] \le  \frac{2\mu_{\max}v_d}{p h_{a,k}^d},
	\end{align*}
	where the first inequality follows from the fact that $\mu_{a, k}(x) \le \frac{2\mu_{\max}}{p}$.

	By Bernstein inequality,
	\begin{align*}
	&\mathbb{P}(|(\hat{\mathcal{A}}_{a,k}(x_0))_{r_1,r_2}-(\mathcal{A}_{a,k}(x_0))_{r_1,r_2}|\ge \frac{\lambda_0}{M_{\beta}^2} \mid  \ogood_{k-1}, \omany_{k - 1},  N_{a,k} = n_{a,k}, \mathcal{F}_{k-1} ) \\
	=& \mathbb{P}\bigg(\big\vert \sum_{i\in \mathcal{T}_{a,k}}Z_t(r_1, r_2) -  n_{a,k}\expect(\sum_{i\in \mathcal{T}_{a,k}}Z_t(r_1, r_2) \mid \ogood_{k-1}, \omany_{k - 1}, N_{a,k} = n_{a,k}, \mathcal{F}_{k-1}) \big\vert \\
	& \qquad\qquad \qquad \qquad \qquad \qquad  \ge n_{a,k}\frac{\lambda_0}{M_{\beta}^2} \mid \ogood_{k-1},  \omany_{k - 1}, N_{a,k} = n_{a,k}, \mathcal{F}_{k-1}\bigg) \\
	\le& 2\exp(-\frac{3p n_{a,k}h_{a,k}^d\lambda_0^2}{12M_{\beta}^4\mu_{\max}v_d+2p\lambda_0 M^2_\beta}).
	\end{align*}
	By taking union bound over all possible $r_1, r_2$,
	\begin{align}\label{eq: eigen-sample}
	&\mathbb{P}(\sum_{|r_1|,|r_2|\le \floor*{\beta}}  |(\hat{\mathcal{A}}_{a,k}(x_0))_{r_1,r_2}-(\mathcal{A}_{a,k}(x_0))_{r_1,r_2}|\ge \lambda_0 | \ogood_{k-1}, \omany_{k - 1},  N_{a,k} = n_{a,k}, \mathcal{F}_{k-1}) \nonumber\\
	\le & 2M^2_{\beta}\exp(-\frac{3p n_{a,k}h_{a,k}^d\lambda_0^2}{12M^4_{\beta}\mu_{\max}v_d+2p\lambda_0 M^2_{\beta}}) \nonumber \\
	\le & 2M^2_{\beta}\exp\big\{-C_0\big(4(1+ L_1\sqrt{d})^2\big)n_{a, k}h_{a, k}^d\big\}  \nonumber \\
	\le & 2M^2_{\beta}\exp\big\{-C_0\big(4(1+ L_1\sqrt{d})^2\big)n_{a, k}^{\frac{2\beta}{2\beta + d}}\big\}
	\end{align}

	According to \cref{eq: eigen-diff,eq: eigen-pop,eq: eigen-sample}, with high probability $1 - 2M^2_{\beta}\exp\big\{-C_0\big(4(1+ L_1\sqrt{d})^2\big)n_{a, k}^{\frac{2\beta}{2\beta + d}}\big\}$,
	\[
	\lambda_{\min}(\hat{\mathcal{A}}_{a,k}(x_0)) \ge \lambda_0.
	\]
\endproof

\proof{Proof of \cref{thm: good-prob}.}

	\textbf{I. Proof for $\pr (\good_{k}^C \mid \ogood_{k-1},  \omany_k )
			\le \frac{8+4M^2_{\beta}}{T}$.} When
      \edit{$T\ge \exp(1 \vee \frac{36{M}_{\beta}L^2v^2_d\mu^2_{\max}C_0(2\beta + d)}{p^2\lambda^2_0 (2\beta + d + \beta d)}) $},
	\begin{align*}
		\delta^{-d}T \ge \exp(\frac{36{M}_{\beta}L^2v^2_d\mu^2_{\max}C_0}{p^2\lambda^2_0}),
	\end{align*}
	which implies that $N_{\pm 1, k} \ge  (\frac{\log(T\delta^{-d})}{C_0\epsilon_k^2})^{\frac{2\beta +d}{2\beta}} \ge (\frac{6\sqrt{M_{\beta}}L v_d\mu_{\max}}{p\lambda_0 \epsilon_k})^{\frac{2\beta+d}{\beta}}$ when $\many_k $ holds.

	So all conditions in \cref{lemma: convergence} are satisfied. Thus the conclusion follows from the fact that
	\[
	\delta^{-d}{(8+4M^2_{\beta})} \exp(-C_0 \min\{N_{-1,k}, N_{1,k}\}^{\frac{2\beta}{2\beta+d}}\epsilon_k^2) \le \frac{8+4M^2_{\beta}}{T}.
	\]
	since  $\many_k $ states that
	\[
	\min\{N_{1, k}, N_{-1, k}\}\ge (\frac{\log(T\delta^{-d})}{C_0\epsilon_k^2})^{\frac{2\beta +d}{2\beta}}.
	\]

	\textbf{II. Proof for $\pr(\many_k^C \mid \ogood_{k - 1}, \omany_{k-1})
			\le \frac{2}{T}$.}
	\cref{lemma: decision-set-characterization} implies that given $\ogood_{k-1} \cap \omany_{k - 1}$,  $\{x:\tau(x)\ge 0\}\subseteq\bigcup_{j=1}^k\mathcal{E}_{1,j}\cup \mathcal{R}_k$. Thus
	\begin{align*}
	\expect(N_{1,k} \mid \ogood_{k-1}, \omany_{k - 1}) &= \expect( \sum_{t\in \mathcal{T}_k}\mathbb{I}\{X_t\in \bigcup_{j=1}^k\mathcal{E}_{1,j}\} + \sum_{t\in \mathcal{T}_k}\mathbb{I}\{X_t\in \mathcal{R}_k, A_t = 1\} \mid \ogood_{k-1}, \omany_{k - 1})\\
	& = n_k\mathbb{P}(X_t\in \bigcup_{j=1}^k\mathcal{E}_{1,j} \mid \ogood_{k-1}, \omany_{k - 1}) + \frac{1}{2}n_k\mathbb{P}(X_t\in \mathcal{R}_k \mid \ogood_{k-1}, \omany_{k - 1})\\
	&\ge \frac{1}{2}n_k\mathbb{P}(X_t\in \bigcup_{j=1}^k\mathcal{E}_{1,j}\cup \mathcal{R}_k \mid \ogood_{k-1}, \omany_{k - 1})\\
	&\ge \frac{1}{2}n_k\mathbb{P}(\tau(X_t) \ge 0 \mid \ogood_{k - 1}, \omany_{k - 1}) \ge \frac{p}{2}n_k,
	\end{align*}
	where the last inequality uses Assumption \ref{assump: decision-set}.

		By Hoeffding's inequality,
	\begin{align*}
	&\mathbb{P}\bigg(N_{1,k}< (\frac{\log(T\delta^{-d})}{C_0\epsilon_k^2})^{\frac{2\beta +d}{2\beta}}\mid \ogood_{k-1}, \omany_{k - 1}\bigg) \\
	\le & \mathbb{P}\bigg(\expect(N_{1,k}\mid \ogood_{k-1}, \omany_{k - 1})-N_{1,k}>\frac{p}{2}n_k-(\frac{\log(T\delta^{-d})}{C_0\epsilon_k^2})^{\frac{2\beta +d}{2\beta}}\mid \ogood_{k-1}, \omany_{k - 1}\bigg)\\
	\le & \exp\bigg(-\frac{2}{n_k}\big[\frac{p}{2}n_k-(\frac{\log(T\delta^{-d})}{C_0\epsilon_k^2})^{\frac{2\beta +d}{2\beta}}\big]^2\bigg).
	\end{align*}

	When $n_k \ge \frac{4}{p}(\frac{\log(T\delta^{-d})}{C_0\epsilon_k^2})^{\frac{2\beta +d}{2\beta}}+\frac{2}{p^2}\log T$
	\[
		\frac{2}{n_k}\big[\frac{p}{2}n_k-(\frac{\log(T\delta^{-d})}{C_0\epsilon_k^2})^{\frac{2\beta +d}{2\beta}}\big]^2 \ge \frac{p^2}{2}n_k - 2p(\frac{\log(T\delta^{-d})}{C_0\epsilon_k^2})^{\frac{2\beta +d}{2\beta}} \ge \log T.
	\]

	Thus
	\[
		\mathbb{P}\bigg(N_{1,k}< (\frac{\log(T\delta^{-d})}{C_0\epsilon_k^2})^{\frac{2\beta +d}{2\beta}}\mid \ogood_{k-1}, \omany_{k - 1}\bigg) \le \frac{1}{T}.
	\]

	Similarly, we can prove the result  for $N_{-1, k}$. Then the conclusion follows by taking union bound for $N_{-1, k}$ and $N_{1, k}$.

	\textbf{III. Proof for $\pr(\ogood_k^C \cup \omany_{k}^C) \le \frac{(10 + 4M^2)k}{T}$.} Recall that $\ogood_k = \cap_{j = 1}^k \good_j$ and $\omany_k = \cap_{j = 1}^k \many_j$.  It is easy to verify that
	\[
		\ogood_k^C \cup \omany_{k}^C \subseteq \big(\bigcup_{j = 0}^{k - 1} \good_{j + 1}^C \cap \omany_{j + 1} \cap \ogood_{j} \big)\cup \big(\bigcup_{j = 0}^{k - 1}\many_{j + 1}^C \cap \ogood_j \cap \omany_j \big).
	\]
	It follows that
	\begin{align*}
		\pr(\ogood_k^C \cup \omany_{k}^C)
			&\le \sum_{j = 0}^{k - 1}\pr\big( \good_{j + 1}^C \cap \omany_{j + 1} \cap \ogood_{j} \big) +  \sum_{j = 0}^{k - 1}\pr\big(\many_{j + 1}^C \cap \ogood_j \cap \omany_j \big) \\
			&\le \sum_{j = 0}^{k - 1}\pr( \good_{j + 1}^C \mid \ogood_{j}, \omany_{j + 1}) +  \sum_{j = 0}^{k - 1}\pr(\many_{j + 1}^C \mid \ogood_j, \omany_j) \\
			&\le \sum_{j = 0}^{k - 1} \frac{8 + 4M_{\beta}^2}{T} + \frac{2}{T} = \frac{(10 + 4M_{\beta}^2)k}{T}.
	\end{align*}
 \endproof

\proof{Proof of \cref{thm: regret}.}
	According to the definition of the expected cumulative regret,
	\begin{align*}
		R_T(\hat{\pi})
			&= \expect (\sum_{t=1}^{T} Y_{t}(\pi^*(X_t)) - Y_{t}(A_t)) \\
			&=  \sum_{k=1}^K \sum_{t\in\mathcal{T}_k} \expect (Y_{t}(\pi^*(X_t)) - Y_{t}(A_t)) \\
			&\le \sum_{k=1}^K\sum_{t\in\mathcal{T}_k} \expect( Y_{t}(\pi^*(X_t)) - Y_{t}(A_t) \mid \ogood_{k - 1} \cap \omany_{k - 1}) \\
			&+ \sum_{k=1}^K\sum_{t\in\mathcal{T}_k} \expect( Y_{t}(\pi^*(X_t)) - Y_{t}(A_t) \mid \ogood_{k - 1}^C \cup \omany_{k - 1}^C)\pr(\ogood_{k - 1}^C \cup \omany_{k - 1}^C) \\
			&\le \sum_{k=1}^K\sum_{t\in\mathcal{T}_k} \expect( Y_{t}(\pi^*(X_t)) - Y_{t}(A_t) \mid \ogood_{k - 1} \cap \omany_{k - 1})  \\
			&+ \sum_{k=1}^K\sum_{t\in\mathcal{T}_k} \pr(\ogood_{k - 1}^C \cup \omany_{k - 1}^C)
	\end{align*}
	where the last inequality follows from the fact that $Y_t(\pm1) \in [0, 1]$.

	\cref{thm: good-prob} states that for $2 \le k \le K$,
	\[
		\sum_{t\in\mathcal{T}_k}\pr(\ogood_{k - 1}^C \cup \omany_{k - 1}^C) \le n_k \frac{(10 + 4M_{\beta}^2)(k - 1)}{T} \le (10 + 4M_{\beta}^2)(k - 1).
	\]
	Furthermore,
	\begin{align*}
		&\sum_{t\in\mathcal{T}_k} \expect( Y_{t}(\pi^*(X_t)) - Y_{t}(A_t) \mid \ogood_{k - 1} \cap \omany_{k - 1}) \\
		=& \sum_{t\in\mathcal{T}_k} \expect( Y_{t}(\pi^*(X_t)) - Y_{t}(A_t) \mid \ogood_{k - 1} \cap \omany_{k - 1}, X_t \in \rand_k)\pr(X_t \in \rand_k) \\
		+& \sum_{t\in\mathcal{T}_k} \expect( Y_{t}(\pi^*(X_t)) - Y_{t}(A_t) \mid \ogood_{k - 1} \cap \omany_{k - 1}, X_t \in \bigcup_{j = 1}^k \exploit_{1, j}\cup \exploit_{-1, j} ) \\
		&\times \pr(X_t \in\bigcup_{j = 1}^k \exploit_{1, j}\cup \exploit_{-1, j}).
	\end{align*}
	\cref{lemma: decision-set-characterization} implies that given $\ogood_{k - 1}$, $\bigcup_{j=1}^{k} \mathcal{E}_{1,j}\subseteq \{x:\tau(x)> 0\}$ and $\bigcup_{j=1}^{k} \mathcal{E}_{-1,j}\subseteq \{x:\tau(x)< 0\}$. This means that the decisions made on these regions exactly coincide with the decisions made by the oracle policy. Therefore,
	\[
		\sum_{t\in\mathcal{T}_k} \expect( Y_{t}(\pi^*(X_t)) - Y_{t}(A_t) \mid \ogood_{k - 1} \cap \omany_{k - 1}, X_t \in \bigcup_{j = 1}^k \exploit_{1, j}\cup \exploit_{-1, j} )  = 0.
	\]
	Moreover, \cref{lemma: decision-set-characterization} states that given  $\ogood_{k - 1}$, $\rand_k\subseteq \{x: |\tau(x)| \le  2\epsilon_{k-1}\}$.
	\begin{align*}
	&\sum_{t\in\mathcal{T}_k} \expect( Y_{t}(\pi^*(X_t)) - Y_{t}(A_t) \mid \ogood_{k - 1} \cap \omany_{k - 1}, X_t \in \rand_k)\pr(X_t \in \rand_k \mid \ogood_{k - 1} \cap \omany_{k - 1})  \\
	= &\sum_{t\in\mathcal{T}_k}\expect ( |\tau(X_t)| \mid \ogood_{k - 1} \cap \omany_{k - 1}, X_t\in \mathcal{R}_k) \mathbb{P}(X_t\in \mathcal{R}_k \mid \ogood_{k - 1} \cap \omany_{k - 1})   \\
	= &\sum_{t\in\mathcal{T}_k} 2\epsilon_{k-1}\mathbb{P}(|\tau(X_t)|\le 2\epsilon_{k-1})\\
	\le & \MarginL2^{1+\alpha}\epsilon_{k-1}^{1+\alpha} n_k,
	\end{align*}
	where the last inequality uses the margin condition in Assumption \ref{assump: margin}.

	Moreover, according to \cref{lemma: k-bound},
	\begin{align*}
		K &\le \lceil \frac{\beta}{(2\beta+d)\log 2}\log T \rceil.
	\end{align*}
	Thus,
	\begin{align*}
		R_T(\hat{\pi})
			&\le \sum_{k=1}^K \MarginL2^{1+\alpha}\epsilon_{k-1}^{1+\alpha} n_k + (10 + 4M^2)(k - 1) \\
			&\le \frac{\MarginL}{p}4^{2+\alpha}C_0^{-\frac{2\beta+d}{2\beta}} \frac{2^{2(\beta+d-\alpha\beta)/\beta}}{2^{(\beta+d-\alpha\beta)/\beta}-1} T^{\frac{\beta+d-\alpha\beta}{2\beta+d}}  \\
			&\times (\frac{2\beta + d + \beta d}{2\beta + d}\log T)^{\frac{2\beta+d}{2\beta}}
			+ \frac{4}{p^2(2^{1 + \alpha} - 1)}\MarginL4^{1+\alpha}\log T \\
			&+ (5 + 4M^2)\frac{\beta^2 \log^2 T}{(2\beta + d)^2\log^2 2} \\
			&= \Tilde{O}(T^{\frac{\beta+d-\alpha\beta}{2\beta+d}}).
	\end{align*}
\endproof

\subsection[appendix]{Proofs for \cref{section: lower bound}}

\proof{Proof of \cref{lb}.}

Our proof of \cref{lb} combines ideas from the proofs of Theorem 3.5 in \cite{audibert2007} and Theorem 4.1 in \cite{Rigollet10}.
\edit{Note that
the instance constructed in \cite{Rigollet10} is different from our instance here, and the expected reward function in their instance is non-differentiable, so it is not suitable for our purpose. Instead, we adapt an argument from \cite{audibert2007} for offline nonparametric classification to our online bandit setting.}

\textbf{First, we construct a class $\mathcal{H}=\{\mathbb{P}_{\sigma}:\sigma\in \Sigma_{m} = \{-1,1\}^{m}\}$ of probability distributions of $(X, Y_1, Y_{-1})$.}

Fix constants $\delta_0\in (0,\frac{1}{2})$, $\kappa^2 = \frac{1}{4}-\delta_0^2$, $q = \lceil (\frac{T}{4 e \kappa^2})^{\frac{1}{2\beta+d}}\rceil$, $m = \lceil  q^{d-\alpha\beta}\rceil$ and $\omega =   q^{-d}$. Assume $T$ is sufficiently large so that $m\le q^d$, $\omega\le \frac{1}{m}$ and $T>4\kappa^2q^{2\beta+d}$.

Define
$$G_q = \{(\frac{2j_1 +1}{2q},\dots,\frac{2j_d+1}{2q}): j_i\in\{0,\dots,q-1\}, i=1,\dots, d\},$$
and we number the points in $G_q$ as $x_1,\dots, x_{q^d}$.
For any $x\in  [0,1]^d$, we denote by $g_q(x)=\arg\min_{x'\in G'}\magd{x-x'}$ the closest point to $x$ in $G'$.
If there are multiple closest points to $x$, we choose $g_q(x)$ to be the one closest to $(0, 0, \cdots, 0)$.
All points that share the same closest grid point $g_q(x)$ belong to a hypercube with length $\frac{1}{q}$ and center $g_q(x)$. We denote this hypercube as $\cube_q(x) = \{x'\in \mathcal{X}: g_q(x') = g_q(x)\}$.
Define $\mathcal{X}_i = \cube_q(x_i)$ for $i = 1,\dots,m$ and $\mathcal{X}_0 = [0,1]^d-\bigcup_{i=1}^m \mathcal{X}_i$.

The marginal distribution of $X$ (denoted by $\mathbb{P}_X$) does not depend on $\sigma$. Its density in $\mathbb{R}^d$ is
\[
    \mu_X(x)=
\begin{cases}
    \frac{\omega}{\text{Leb}[\mathcal{B}(0,\frac{1}{4q})]},& \text{if } x \in  \bigcup_{i=1}^m \mathcal{B}(x_i, \frac{1}{4q}),\\
    \frac{1-m\omega}{\text{Leb}[A_0]},              & \text{if } x\in \mathcal{X}_0,\\
    0, & \text{otherwise}.
\end{cases}
\]

The conditional distribution of $Y(-1)$ given $X$ does not depend on $\sigma$ either. It is simply Bernoulli distribution with conditional expectation is $\eta^{\sigma}_{-1} \equiv \frac{1}{2}$.

We now define the conditional distribution of $Y(1)$ given $X$ for $\mathbb{P}_{\sigma}\in \mathcal{H}$.

Consider an infinitely differentiable function $u_1$ defined as
\[
    u_1(x)=
\begin{cases}
    \exp\{-\frac{1}{(\frac{1}{2}-x)(x-\frac{1}{4})}\},& \text{if } x \in (\frac{1}{4},\frac{1}{2}),\\
    0,              & \text{otherwise},
\end{cases}
\]
and take $u: \mathbb{R}_+ \rightarrow \mathbb{R}_+$ to be
$$u(x) = (\int_{\frac{1}{4}}^{\frac{1}{2}}u_1(t)d t)^{-1}\int_x^{\infty}u_1(t)d t.$$
It is easy to verify that $u$ is a non-increasing infinitely differentiable function satisfying $u=1$ on $[0,\frac{1}{4}]$ and $u=0$ on $[\frac{1}{2},\infty)$. Moreover, for any integer $l\ge 1$, the $l$-th derivative of $u(x)$ at $x\in (\frac{1}{4},\frac{1}{2})$ is in the form of $\frac{poly(x)}{((\frac{1}{2}-x)(x-\frac{1}{4}))^{2(l-1)}}\exp(-\frac{1}{(\frac{1}{2}-x)(x-\frac{1}{4})})$, which is bounded in the domain.
\edit{
Therefore, we can find small enough constant $C_{\phi}\in(0,\delta_0]$ such that $\phi: \mathbb{R}^d \rightarrow R_+$ defined as
\begin{align} \label{eq: phi function}
\phi(x)\triangleq C_{\phi}u(||x||)
\end{align}
satisfies the condition that for any $x,x'\in \mathbb{R}^d$ and $|r| = \floor*{\beta}$, $|D^r \phi(x') - D^r \phi_x(x')|\le \frac{L}{C_{\beta}}||x'-x||^{\beta - \floor*{\beta}}$ and $|\phi(x') - \phi(x)|\le L_1||x'-x||$, where $C_{\beta} = \sum_{|r|=\floor*{\beta}} \frac{1}{r!}$.
}

Define $\eta^{\sigma}_1: [0,1]^d\rightarrow \mathbb{R}$ as
$$\eta^{\sigma}_1(x) = \frac{1}{2}+\sum_{j=1}^m \sigma_j\varphi_j(x),$$
where $\varphi_j(x) = q^{-\beta}\phi(q[x-x_j])\mathbb{I}(x\in \mathcal{X}_j)$. Since $C_{\phi}\le \delta_0$, $\eta^{\sigma}_1 \in [\frac{1}{2}-\delta_0, \frac{1}{2}+\delta_0]\subset [0,1]$. Therefore, we can define $\expect_{\sigma}(Y(1)|X) = \eta^{\sigma}_1(X)$, and the conditional distribution of $Y(1)$ given $X$ is Bernoulli distribution with mean $\eta^{\sigma}_1(X)$. This completes the construction.

\textbf{Second, we check that $\mathcal{H}\subseteq\mathcal{P}$.}

Fix any $\sigma\in \Sigma_m$ and consider distribution $\mathbb{P}_{\sigma}$.
\begin{enumerate}
    \item Smooth Conditional Expected Rewards (Assumption \ref{assump: smoothness})

    The verification for $\eta^{\sigma}_{-1}$ is trivial.

    \edit{
    For any $l\in \mathbb{N}_+^d$ such that $\abs{l}\le \floor*{\beta}$ and $j\in\{1,\dots, m\}$,
    $$D^l \varphi_j(x) = q^{|l|-\beta}D^l \phi(q[x-g_q(x)])\mathbb{I}\{x\in \mathcal{X}_j\}.$$
    Recall that our choice of $C_{\phi}$ ensures that for any $|r|  = \floor*{\beta}$, $|D^r \phi(x') - D^r \phi_x(x')|\le \frac{L}{C_{\beta}}||x'-x||^{\beta - \floor*{\beta}}$.
    When $x$ and $x'$ are in the same cube, we have
    \begin{align*}
      \|D^{r}\eta_1^{\sigma}(x') - D^{r}\eta_1^{\sigma}(x)\| \le &
    \|D^{r}\varphi_j(x') - D^{r}\varphi_j(x)\|\\ \le&q^{\floor*{\beta} - \beta}\|D^{r}\phi(q(x' - g_q(x'))) - D^{r}\phi(q(x - g_q(x))) \| \\
    \le& \frac{L}{C_{\beta}} q^{\floor*{\beta} - \beta}\|q(x' - g_q(x')) - q(x - g_q(x))\|^{\beta - \floor*{\beta}}\\ \le & \frac{L}{C_{\beta}}\|(x' - g_q(x')) - (x - g_q(x))\|^{\beta - \floor*{\beta}} \\
    = & \frac{L}{C_{\beta}}\|x' - x\|^{\beta - \floor*{\beta}}.
    \end{align*}
    When $x$ and $x'$ are in different cubes, we can always find $x''$ in the same cube as $x$ such that $D^{r}\eta_1^{\sigma}(x'') = D^{r}\eta_1^{\sigma}(x')$ and
    \begin{align*}
      \|D^{r}\eta_1^{\sigma}(x') - D^{r}\eta_1^{\sigma}(x)\| =   \|D^{r}\eta_1^{\sigma}(x'') - D^{r}\eta_1^{\sigma}(x)\| \le \frac{L}{C_{\beta}}\|x'' - x\|^{\beta - \floor*{\beta}} \le
      \frac{L}{C_{\beta}}\|x' - x\|^{\beta - \floor*{\beta}}.
    \end{align*}
     Therefore, for any $x,x'\in [0,1]^d$, there exists $\lambda\in[0,1]$ such that
\begin{align*}
  \|\eta_1^{\sigma}(x') - \sum_{|r| \le \floor*{\beta}}\frac{(x' - x)^r}{r!}D^r \eta_1^{\sigma}(x) \|\le& \|\sum_{|r| = \floor*{\beta}} \frac{(x' - x)^r}{r!}D^r \eta_1^{\sigma}(x' + \lambda(x - x')) - \sum_{|r| = \floor*{\beta}} \frac{(x' - x)^r}{r!}D^r \eta_1^{\sigma}(x) \| \\
  \le& C_{\beta}\|x' - x\|^{\floor*{\beta}} \frac{L}{C_{\beta}} \|x' - x\|^{\beta - \floor*{\beta}}\\ = &L\|x' - x\|^\beta,
\end{align*}
and $|\eta^{\sigma}_1(x') - \eta^{\sigma}_1(x)|\le L_1||x'-x||$.
}

  \item Optimal Decision Regions (Assumption \ref{assump: decision-set})

  \edit{
  Note that by construction $\mathcal{X}_0 \subseteq \mathcal Q_a= \{x:a\tau(x)\ge 0\} \cap \mathcal{X}$.
  Moreover, by the construction of $m$ and $w$, $\text{Leb}[\mathcal{X}_0] = (1+o(1))\text{Leb}([0,1]^d) = (1+o(1))\text{Leb}(\mathcal X)$, or equivalently $\text{Leb}\left[\cup_{i = 1}^m \mathcal X_i\right] \ll \text{Leb}[\mathcal{X}_0]$. This means that for sufficiently large $T$, the whole covariate support is almost entirely occupied by $\mathcal X_0$.
  }

  \edit{
  Let's fix a point $x \in \mathcal Q_a$ and a sufficiently large $T$. We consider two cases:
  \begin{itemize}
   \item When $r \gg \frac{1}{q}$, then by the fact that the whole covariate support is almost entirely occupied by $\mathcal X_0$, $\text{Leb}[\mathcal{B}(x,r)\cap\mathcal Q_a] \ge \text{Leb}[\mathcal{B}(x,r)\cap\mathcal X_0] = (1 + o(1))\text{Leb}[\mathcal{B}(x,r)]$, so regularity is satisfied.
   \item When $r = O(\frac{1}{q})$, if $x \in \mathcal X_0$, the regularity condition is satisfied because each hypercube satisfies the regularity condition.
   On the other hand, if $x \in \mathcal Q_a \setminus \mathcal X_0 = \cup_{i = 1}^m \mathcal X_i \cap \mathcal Q_a$ (and without loss of generality let $x\in \mathcal{B}(x_1, \frac{1}{4q})$), then $\text{Leb}[\mathcal{B}(x,r) \cap \mathcal Q_a] \ge \text{Leb}[\mathcal{B}(x,r) \cap \mathcal{B}(x_1, \frac{1}{4q})]$. Then the regularity condition is satisfied by the fact that the ball $\mathcal{B}(x_1, \frac{1}{4q})$ is regular.
  \end{itemize}
  }

    \item Strong Density (Assumption \ref{assump: support})

     The support of $X$ is $\mathcal{X} = \bigcup_{i=1}^m \mathcal{B}(x_i, \frac{1}{4q}) \cup \mathcal X_0$, which is compact. By definition,
     \[
    \mu_X(x)=
\begin{cases}
     \frac{4^d }{v_d},& \text{if } x \in  \bigcup_{i=1}^m \mathcal{B}(x_i, \frac{1}{4q}),\\
    \frac{1-m\omega}{1-m q^{-d}} = 1,              & \text{if } x\in \mathcal{X}_0.
\end{cases}
\]

The strong density condition is satisfied with $\mu_{\max} = \frac{4^d}{v_d}$ and $\mu_{\min} = 1$.

    \item Margin Condition (Assumption \ref{assump: margin})

    Let $x_0 = (\frac{1}{2q}, \dots, \frac{1}{2q})$. We have
    \begin{align*}
        \mathbb{P}_{\sigma}(0<|\tau(X)|\le t) &= \mathbb{P}_{\sigma}(0<|\eta_1^{\sigma}(X)-\frac{1}{2}|\le t)\\
        & = m\mathbb{P}_{\sigma}(0<\phi[q(X-x_0)]\le t q^{\beta})\\
        & = m\int_{\mathcal{B}(x_0,\frac{1}{4q})} \mathbb{I}\{0<\phi[q(x-x_0)]\le t q^{\beta}\} \frac{\omega}{\text{Leb}[\mathcal{B}(0,\frac{1}{4q})]} d x \\
        & = \frac{m\omega }{\text{Leb}[\mathcal{B}(0,\frac{1}{4})]}\int_{\mathcal{B}(0,\frac{1}{4})}\mathbb{I}\{\phi(x)\le t q^{\beta}\}d x\\
        & = m\omega\mathbb{I}\{t\ge C_{\phi}q^{-\beta}\}.
    \end{align*}
    Note that $m\omega = O(q^{-\alpha\beta})$, and Assumption \ref{assump: margin} with is satisfied with $\MarginL =2C_{\phi}^{-\alpha}$.

\end{enumerate}

\textbf{Finally, we prove a lower bound for $\expect I_t(\pi)$ based on problem instances in $\mathcal{H}$.}

For any policy $\pi$ and any $t = 1, \dots, T$, denote by $\mathbb{P}_{\pi,\sigma}^t$ the joint distribution of
$$(X_1, Y_1(\pi_1(X_1))), \dots, (X_t, Y_t(\pi_t(X_t)))$$
where $(X_t, Y_t(1), Y_t(-1))$ are generated i.i.d from $\mathbb{P}_{\sigma}$, and $\expect_{\pi,\sigma}^t$ the corresponding expectation.

Observe that
\begin{align*}
    \sup_{\sigma\in\Sigma_m} \expect_{\sigma}I_t(\pi) & =  \sup_{\sigma\in\Sigma_m}\sum_{t=1}^T \expect_{\pi,\sigma}^{t-1} \mathbb{P}_X[\pi_t(X_t) \neq \sign(\eta_1^{\sigma}(X_t))]\\
    & = \sup_{\sigma\in\Sigma_m}\sum_{j=1}^m\sum_{t=1}^T \expect_{\pi,\sigma}^{t-1} \mathbb{P}_X[\pi_t(X_t) \neq \sigma_j, X_t\in \mathcal{X}_j]\\
    & \ge \frac{1}{2^m}\sum_{j=1}^m\sum_{t=1}^T \sum_{\sigma\in \Sigma_m} \expect_{\pi,\sigma}^{t-1} \mathbb{P}_X[\pi_t(X_t) \neq \sigma_j, X_t\in \mathcal{X}_j]\\
    & =  \frac{1}{2^m}\sum_{j=1}^m\sum_{t=1}^T \sum_{\sigma_{[-j]}\in \Sigma_{m-1}}\sum_{i\in\{-1,1\}} \expect_{\pi,\sigma_{[-j]}^i}^{t-1} \mathbb{P}_X[\pi_t(X_t) \neq i, X_t\in \mathcal{X}_j],
\end{align*}
where $\sigma_{[-j]} = (\sigma_1,\dots,\sigma_{j-1},\sigma_{j+1},\dots,\sigma_m)$ and $\sigma^i_{[-j]} = (\sigma_1,\dots,\sigma_{j-1},i,\sigma_{j+1},\dots,\sigma_m)$.

For $j=1,\dots,m$, define $\mathbb{P}_X^j(\cdot) = \mathbb{P}_X(\cdot|X\in \mathcal{X}_j)$. By Theorem $2.2(iii)$ in \cite{Tsybakov:2008:INE:1522486},
\begin{align*}
    \sum_{i\in\{-1,1\}}\expect_{\pi,\sigma_{[-j]}^i}^{t-1} \mathbb{P}_X[\pi_t(X_t) \neq i, X_t\in \mathcal{X}_j] & = \frac{1}{q^d}\sum_{i\in\{-1,1\}}\expect_{\pi,\sigma_{[-j]}^i}^{t-1} \mathbb{P}^j_X[\pi_t(X_t) \neq i]\\
    & \ge \frac{1}{4 q^d} \exp[-\mathcal{K}(\mathbb{P}_{\pi,\sigma_{[-j]}^{-1}}^{t-1}\times \mathbb{P}^j_X, \mathbb{P}_{\pi,\sigma_{[-j]}^1}^{t-1}\times \mathbb{P}^j_X)] \\
    & = \frac{1}{4 q^d} \exp[-\mathcal{K}(\mathbb{P}_{\pi,\sigma_{[-j]}^{-1}}^{t-1}, \mathbb{P}_{\pi,\sigma_{[-j]}^1}^{t-1})],
\end{align*}
where $\mathcal{K}(\cdot, \cdot)$ denotes the KL-divergence of two probability distributions (\cite{kullback1951}).

For $t = 2,\dots, T$, denote by $\mathcal{F}^+_t$ the $\sigma$-algebra generated by $X_t$ and $(X_s, Y_s(\pi_s(X_s))), s=1,\dots, t-1$. Denote by $\mathbb{P}_{\pi,\sigma}^{\cdot|\mathcal{F}_t^+}$ the conditional distribution given $\mathcal{F}_t^+$. By the chain rule for KL divergence, for $t=1,\dots, T$,
 \begin{align*}
     \mathcal{K}(\mathbb{P}_{\pi,\sigma_{[-j]}^{-1}}^t, \mathbb{P}_{\pi,\sigma_{[-j]}^1}^t) & = \mathcal{K}(\mathbb{P}_{\pi,\sigma_{[-j]}^{-1}}^{t-1}, \mathbb{P}_{\pi,\sigma_{[-j]}^1}^{t-1}) + \mathbb{E}_{\pi,\sigma_{[-j]}^{-1}}^{t-1}\expect_{X}[\mathcal{K}(\mathbb{P}_{\pi,\sigma_{[-j]}^{-1}}^{(X_t, Y_t(\pi_t(X_t)))|\mathcal{F}_t^+}, \mathbb{P}_{\pi,\sigma_{[-j]}^1}^{(X_t, Y_t(\pi_t(X_t)))|\mathcal{F}_t^+})]\\
     & = \mathcal{K}(\mathbb{P}_{\pi,\sigma_{[-j]}^{-1}}^{t-1}, \mathbb{P}_{\pi,\sigma_{[-j]}^1}^{t-1}) + \mathbb{E}_{\pi,\sigma_{[-j]}^{-1}}^{t-1}\expect_{X}[\mathcal{K}(\mathbb{P}_{\pi,\sigma_{[-j]}^{-1}}^{ Y_t(\pi_t(X_t))|\mathcal{F}_t^+}, \mathbb{P}_{\pi,\sigma_{[-j]}^1}^{ Y_t(\pi_t(X_t))|\mathcal{F}_t^+})].
 \end{align*}
Lemma 4.1 in \cite{Rigollet10} shows that for any $\eta',\eta''\in (\frac{1}{2}-\delta_0,\frac{1}{2}+\delta_0)$, the KL-divergence of two Bernoulli distributions with mean $\eta'$ and $\eta''$ respectively satisfies
$$\mathcal{K}(\text{Bernoulli}(\eta'),\text{Bernoulli}(\eta''))\le \frac{1}{\kappa^2}(\eta'-\eta'')^2.$$
This implies,
 \begin{align*}
     \mathcal{K}(\mathbb{P}_{\pi,\sigma_{[-j]}^{-1}}^t, \mathbb{P}_{\pi,\sigma_{[-j]}^1}^t)
     & \le \mathcal{K}(\mathbb{P}_{\pi,\sigma_{[-j]}^{-1}}^{t-1}, \mathbb{P}_{\pi,\sigma_{[-j]}^1}^{t-1}) + \mathbb{E}_{\pi,\sigma_{[-j]}^{-1}}^{t-1}\expect_{X}[\frac{1}{\kappa^2}(\eta_1^{\sigma_{[-j]}^1}(X_t)-\eta_1^{\sigma_{[-j]}^{-1}}(X_t))^2\mathbb{I}\{\pi_t(X_t) = 1\}]\\
     & \le \mathcal{K}(\mathbb{P}_{\pi,\sigma_{[-j]}^{-1}}^{t-1}, \mathbb{P}_{\pi,\sigma_{[-j]}^1}^{t-1}) + \mathbb{E}_{\pi,\sigma_{[-j]}^{-1}}^{t-1}\expect_{X}[\frac{4C_{\phi}^2}{\kappa^2 q^{2\beta}}\mathbb{I}\{\pi_t(X_t) = 1, X_t\in\mathcal{X}_j\}]\\
     & \le \mathcal{K}(\mathbb{P}_{\pi,\sigma_{[-j]}^{-1}}^{t-1}, \mathbb{P}_{\pi,\sigma_{[-j]}^1}^{t-1}) + \mathbb{E}_{\pi,\sigma_{[-j]}^{-1}}^{t-1}\expect_{X}[\frac{1}{\kappa^2 q^{2\beta}}\mathbb{I}\{\pi_t(X_t) = 1, X_t\in\mathcal{X}_j\}],
 \end{align*}
 where the last inequality follows from $C_{\phi}\le \delta_0< \frac{1}{2}$. Define
 $$\mathcal{N}_{j,\pi} = \expect_{\pi,\sigma_{[-j]}^{-1}}^{T-1}\expect_X[\sum_{t=1}^T\mathbb{I}\{\pi_t(X_t) = 1, X_t\in\mathcal{X}_j\}].$$
 By induction, for $t=1,\dots, T$,
 $$\mathcal{K}(\mathbb{P}_{\pi,\sigma_{[-j]}^{-1}}^{t-1}, \mathbb{P}_{\pi,\sigma_{[-j]}^1}^{t-1}) \le \frac{1}{\kappa^2 q^{2\beta}}\mathcal{N}_{j,\pi}.$$
 This implies
 $$\sum_{\sigma_{[-j]}\in \Sigma_{m-1}}\sum_{i\in\{-1,1\}} \expect_{\pi,\sigma_{[-j]}^i}^{t-1} \mathbb{P}_X[\pi_t(X_t) \neq i, X_t\in \mathcal{X}_j]\ge \frac{2^{m-1}}{4q^d}\exp(-\frac{1}{\kappa^2q^{2\beta}}\mathcal{N}_{j,\pi}).$$
  Moreover, it is trivially true that
 $$\sum_{t=1}^T \sum_{\sigma\in \Sigma_m} \expect_{\pi,\sigma}^{t-1} \mathbb{P}_X[\pi_t(X_t) \neq \sigma_j, X_t\in \mathcal{X}_j]\ge 2^{m-1}\mathcal{N}_{j,\pi}.$$
 Therefore,
 \begin{align*}
   \sup_{\sigma\in\Sigma_m} \expect_{\sigma}I_t(\pi) \ge & \frac{2^{m-1}}{2^m}\sum_{j=1}^m\max\{\frac{T}{4q^d}\exp(-\frac{1}{\kappa^2q^{2\beta}}\mathcal{N}_{j,\pi}), \mathcal{N}_{j,\pi}\} \\
    \ge & \frac{1}{4}\sum_{j=1}^m\{\frac{T}{4q^d}\exp(-\frac{1}{\kappa^2q^{2\beta}}\mathcal{N}_{j,\pi})+ \mathcal{N}_{j,\pi}\}\\
    \ge & \frac{m}{4}\inf_{z\ge 0}\{\frac{T}{4q^d}\exp(-\frac{z}{\kappa^2q^{2\beta}})+ z\}.
 \end{align*}
 Since $T>4\kappa^2q^{2\beta+d}$, we have
 $$z^* = \text{argmin}_{z\ge 0}\{\frac{T}{4q^d}\exp(-\frac{z}{\kappa^2q^{2\beta}})+ z\} = \kappa^2q^{2\beta}\log(\frac{T}{4\kappa^2q^{2\beta+d}}) = c^*T^{\frac{2\beta}{2\beta+d}},$$
 where $c^*$ is a positive constant. It follows that
 $$ \sup_{\sigma\in\Sigma_m} \expect_{\sigma}I_t(\pi) \ge cT^{1-\frac{\alpha\beta}{2\beta+d}}.$$
\endproof

\blockedit
\proof{Proof of \cref{lb2}}
\textbf{First, we construct a class $\mathcal{H}'=\{\mathbb{P}_{\sigma}:\sigma\in \Sigma_{\bar{m}} = \{-1,1\}^{\bar{m}}\}$ of probability distributions of $(X, Y_1, Y_{-1})$.}

Fix constants $\Delta = \frac{1}{2}(\beta+\frac{\beta d}{\alpha\beta + d})$, $\delta_0\in (0,\frac{1}{2})$, $\kappa^2 = \frac{1}{4}-\delta_0^2$, $\bar{p} = (\frac{v_d T}{2^d e \kappa^2})^{\frac{1}{2\Delta+d}}$, $\bar{q} = \lceil \bar{p}^{\Delta/\beta} \rceil$, $\bar{m} = \lceil  \bar{p}^{d-\alpha\Delta}\rceil$ and $\bar{\omega} =  \frac{1}{1 - \bar{m} (\bar{q}^{-d} - v_d (2\bar{p})^{-d})} $. Assume $T$ is sufficiently large so that $\bar{m}\le \bar{q}^d/2$ and $2\bar{q}\le \bar{p}$.

Define
$$G_{\bar{q}} = \{(\frac{2j_1 +1}{2\bar{q}},\dots,\frac{2j_d+1}{2\bar{q}}): j_i\in\{0,\dots,\bar{q}-1\}, i=1,\dots, d\},$$
and we number the points in $G_{\bar{q}}$ as $x_1,\dots, x_{\bar{q}^d}$.
For any $x\in  [0,1]^d$, we denote by $g_{\bar{q}}(x)=\arg\min_{x'\in G_{\bar{q}}}\magd{x-x'}$ the closest point to $x$ in $G_{\bar{q}}$.
If there are multiple closest points to $x$, we choose $g_{\bar{q}}(x)$ to be the one closest to $(0, 0, \cdots, 0)$.
All points that share the same closest grid point $g_{\bar{q}}(x)$ belong to a hypercube with length $\frac{1}{\bar{q}}$ and center $g_{\bar{q}}(x)$. We denote this hypercube as $\cube_{\bar{q}}(x) = \{x'\in \mathcal{X}: g_{\bar{q}}(x') = g_{\bar{q}}(x)\}$.
Define $\mathcal{X}_i = \cube_{\bar{q}}(x_i)$ for $i = 1,\dots,\bar{m}$ and $\mathcal{X}_0 = [0,1]^d-\bigcup_{i=1}^{\bar{m}} \mathcal{X}_i$.

The marginal distribution of $X$ (denoted by $\mathbb{P}_X$) does not depend on $\sigma$. Its density in $\mathbb{R}^d$ is
\[
    \mu_X(x)=
\begin{cases}
    \bar{\omega} ,& \text{if } x \in  \bigcup_{i=1}^{\bar{m}} \mathcal{B}(x_i, \frac{1}{2\bar{p}}) \cup \mathcal{X}_0,\\
    0, & \text{otherwise}.
\end{cases}
\]
Note here the radius of the $\bar{m}$ balls is $\frac{1}{2\bar{p}}$ and it is (much) smaller than the cube size $\frac{1}{\bar{q}}$.

The conditional distribution of $Y(-1)$ given $X$ does not depend on $\sigma$ either. It is simply Bernoulli distribution with conditional expectation is $\eta^{\sigma}_{-1} \equiv \frac{1}{2}$.

We now define the conditional distribution of $Y(1)$ given $X$ for $\mathbb{P}_{\sigma}\in \mathcal{H}$.
Let $\phi(x)$ be defined as in \cref{eq: phi function}. Define $\eta^{\sigma}_1: [0,1]^d\rightarrow \mathbb{R}$ as
$$\eta^{\sigma}_1(x) = \frac{1}{2}+\sum_{j=1}^{\bar{m}} \sigma_j\varphi_j(x),$$
where $\varphi_j(x) = \bar{q}^{-\beta}\phi(\bar{q}[x-n_{\bar{q}}(x)])\mathbb{I}(x\in \mathcal{X}_j)$. Since $C_{\phi}\le \delta_0$, $\eta^{\sigma}_1 \in [\frac{1}{2}-\delta_0, \frac{1}{2}+\delta_0]\subset [0,1]$. Therefore, we can define $\expect_{\sigma}(Y(1)|X) = \eta^{\sigma}_1(X)$, and the conditional distribution of $Y(1)$ given $X$ is Bernoulli distribution with mean $\eta^{\sigma}_1(X)$. This completes the construction.

\textbf{Second, we check that functions in $\mathcal{H}'$ satisfy \cref{assump: smoothness}, \cref{assump: support} and \cref{assump: margin}. }

Fix any $\sigma\in \Sigma_{\bar{m}}$ and consider distribution $\mathbb{P}_{\sigma}$.

\begin{enumerate}
    \item Smooth Conditional Expected Rewards (\cref{assump: smoothness})

    This can be verified in the same way as discussed in the proof of \cref{lb} (we only need to change $q$ to $\bar{q}$), so we omit the details here.

    \item Strong Density (\cref{assump: support})

     The support of $X$ is $\mathcal{X} = \bigcup_{i=1}^{\bar{m}} \mathcal{B}(x_i, \frac{1}{2 \bar{p}}) \cup \mathcal{X}_0$, which is compact. By definition, since $\bar{m}\le \bar{q}^d/2$, $\mu_X(x) = \bar{\omega} \in (1,2)$ on its support, so the strong density condition is satisfied with $\mu_{\max} = 2$ and $\mu_{\min} = 1$.

    \item Margin Condition (Assumption \ref{assump: margin})

    Let $x_0 = (\frac{1}{2\bar{q}}, \dots, \frac{1}{2\bar{q}})$. We have
    \begin{align*}
        \mathbb{P}_{\sigma}(0<|\tau(X)|\le t) &= \mathbb{P}_{\sigma}(0<|\eta_1^{\sigma}(X)-\frac{1}{2}|\le t)\\
        & = \bar{m}\mathbb{P}_{\sigma}(0<\phi[\bar{q}(X-x_0)]\le t \bar{q}^{\beta})\\
        & = \bar{m}\bar{\omega}\int_{\mathcal{B}(x_0,\frac{1}{2\bar{p}})} \mathbb{I}\{0<\phi[\bar{q}(x-x_0)]\le t \bar{q}^{\beta}\}  d x \\
        & = \frac{\bar{m}\bar{\omega} v_d}{(2\bar{p})^d}\mathbb{I}\{t\ge C_{\phi}\bar{q}^{-\beta}\}.
    \end{align*}
   Since $\frac{\bar{m}\bar{\omega}}{\bar{p}^d} = O(\bar{p}^{-\alpha\Delta}) = O(\bar{q}^{-\alpha\beta})$, Assumption \ref{assump: margin} with is satisfied.

\end{enumerate}

Note here $\frac{1}{2\bar{p}} = o(\frac{1}{\bar{q}})$, i.e., the radius of the $\bar{m}$ balls is (much) smaller than the hypercube size, in contrast to instances in the proof of \cref{lb} where the ball size is proportional to the hypercube size. As a result, for any $x \in \bigcup_{i=1}^{\bar{m}} \mathcal{B}(x_i, \frac{1}{2\bar{p}}) \cap \mathcal Q_a$ and $r = \Theta(\frac{1}{\bar q})$, $\text{Leb}[\mathcal B(x, r)\cap \mathcal Q_a] = O\left(\text{Leb}[\mathcal B(x, r)\cap \mathcal \mathcal{B}(g_{\bar q}(x), \frac{1}{2\bar{p}})]\right) = O\left(\text{Leb}[\mathcal \mathcal{B}(g_{\bar q}(x), \frac{1}{2\bar{p}})]\right) = o(\text{Leb}[\mathcal B(x, r)])$, i.e., \cref{assump: decision-set} is violated.

\textbf{Finally, we prove a lower bound for $\expect I_t(\pi)$ based on problem instances in $\mathcal{H}$.}

For any policy $\pi$ and any $t = 1, \dots, T$, denote by $\mathbb{P}_{\pi,\sigma}^t$ the joint distribution of
$$(X_1, Y_1(\pi_1(X_1))), \dots, (X_t, Y_t(\pi_t(X_t)))$$
where $(X_t, Y_t(1), Y_t(-1))$ are generated i.i.d from $\mathbb{P}_{\sigma}$, and $\expect_{\pi,\sigma}^t$ the corresponding expectation.

Observe that
\begin{align*}
    \sup_{\sigma\in\Sigma_{\bar{m}}} \expect_{\sigma}I_t(\pi) & =  \sup_{\sigma\in\Sigma_{\bar{m}}}\sum_{t=1}^T \expect_{\pi,\sigma}^{t-1} \mathbb{P}_X[\pi_t(X_t) \neq \sign(\eta_1^{\sigma}(X_t))]\\
    & = \sup_{\sigma\in\Sigma_{\bar{m}}}\sum_{j=1}^{\bar{m}}\sum_{t=1}^T \expect_{\pi,\sigma}^{t-1} \mathbb{P}_X[\pi_t(X_t) \neq \sigma_j, X_t\in \mathcal{X}_j]\\
    & \ge \frac{1}{2^{\bar{m}}}\sum_{j=1}^{\bar{m}}\sum_{t=1}^T \sum_{\sigma\in \Sigma_{\bar{m}}} \expect_{\pi,\sigma}^{t-1} \mathbb{P}_X[\pi_t(X_t) \neq \sigma_j, X_t\in \mathcal{X}_j]\\
    & =  \frac{1}{2^{\bar{m}}}\sum_{j=1}^{\bar{m}}\sum_{t=1}^T \sum_{\sigma_{[-j]}\in \Sigma_{{\bar{m}}-1}}\sum_{i\in\{-1,1\}} \expect_{\pi,\sigma_{[-j]}^i}^{t-1} \mathbb{P}_X[\pi_t(X_t) \neq i, X_t\in \mathcal{X}_j],
\end{align*}
where $\sigma_{[-j]} = (\sigma_1,\dots,\sigma_{j-1},\sigma_{j+1},\dots,\sigma_m)$ and $\sigma^i_{[-j]} = (\sigma_1,\dots,\sigma_{j-1},i,\sigma_{j+1},\dots,\sigma_m)$.

For $j=1,\dots,\bar{m}$, define $\mathbb{P}_X^j(\cdot) = \mathbb{P}_X(\cdot|X\in \mathcal{X}_j)$. By Theorem $2.2(iii)$ in \cite{Tsybakov:2008:INE:1522486},
\begin{align*}
    \sum_{i\in\{-1,1\}}\expect_{\pi,\sigma_{[-j]}^i}^{t-1} \mathbb{P}_X[\pi_t(X_t) \neq i, X_t\in \mathcal{X}_j] & = \frac{v_d \bar{\omega}}{(2\bar{p})^d}\sum_{i\in\{-1,1\}}\expect_{\pi,\sigma_{[-j]}^i}^{t-1} \mathbb{P}^j_X[\pi_t(X_t) \neq i]\\
    & \ge \frac{v_d }{4 (2\bar{p})^d} \exp[-\mathcal{K}(\mathbb{P}_{\pi,\sigma_{[-j]}^{-1}}^{t-1}\times \mathbb{P}^j_X, \mathbb{P}_{\pi,\sigma_{[-j]}^1}^{t-1}\times \mathbb{P}^j_X)] \\
    & = \frac{v_d}{4 (2\bar{p})^d} \exp[-\mathcal{K}(\mathbb{P}_{\pi,\sigma_{[-j]}^{-1}}^{t-1}, \mathbb{P}_{\pi,\sigma_{[-j]}^1}^{t-1})],
\end{align*}
where $\mathcal{K}(\cdot, \cdot)$ denotes the KL-divergence of two probability distributions (\cite{kullback1951}).

For $t = 2,\dots, T$, denote by $\mathcal{F}^+_t$ the $\sigma$-algebra generated by $X_t$ and $(X_s, Y_s(\pi_s(X_s))), s=1,\dots, t-1$. Denote by $\mathbb{P}_{\pi,\sigma}^{\cdot|\mathcal{F}_t^+}$ the conditional distribution given $\mathcal{F}_t^+$. By the chain rule for KL divergence, for $t=1,\dots, T$,
 \begin{align*}
     \mathcal{K}(\mathbb{P}_{\pi,\sigma_{[-j]}^{-1}}^t, \mathbb{P}_{\pi,\sigma_{[-j]}^1}^t) & = \mathcal{K}(\mathbb{P}_{\pi,\sigma_{[-j]}^{-1}}^{t-1}, \mathbb{P}_{\pi,\sigma_{[-j]}^1}^{t-1}) + \mathbb{E}_{\pi,\sigma_{[-j]}^{-1}}^{t-1}\expect_{X}[\mathcal{K}(\mathbb{P}_{\pi,\sigma_{[-j]}^{-1}}^{(X_t, Y_t(\pi_t(X_t)))|\mathcal{F}_t^+}, \mathbb{P}_{\pi,\sigma_{[-j]}^1}^{(X_t, Y_t(\pi_t(X_t)))|\mathcal{F}_t^+})]\\
     & = \mathcal{K}(\mathbb{P}_{\pi,\sigma_{[-j]}^{-1}}^{t-1}, \mathbb{P}_{\pi,\sigma_{[-j]}^1}^{t-1}) + \mathbb{E}_{\pi,\sigma_{[-j]}^{-1}}^{t-1}\expect_{X}[\mathcal{K}(\mathbb{P}_{\pi,\sigma_{[-j]}^{-1}}^{ Y_t(\pi_t(X_t))|\mathcal{F}_t^+}, \mathbb{P}_{\pi,\sigma_{[-j]}^1}^{ Y_t(\pi_t(X_t))|\mathcal{F}_t^+})].
 \end{align*}
Lemma 4.1 in \cite{Rigollet10} shows that for any $\eta',\eta''\in (\frac{1}{2}-\delta_0,\frac{1}{2}+\delta_0)$, the KL-divergence of two Bernoulli distributions with mean $\eta'$ and $\eta''$ respectively satisfies
$$\mathcal{K}(\text{Bernoulli}(\eta'),\text{Bernoulli}(\eta''))\le \frac{1}{\kappa^2}(\eta'-\eta'')^2.$$
This implies
 \begin{align*}
     \mathcal{K}(\mathbb{P}_{\pi,\sigma_{[-j]}^{-1}}^t, \mathbb{P}_{\pi,\sigma_{[-j]}^1}^t)
     & \le \mathcal{K}(\mathbb{P}_{\pi,\sigma_{[-j]}^{-1}}^{t-1}, \mathbb{P}_{\pi,\sigma_{[-j]}^1}^{t-1}) + \mathbb{E}_{\pi,\sigma_{[-j]}^{-1}}^{t-1}\expect_{X}[\frac{1}{\kappa^2}(\eta_1^{\sigma_{[-j]}^1}(X_t)-\eta_1^{\sigma_{[-j]}^{-1}}(X_t))^2\mathbb{I}\{\pi_t(X_t) = 1\}]\\
     & \le \mathcal{K}(\mathbb{P}_{\pi,\sigma_{[-j]}^{-1}}^{t-1}, \mathbb{P}_{\pi,\sigma_{[-j]}^1}^{t-1}) + \mathbb{E}_{\pi,\sigma_{[-j]}^{-1}}^{t-1}\expect_{X}[\frac{4C_{\phi}^2}{\kappa^2 \bar{q}^{2\beta}}\mathbb{I}\{\pi_t(X_t) = 1, X_t\in\mathcal{X}_j\}]\\
     & \le \mathcal{K}(\mathbb{P}_{\pi,\sigma_{[-j]}^{-1}}^{t-1}, \mathbb{P}_{\pi,\sigma_{[-j]}^1}^{t-1}) + \mathbb{E}_{\pi,\sigma_{[-j]}^{-1}}^{t-1}\expect_{X}[\frac{1}{\kappa^2 \bar{p}^{2\Delta}}\mathbb{I}\{\pi_t(X_t) = 1, X_t\in\mathcal{X}_j\}],
 \end{align*}
 where the last inequality follows from $C_{\phi}\le \delta_0< \frac{1}{2}$ and $\bar{q}^{-2\beta} \le \bar{p}^{-2\Delta}$. Define
 $$\mathcal{N}_{j,\pi} = \expect_{\pi,\sigma_{[-j]}^{-1}}^{T-1}\expect_X[\sum_{t=1}^T\mathbb{I}\{\pi_t(X_t) = 1, X_t\in\mathcal{X}_j\}].$$
 By induction, for $t=1,\dots, T$,
 $$\mathcal{K}(\mathbb{P}_{\pi,\sigma_{[-j]}^{-1}}^{t-1}, \mathbb{P}_{\pi,\sigma_{[-j]}^1}^{t-1}) \le \frac{1}{\kappa^2 \bar{p}^{2\Delta}}\mathcal{N}_{j,\pi}.$$
 This implies
 $$\sum_{\sigma_{[-j]}\in \Sigma_{\bar{m}-1}}\sum_{i\in\{-1,1\}} \expect_{\pi,\sigma_{[-j]}^i}^{t-1} \mathbb{P}_X[\pi_t(X_t) \neq i, X_t\in \mathcal{X}_j]\ge \frac{2^{\bar{m}-1} v_d}{4(2\bar{p})^d}\exp(-\frac{1}{\kappa^2  \bar{p}^{2\Delta}}\mathcal{N}_{j,\pi}).$$
  Moreover, it is trivially true that
 $$\sum_{t=1}^T \sum_{\sigma\in \Sigma_{\bar{m}}} \expect_{\pi,\sigma}^{t-1} \mathbb{P}_X[\pi_t(X_t) \neq \sigma_j, X_t\in \mathcal{X}_j]\ge 2^{\bar{m}-1}\mathcal{N}_{j,\pi}.$$
 Therefore,
 \begin{align*}
   \sup_{\sigma\in\Sigma_{\bar{m}}} \expect_{\sigma}I_t(\pi) \ge & \frac{2^{\bar{m}-1}}{2^{\bar{m}}}\sum_{j=1}^{\bar{m}}\max\{\frac{ v_d T}{4(2\bar{p})^d}\exp(-\frac{1}{\kappa^2  \bar{p}^{2\Delta}}\mathcal{N}_{j,\pi}), \mathcal{N}_{j,\pi}\} \\
    \ge & \frac{1}{4}\sum_{j=1}^{\bar{m}}\{\frac{ v_d T}{4(2\bar{p})^d}\exp(-\frac{1}{\kappa^2  \bar{p}^{2\Delta}}\mathcal{N}_{j,\pi})+ \mathcal{N}_{j,\pi}\}\\
    \ge & \frac{\bar{m}}{4}\inf_{z\ge 0}\{\frac{ v_d T}{4(2\bar{p})^d}\exp(-\frac{z}{\kappa^2  \bar{p}^{2\Delta}})+ z\}.
 \end{align*}
 Since $T>\frac{2^d k^2\bar{p}^{2\Delta+d}}{v_d}$, we have
 $$z^* = \text{argmin}_{z\ge 0}\{\frac{ v_d T}{4(2\bar{p})^d}\exp(-\frac{z}{\kappa^2  \bar{p}^{2\Delta}})+ z\} = \frac{1}{2}\kappa^2\bar{p}^{2\Delta}\log(\frac{v_d T}{2^d\kappa^2\bar{p}^{2\Delta+d}}) = c^*T^{\frac{2\Delta}{2\Delta+d}},$$
 where $c^*$ is a positive constant. It follows that
 $$ \sup_{\sigma\in\Sigma_{\bar{m}}} \expect_{\sigma}I_t(\pi) \ge cT^{1-\frac{\alpha\Delta}{2\Delta+d}}.$$
\endproof

\subsection[appendix]{Proofs for Appendix \ref{section: k arm}}\label{sec: proof-multi-arm}
\proof{Proof of \cref{lemma: decision-set-characterization-multi-arm}}
We prove this lemma by induction.
Suppose the statements above hold up to $k = k_0$th epoch.

Statement \ref{decision-set-1}: $D_{\text{Est}, k_0 + 1}(a)$ is the region within $\mathcal R_{k_0}(a)$ where arm $a$ is estimable, and it was outperformed by another estimable arm according to the estimated expected reward $\hat\eta_{a, k_0}$. Thus under $\mathcal G_{k_0}$, it is indeed outperformed by another arm, i.e., $D_{\text{Est}, k_0 + 1}(a) \subseteq  \braces{x \in \mathcal X: \eta_{a}(x) < \max_{a' \in \mathcal A}\eta_{a'}(x)}$.

Statement \ref{decision-set-2}: it holds for $k = k_0 + 1$ because compared to $\mathcal R_{k_0}$ we have only removed
strictly suboptimal arms according to statement \ref{decision-set-1} for $k = k_0 + 1$ and statement \ref{decision-set-5} for $k = k_0$.

Statement  \ref{decision-set-3}: (1) only one arm remains in $k_0 + 1$ so the statement is trivially true; (2) at least two arms remain, and given that inestimable arms have been removed, these remaining arms have to be estimable with close estimated expected rewards, then $\mathcal G_{k_0}$ implies that their true expected rewards should be close.

Statement  \ref{decision-set-4}: $\prns{\bigcup_{\mathcal A_0: a \in \mathcal A_0} \mathcal R_{k_0 + 1}(\mathcal A_0) \cap \mathcal X}^C$ are regions where arm $a$ has already been removed, which means
\begin{align*}
\prns{\bigcup_{\mathcal A_0: a \in \mathcal A_0} \mathcal R_{k_0 + 1}(\mathcal A_0) \cap \mathcal X}^C \subseteq \prns{D_{\text{Irr}, k_0}(a) \cup D_{\text{Est}, k_0 + 1}(a)} \cap \mathcal X \subseteq \braces{x \in \mathcal X: \eta_{a}(x) < \max_{a' \in \mathcal A}\eta_{a;}(x)} \subseteq \mathcal Q_a^C,
\end{align*}
according to statement 1 for $k = k_0 + 1$ and statement 5 for $k = k_0$.
Statement 4 for time $k = k_0 + 1$ then follows.

Statement \ref{decision-set-5}: $\mathcal D_{\text{Irr}, k_0 + 1}(a) \subseteq \bigcup_{\mathcal A_0: a \in \mathcal A_0} \mathcal R_{k_0 + 1}(\mathcal A_0) \cap \mathcal X$ and by statement 4 for $k = k_0 + 1$, we know $\mathcal Q_a \subseteq \bigcup_{\mathcal A_0: a \in \mathcal A_0} \mathcal R_{k_0 + 1}(\mathcal A_0) \cap \mathcal X$, which implies that for any point $x$ in $\mathcal Q_a$, $ \bigcup_{\mathcal A_0: a \in \mathcal A_0} \mathcal R_{k_0 + 1}(\mathcal A_0) \cap \mathcal X$ should be regular at $x$ by the regularity of $\mathcal Q_a$. This  means that $x$ does not belong to $D_{\text{Irr}, k_0 + 1}(a)$. Thus $D_{\text{Irr}, k_0 + 1}(a) \subseteq \mathcal Q_a^C = \braces{x \in \mathcal X: \eta_{a}(x) < \max_{a' \in \mathcal A}\eta_{a'}(x)}$.
\endproof

\proof{Proof of \cref{lemma: strong-density-multi-arm}.}
	When $k  = 1$, the conclusions hold trivially since $\mathcal R_k(\mathcal A) = \mathcal{X}$, \ie, we pull each arm with equal probability for all samples in the first stage. This implies that $\mu_{a, 1} = \mu(x)$ for $x \in \mathcal{X}$. Then the conclusion follows from \cref{assump: multi-arm} condition iii. Denote $\mathcal{T}_{a, k} = \{t \in \mathcal{T}_k: A_t = a\}$, $N_{\mathcal A, k} = \{N_{a, k}\}_{a \in \mathcal A}$ and $n_{\mathcal A, k} = \{n_{a, k}\}_{a \in \mathcal A}$.

	When $k \ge 2$, $\{(X_t, Y_t), t\in \mathcal{T}_{a, k}\}$ are obviously i.i.d conditionally on $\mathcal{F}_{k-1}$, $\ogood_{k - 1}$ and $\omany_{k-1}$, since $A_t$ only depends on $X_t$ and $\mathcal{F}_{k - 1}$. This implies that $\{X_t:  t\in \mathcal{T}_{a, k}\}$ are i.i.d conditionally on $\mathcal{F}_{k-1}$, $\ogood_{k - 1}$, $\omany_{k - 1}$. Furthermore, $X_t {\perp\!\!\!\perp} N_{a,k}|A_t$, thus $\{X_t:  t\in \mathcal{T}_{a, k}\}$ are i.i.d given $\mathcal{F}_{k-1}$, $\ogood_{k - 1}$, $\omany_{k - 1}$, $N_{a, k} = n_{a, k}$. This also implies that for any $x \in \mathcal{X}$,
	\begin{align*}
		\mu_{a,k}(x)
			&= \mu_{X_t \mid A_t=a, \ogood_{k-1}, \omany_{k-1}, N_{\mathcal A, k} = n_{\mathcal A, k}, \mathcal{F}_{k-1}}(x) \\
			&= \mu_{X_t \mid A_t=a, \ogood_{k-1}, \omany_{k-1}, \mathcal{F}_{k-1}}(x)
	\end{align*}

	For any $x\in \mathcal{X},$ obviously $\mu_{X_t| \ogood_{k-1}, \omany_{k - 1}, \mathcal{F}_{k-1}}(x) = \mu_{X_t}(x)$ for $t \in \mathcal{T}_k$.  Thus
	\begin{align}\label{eq: density-bound}
		&\mu_{X_t|A_t=a, \ogood_{k-1}, \omany_{k - 1}, \mathcal{F}_{k-1}}(x) \nonumber \\
		& = \frac{\mathbb{P}(A_t=a \mid X_t=x, \ogood_{k-1}, \omany_{k-1}, \mathcal{F}_{k-1})}{\mathbb{P}(A_t=a \mid \ogood_{k-1},\omany_{k - 1}, \mathcal{F}_{k-1})}\mu_{X_t \mid \ogood_{k-1}, \omany_{k -1}, \mathcal{F}_{k-1}}(x) \nonumber \\ & = \frac{\mathbb{P}(A_t=a \mid X_t=x, \ogood_{k-1}, \omany_{k -1}, \mathcal{F}_{k-1})}{\mathbb{P}(A_t=a \mid \ogood_{k-1}, \omany_{k -1}, \mathcal{F}_{k-1})}\mu_{X_t}(x).
	\end{align}

	For any $x \in \bigcup_{\mathcal A_0: a \not\in \mathcal{A}_0}\mathcal{R}_{k - 1}(\mathcal A_0) \cap \mathcal X$, our algorithm ensures that
	\[
		\mathbb{P}(A_t=a|X_t=x, \ogood_{k-1}, \omany_{k - 1},\mathcal{F}_{k-1})=0.
	\]
	Therefore, for $x \in \bigcup_{\mathcal A_0: a \not\in \mathcal{A}_0}\mathcal{R}_{k - 1}(\mathcal A_0) \cap \mathcal X$, $\mu_{a,k}(x)=0$, which proves statement 2.

	For any $x \in \bigcup_{\mathcal A_0: a \in \mathcal{A}_0}\mathcal{R}_{k - 1}(\mathcal A_0) \cap \mathcal X$, our algorithm ensures that
	\[
		 \mathbb{P}(A_t=a|X_t=x, \ogood_{k-1}, \omany_{k - 1}, \mathcal{F}_{k-1}) \ge \frac{1}{|\mathcal A|}.
	\]
	Plus, since $T \ge T_0 \vee \prns{\exp(1 \vee \frac{C_{\mathcal A}(2\beta +d)}{4 (2r_0)^{2\beta}(2\beta+d + \beta d)})} $, \cref{lemma: decision-set-characterization-multi-arm} implies that $\mathcal Q_a \subseteq \bigcup_{\mathcal A_0: a \in \mathcal A_0} \mathcal R_k(\mathcal A_0) \cap \mathcal X$. By following \cref{lemma: nomempty-opt}, we can analogously prove that $\pr(X \in \mathcal Q_a) \ge p$. It follows from  \cref{assump: decision-set} that
	\begin{align*}
	&\mathbb{P}(A_t=a| \ogood_{k-1}, \omany_{k - 1}, \mathcal{F}_{k-1})\\
	= & \mathbb{P}(A_t=a|X_t\in \bigcup_{\mathcal A_0: a \in \mathcal A_0} \mathcal R_k(\mathcal A_0) \cap \mathcal X, \ogood_{k-1}, \omany_{k - 1}, \mathcal{F}_{k-1})  \\
	& \times \mathbb{P}(X_t \in \bigcup_{\mathcal A_0: a \in \mathcal A_0} \mathcal R_k(\mathcal A_0) \cap \mathcal X \mid \ogood_{k-1}, \omany_{k - 1}, \mathcal{F}_{k-1})
	 \\ \ge & \frac{1}{|\mathcal A|}\pr(X_t \in \mathcal Q_a) \ge \frac{p}{|\mathcal A|},
	\end{align*}
	where the last inequality follows from \cref{lemma: nomempty-opt}.

	Then it follows from \cref{assump: support,eq: density-bound} that for any $x\in \bigcup_{\mathcal A_0: a \in \mathcal A_0} \mathcal R_k(\mathcal A_0) \cap \mathcal X$,
	\[
		\frac{1}{|\mathcal A|}\mu_{\min}\le \mu_{a,k}(x)\le \frac{|\mathcal A|\mu_{\max}}{p},
	\]
	which proves statement 1.
\endproof

\proof{Proof of \cref{lemma: eigenvalue-multi-arm}.}
We can condition on $\ogood_{k-1}, \omany_{k - 1}, N_{a,k} = n_{a,k}$, and $\mathcal{F}_{k - 1}$, and then follow the proof of  \cref{lemma: eigenvalue}. The only difference is that now  samples $S_{a, k} = \{(X_t, Y_t): A_t = a, t \in\mathcal{T}_k \} = \{(X_t, Y_t): t \in\mathcal{T}_{k, a} \}$ are i.i.d whose conditional density for $X_t$ has different strong density parameters: $\frac{1}{|\mathcal A|}\mu_{\min}\le \mu_{a,k}(x)\le \frac{|\mathcal A|\mu_{\max}}{p}$ for any $x \in \bigcup_{\mathcal A_0: a \in \mathcal{A}_0}\mathcal{R}_{k - 1}(\mathcal A_0) \cap \mathcal X$. This only influences the estimation error bound of $|(\hat{\mathcal{A}}_{a,k}(x_0))_{r_1,r_2}-(\mathcal{A}_{a,k}(x_0))_{r_1,r_2}|$. Note that
For $t \in S_{a, k}$ and $|r_1|, |r_2| \le \beta$, and $Z_t(r_1, r_2) =  \frac{1}{h_{a,k}^d}(\frac{X_t-x_0}{h_{a,k}})^{r_1+r_2}K(\frac{X_t-x_0}{h_{a,k}})$, we obviously have $|Z_t(r_1, r_2)|\le \frac{1}{h^d_{a,k}}$, and
	\begin{align*}
	&\qquad \expect( Z_t^2(r_1, r_2)| A_t = a, \ogood_{k-1}, \omany_{k - 1}, N_{\mathcal A,k} = n_{a,k}, \mathcal{F}_{k-1})   \\
		&= \frac{1}{h_{a,k}^{2d}}\int_{\mathcal{X}}(\frac{x-x_0}{h_{a,k}})^{2(r_1+r_2)}K^2(\frac{x-x_0}{h_{a,k}})\mu_{a,k} (x)dx\\
		& \le \frac{|\mathcal A|\mu_{\max}}{p h_{a,k}^d}\int_{B_{a,k}(x_0)}u^{2(r_1+r_2)}du
		\le \frac{|\mathcal A|\mu_{\max}}{p h_{a,k}^d}\leb[B_{a,k}(x_0)] \le  \frac{|\mathcal A|\mu_{\max}v_d}{p h_{a,k}^d},
	\end{align*}
	where the first inequality follows from the fact that $\mu_{a, k}(x) \le \frac{|\mathcal A|\mu_{\max}}{p}$ according to \cref{lemma: strong-density-multi-arm}. Similarly, we can use Bernstein inequality to show that
	\begin{align*}
	&\mathbb{P}(|(\hat{\mathcal{A}}_{a,k}(x_0))_{r_1,r_2}-(\mathcal{A}_{a,k}(x_0))_{r_1,r_2}|\ge \frac{\lambda_0}{M_{\beta}^2} \mid  \ogood_{k-1}, \omany_{k - 1},  N_{a,k} = n_{a,k}, \mathcal{F}_{k-1} )
	\le 2\exp(-\frac{3p n_{a,k}h_{a,k}^d\lambda_0^2}{6|\mathcal A|M_{\beta}^4\mu_{\max}v_d+2p\lambda_0 M^2_\beta}).
	\end{align*}
	Then we can apply union bound to show that
	taking union bound over all possible $r_1, r_2$,
	\begin{align*}
	&\mathbb{P}(\sum_{|r_1|,|r_2|\le \floor*{\beta}}  |(\hat{\mathcal{A}}_{a,k}(x_0))_{r_1,r_2}-(\mathcal{A}_{a,k}(x_0))_{r_1,r_2}|\ge \lambda_0 | \ogood_{k-1}, \omany_{k - 1},  N_{a,k} = n_{a,k}, \mathcal{F}_{k-1}) \nonumber\\
	\le & 2M^2_{\beta}\exp(-\frac{3p n_{a,k}h_{a,k}^d\lambda_0^2}{6|\mathcal A|M^4_{\beta}\mu_{\max}v_d+2p\lambda_0 M^2_{\beta}}) \le  2M^2_{\beta}\exp\big\{-C_{\mathcal A}\big(4(1+ L_1\sqrt{d})^2\big)n_{a, k}^{\frac{2\beta}{2\beta + d}}\big\}.
	\end{align*}
	If follows from the inequality above and \cref{eq: eigen-diff,eq: eigen-pop} that with high probability $1 - 2M^2_{\beta}\exp\big\{-C_{\mathcal A}\big(4(1+ L_1\sqrt{d})^2\big)n_{a, k}^{\frac{2\beta}{2\beta + d}}\big\}$,
	\[
	\lambda_{\min}(\hat{\mathcal{A}}_{a,k}(x_0)) \ge \lambda_0.
	\]
\endproof

\proof{Proof of \cref{thm: good-prob-multi-arm}.}
\textbf{I. Prove the bound on $\pr (\good_{k}^C \mid \ogood_{k-1},  \omany_k )$.} We note that when $T \ge \exp(1\vee \frac{36{M}_{\beta}L^2v^2_d\mu^2_{\max}C_{\mathcal A}(2\beta + d)}{p^2\lambda^2_0 (2\beta + d + \beta d)})$, the proof of \cref{thm: good-prob} Step I shows that all conditions in \cref{lemma: convergence} are satisfied.
	We can straightforwardly revise the Step II and step III in the proof of \cref{lemma: convergence} to show that
	\begin{align*}
	&\mathbb{P}(\sup_{a \in \mathcal A}\sup_{x\in \prns{\bigcup_{\mathcal A_0: a \in \mathcal A_0} \mathcal R_k(\mathcal A_0)} \cap \mathcal D_{\mathrm{Irr}, k - 1}^C(a)}\abs{\hat{\eta}_{k, a}(x)-\eta_{k, a}(x)} \le \epsilon_{k}/2 \mid \ogood_{k-1}, \omany_{k - 1}, N_{a,k} = n_{a,k})\\ \le & \delta^{-d}|\mathcal A|{(4 + 2M_{\beta}^2)} \exp(-C_{\mathcal A} n_{a,k}^{\frac{2\beta}{2\beta+d}}\epsilon_k^2).
	\end{align*}
	where $C_0$ is given in \cref{lemma: eigenvalue-multi-arm}.

	Then we can prove the conclusion by following Step I in the proof of \cref{thm: good-prob}.

\textbf{II. Prove the bound on $\pr(\many_k^C \mid \ogood_{k - 1}, \omany_{k-1})$.} We first note that for any $a \in \mathcal{A}$
\begin{align*}
	\expect(N_{a,k} \mid \ogood_{k-1}, \omany_{k - 1})
	&= \expect( \sum_{t\in \mathcal{T}_k}\mathbb{I}\{X_t\in \bigcup_{\mathcal A_0: a \in \mathcal A_0} \mathcal R_k(\mathcal A_0)\} \mid \ogood_{k-1}, \omany_{k - 1})\\
	& = n_k\mathbb{P}(X_t\in \bigcup_{\mathcal A_0: a \in \mathcal A_0} \mathcal R_k(\mathcal A_0)\mid \ogood_{k-1}, \omany_{k - 1})\\
	&\ge \frac{1}{|\mathcal A|}n_k\mathbb{P}(X_t \in \mathcal Q_a \mid \ogood_{k - 1}, \omany_{k - 1}) \ge \frac{p}{|\mathcal A|}n_k.
	\end{align*}
	where the inequality follows from \cref{lemma: decision-set-characterization-multi-arm} statement \ref{decision-set-4}.

	By Hoeffding's inequality,
	\begin{align*}
	&\mathbb{P}\bigg(N_{a,k}< (\frac{\log(T\delta^{-d})}{C_{\mathcal A}\epsilon_k^2})^{\frac{2\beta +d}{2\beta}}\mid \ogood_{k-1}, \omany_{k - 1}\bigg) \\
	\le & \mathbb{P}\bigg(\expect(N_{a,k}\mid \ogood_{k-1}, \omany_{k - 1})-N_{a,k}>\frac{p}{|\mathcal A|}n_k-(\frac{\log(T\delta^{-d})}{C_{\mathcal A}\epsilon_k^2})^{\frac{2\beta +d}{2\beta}}\mid \ogood_{k-1}, \omany_{k - 1}\bigg)\\
	\le & \exp\bigg(-\frac{2}{n_k}\big[\frac{p}{|\mathcal A|}n_k-(\frac{\log(T\delta^{-d})}{C_{\mathcal A}\epsilon_k^2})^{\frac{2\beta +d}{2\beta}}\big]^2\bigg).
	\end{align*}

	When $n_k \ge \frac{2|\mathcal A|}{p}(\frac{\log(T\delta^{-d})}{C_{\mathcal A}\epsilon_k^2})^{\frac{2\beta +d}{2\beta}}+\frac{|\mathcal A|^2}{2p^2}\log T$
	\[
		\frac{2}{n_k}\big[\frac{p}{|\mathcal A|}n_k-(\frac{\log(T\delta^{-d})}{C_{\mathcal A}\epsilon_k^2})^{\frac{2\beta +d}{2\beta}}\big]^2 \ge 2\frac{p^2}{|\mathcal A|^2}n_k - \frac{4}{|\mathcal A|}p(\frac{\log(T\delta^{-d})}{C_{\mathcal A}\epsilon_k^2})^{\frac{2\beta +d}{2\beta}} \ge \log T.
	\]

	Thus
	\[
		\mathbb{P}\bigg(N_{a,k}< (\frac{\log(T\delta^{-d})}{C_{\mathcal A}\epsilon_k^2})^{\frac{2\beta +d}{2\beta}}\mid \ogood_{k-1}, \omany_{k - 1}\bigg) \le \frac{1}{T}.
	\]
	Accordingly,
	\[
	\mathbb{P}\bigg(\min_{a \in \mathcal A} N_{a,k}< (\frac{\log(T\delta^{-d})}{C_{\mathcal A}\epsilon_k^2})^{\frac{2\beta +d}{2\beta}}\mid \ogood_{k-1}, \omany_{k - 1}\bigg) \le \frac{|\mathcal A|}{T}.
	\]
	\textbf{III. Prove the bound on $\pr(\ogood_k^C \cup \omany_{k}^C)$.} We can follow the proof of \cref{thm: good-prob} and note that
	\begin{align*}
		\pr(\ogood_k^C \cup \omany_{k}^C)
			&\le \sum_{j = 0}^{k - 1}\pr\big( \good_{j + 1}^C \cap \omany_{j + 1} \cap \ogood_{j} \big) +  \sum_{j = 0}^{k - 1}\pr\big(\many_{j + 1}^C \cap \ogood_j \cap \omany_j \big) \\
			&\le \sum_{j = 0}^{k - 1}\pr( \good_{j + 1}^C \mid \ogood_{j}, \omany_{j + 1}) +  \sum_{j = 0}^{k - 1}\pr(\many_{j + 1}^C \mid \ogood_j, \omany_j) \\
			&\le \sum_{j = 0}^{k - 1} \frac{(4 + 2M_{\beta}^2)|\mathcal A|}{T} + \frac{|\mathcal A|}{T} = \frac{(5 + 2M_{\beta}^2)k|\mathcal A|}{T}.
	\end{align*}
\endproof

\proof{Proof of \cref{thm: regret-multi-arm}:}
\begin{align*}\textstyle
		R_T(\hat{\pi})
			&=  \sum_{k=1}^K \sum_{t\in\mathcal{T}_k} \expect [Y_{t}(\pi^*(X_t)) - Y_{t}(A_t)] \\
			&\le \sum_{k=1}^K\sum_{t\in\mathcal{T}_k} \expect[ Y_{t}(\pi^*(X_t)) - Y_{t}(A_t) \mid \ogood_{k - 1} \cap \omany_{k - 1}]
			 + \sum_{k=1}^K\sum_{t\in\mathcal{T}_k} \pr(\ogood_{k - 1}^C \cup \omany_{k - 1}^C).
	\end{align*}
Here
\begin{align*}
&\sum_{t\in\mathcal{T}_k} \expect[ Y_{t}(\pi^*(X_t)) - Y_{t}(A_t) \mid \ogood_{k - 1} \cap \omany_{k - 1}] \\
\le& \sum_{t\in\mathcal{T}_k}\sum_{\mathcal A_0: \emptyset \ne \mathcal A_0\subseteq \mathcal A} \expect\left[\max_{a \in \mathcal A}\eta_a(X_t) - \min_{a \in \mathcal A_0}\eta_a(X_t) \mid \ogood_{k - 1} \cap \omany_{k - 1}, X_t \in \mathcal R_k(\mathcal A_0)\right]\pr\prns{X_t \in \mathcal R_k(\mathcal A_0) \mid  \ogood_{k - 1} \cap \omany_{k - 1}}
\end{align*}
Note that by \cref{lemma: decision-set-characterization-multi-arm} statement \ref{decision-set-2}, we never delete optimal arm from the active arm sets, i.e., $\max_{a \in \mathcal A}\eta_a(X_t) - \min_{a \in \mathcal A_0}\eta_a(X_t) = \max_{a \in \mathcal A_0}\eta_a(X_t) - \min_{a \in \mathcal A_0}\eta_a(X_t)$. Moreover, if $|\mathcal A_0| = 1$, then trivially $\expect\left[\max_{a \in \mathcal A}\eta_a(X_t) - \min_{a \in \mathcal A_0}\eta_a(X_t) \mid \ogood_{k - 1} \cap \omany_{k - 1}, X_t \in \mathcal R_k(\mathcal A_0)\right] = 0$, and if $|\mathcal A_0| > 1$, by \cref{lemma: decision-set-characterization-multi-arm} statement \ref{decision-set-3}, $\max_{a \in \mathcal A}\eta_a(X_t) - \min_{a \in \mathcal A_0}\eta_a(X_t) \le 2\epsilon_{k - 1}$ for $X_t \in \mathcal R_k(\mathcal A_0)$.
We further define $\bar{\mathcal{R}}_k = \bigcup_{\mathcal{A}_0: \mathcal{A}_0 \subseteq \mathcal{A}, |\mathcal{A}_0|>1} \mathcal{R}_k(\mathcal{A}_0)$ as the region where more than one arm is pulled with positive probability.
Obviously, \cref{lemma: decision-set-characterization-multi-arm} statement \ref{decision-set-3} and the fact that any hypercube in $\bar{\mathcal{R}}_k$ explores more than one arm together imply that $\bar{\mathcal{R}}_k \subseteq \{x\in \mathcal{X}: \exists a\neq a' \in \mathcal{A}, |\eta_a(x) - \eta_{a'}(x)|\le 2\epsilon_{k-1}\}$.
Combining all these facts, we get
	\begin{align*}
	&\sum_{t\in\mathcal{T}_k} \expect( Y_{t}(\pi^*(X_t)) - Y_{t}(A_t) \mid \ogood_{k - 1} \cap \omany_{k - 1}) \\
	= &  \sum_{t\in\mathcal{T}_k} \sum_{\mathcal{A}_0: \mathcal{A}_0 \subseteq \mathcal{A}, |\mathcal{A}_0|>1} \expect( Y_{t}(\pi^*(X_t)) - Y_{t}(A_t) \mid \ogood_{k - 1} \cap \omany_{k - 1}, X_t \in \mathcal{R}_k(\mathcal{A}_0))\pr(X_t \in \mathcal{R}_k(\mathcal{A}_0) \mid \ogood_{k - 1} \cap \omany_{k - 1})  \\
	\le &  \sum_{t\in\mathcal{T}_k} \sum_{\mathcal{A}_0: \mathcal{A}_0 \subseteq \mathcal{A}, |\mathcal{A}_0|>1} 2\epsilon_{k-1}\pr(X_t \in \mathcal{R}_k(\mathcal{A}_0) \mid \ogood_{k - 1} \cap \omany_{k - 1})  \\
	= &\sum_{t\in\mathcal{T}_k}  2\epsilon_{k-1} \mathbb{P}(X_t\in \bar{\mathcal{R}}_k \mid \ogood_{k - 1} \cap \omany_{k - 1})   \\
	\le & \sum_{t\in\mathcal{T}_k}  2\epsilon_{k-1} \sum_{a, a'\in \mathcal{A}}\mathbb{P}(0 < |\eta_a(x) - \eta_{a'}(x)|\le 2\epsilon_{k-1} \mid \ogood_{k - 1} \cap \omany_{k - 1}) \\
	\le & \MarginL |\mathcal{A}|^2 2^{1+\alpha}\epsilon_{k-1}^{1+\alpha} n_k,
	\end{align*}
	where the second last inequality follows from the margin condition (condition iv in \cref{assump: multi-arm}). This implies that
	\[
		\sum_{t\in\mathcal{T}_k} \expect( Y_{t}(\pi^*(X_t)) - Y_{t}(A_t) \mid \ogood_{k - 1} \cap \omany_{k - 1})  = \Tilde{O}(|\mathcal A|^3 T^{\frac{\beta+d-\alpha\beta}{2\beta+d}}).
	\]

Moreover, we can follow the proof of \cref{thm: regret} to show that the conclusions of \cref{thm: good-prob-multi-arm} imply
\[
	\sum_{k=1}^K\sum_{t\in\mathcal{T}_k} \pr(\ogood_{k - 1}^C \cup \omany_{k - 1}^C) = \tilde O(1).
\]
\endproof

\end{document}